\documentclass{article}

\PassOptionsToPackage{numbers, compress, sort}{natbib}

 \usepackage[main, final]{neurips_2026}

\usepackage[utf8]{inputenc} 
\usepackage[T1]{fontenc}    
\usepackage{url}            
\usepackage{booktabs}       
\usepackage{amsfonts}       
\usepackage{nicefrac}       
\usepackage{microtype}      
\usepackage{xcolor}         

\usepackage[english]{babel}
\usepackage{microtype}  
\usepackage{graphicx}   
\usepackage{wrapfig}
\usepackage{booktabs}
\usepackage{natbib}
\usepackage[colorlinks=true,
            linkcolor=teal,
            citecolor=blue,
            urlcolor=black]{hyperref}
\usepackage{amsmath}
\usepackage{multirow}
\usepackage{subcaption}
\usepackage{caption}
\usepackage[noabbrev]{cleveref}

\usepackage[table]{xcolor}
\usepackage{float}
\usepackage{makecell}
\usepackage{nicefrac}
\title{Visual Instruction Tuning \\ Aligns Modalities through Abstraction}

\makeatletter
\newcommand{\samethanks}[1]{\footnotemark[#1]}
\makeatother

\author{
\textbf{Luis Palacios $^{1,}$}\thanks{Equal contribution.\quad $^\dagger$Equal supervision.}
\And
\textbf{Lorenzo Basile $^{1,}$}\samethanks{1}
\And
\textbf{Diego Doimo $^{1,}$}$^\dagger$
\And
\textbf{Alberto Cazzaniga $^{1,2,3,}$}$^\dagger$
\AND
\normalfont $^1$ Area Science Park, Trieste, Italy \\
$^2$ Université Côte d'Azur, 3IA Côte d'Azur, Nice, France \\  $^3$ Inria, Maasai Project Team, Sophia Antipolis, France
\\ \texttt{{\{firstname.lastname\}@areasciencepark.it
}}
}

\begin{document}

\maketitle

\let\thefootnote\relax\footnotetext{Code is available at \href{https://github.com/RitAreaSciencePark/MultimodalAbstraction}{\texttt{github.com/RitAreaSciencePark/MultimodalAbstraction}}.}

\begin{abstract}
    \looseness=-1Visual instruction tuning effectively adapts a pre-trained Large Language Model (LLM) to process image information alongside text. Yet it remains unclear how visual features are embedded within the layer-wise hierarchy of abstractions in the LLM backbone.
    Across a diverse set of vision-language architectures, we show that instruction tuning primarily serves as a bridge, embedding visual features directly into the intermediate semantic layers of the LLM, bypassing the early layers devoted to unimodal processing.
    With probing analyses and causal interventions, we show that these intermediate layers constitute the semantic core of vision-language processing and play a critical role in performance across a broad set of multimodal benchmarks. 
    In addition, by comparing the geometry of semantically equivalent visual and textual representations, we find that fine-tuning extends and strengthens the existing abstraction phase, aligning visual features with pre-existing textual ones.
    Finally, we confirm the functional role of this localized alignment by restricting fine-tuning to intermediate layers alone: this strategy preserves the performance of full fine-tuning on vision-centric benchmarks while reducing training time.
    Our results suggest that multimodal integration is a localized phenomenon driven by the repurposing of the internal abstraction engine of the LLM.
\end{abstract}

\section{Introduction}

Incorporating visual perception into language-based architectures is a fundamental step toward versatile multimodal systems. Modern generative Vision-Language Models (VLMs) \cite{alayrac2022flamingo, liu2023visual, Qwen-VL, team2025gemma} achieve this by processing visual features extracted from encoders like CLIP \cite{radford2021learning} alongside textual prompts. These models typically build upon Large Language Models (LLMs) pre-trained on massive text corpora, adapted to multimodal tasks via a two-stage fine-tuning pipeline. An initial alignment stage trains a connector to project visual latents into the LLM.
A second visual instruction tuning stage then reshapes the internal representations of the LLM to support multimodal 
understanding.

Despite the empirical success of this approach, the internal mechanisms of multimodal integration remain poorly understood. This contrasts with the increasingly detailed picture emerging for text-only LLMs, which are known to develop a hierarchy of abstract representations during pre-training, with semantic content concentrated in intermediate layers \cite{valeriani2023, doimo2024representation, chengemergence, skeanlayer}.
A recent line of work has further shown that during supervised fine-tuning, the weights that mostly affect downstream performance are located in the middle layers \cite{harada-etal-2025-massive}, and that selectively tuning these layers is sufficient for cross-lingual transfer~\cite{liu-niehues-2025-middle}. 
Parallel evidence from multimodal interpretability suggests that the same intermediate region is where visual and textual features become more similar \cite{neo2025towards, wu2025the} and that cross-modal information is transferred between vision and language tokens in early-to-middle layers \cite{zhang2025cross}.
Together, these findings motivate intermediate layers as a candidate site for cross-modal integration.
The picture, however, remains fragmented: existing analyses are typically confined to a single architecture, methodology, or modality, and stop short of asking whether the alignment of visual features with intermediate textual abstractions is causally responsible for multimodal performance.

In this work, we go beyond these findings and provide interventional evidence across seven open-weight VLM variants and an extensive set of multimodal benchmarks, showing that the intermediate semantic layers are the most critical for multimodal integration. 
Furthermore, for the subset of variants with reproducible fine-tuning pipelines and training data, we show that actively restricting the tuning process to these layers yields performance comparable to full fine-tuning.

\begin{figure*}
\centering
\includegraphics[width=0.9\textwidth]{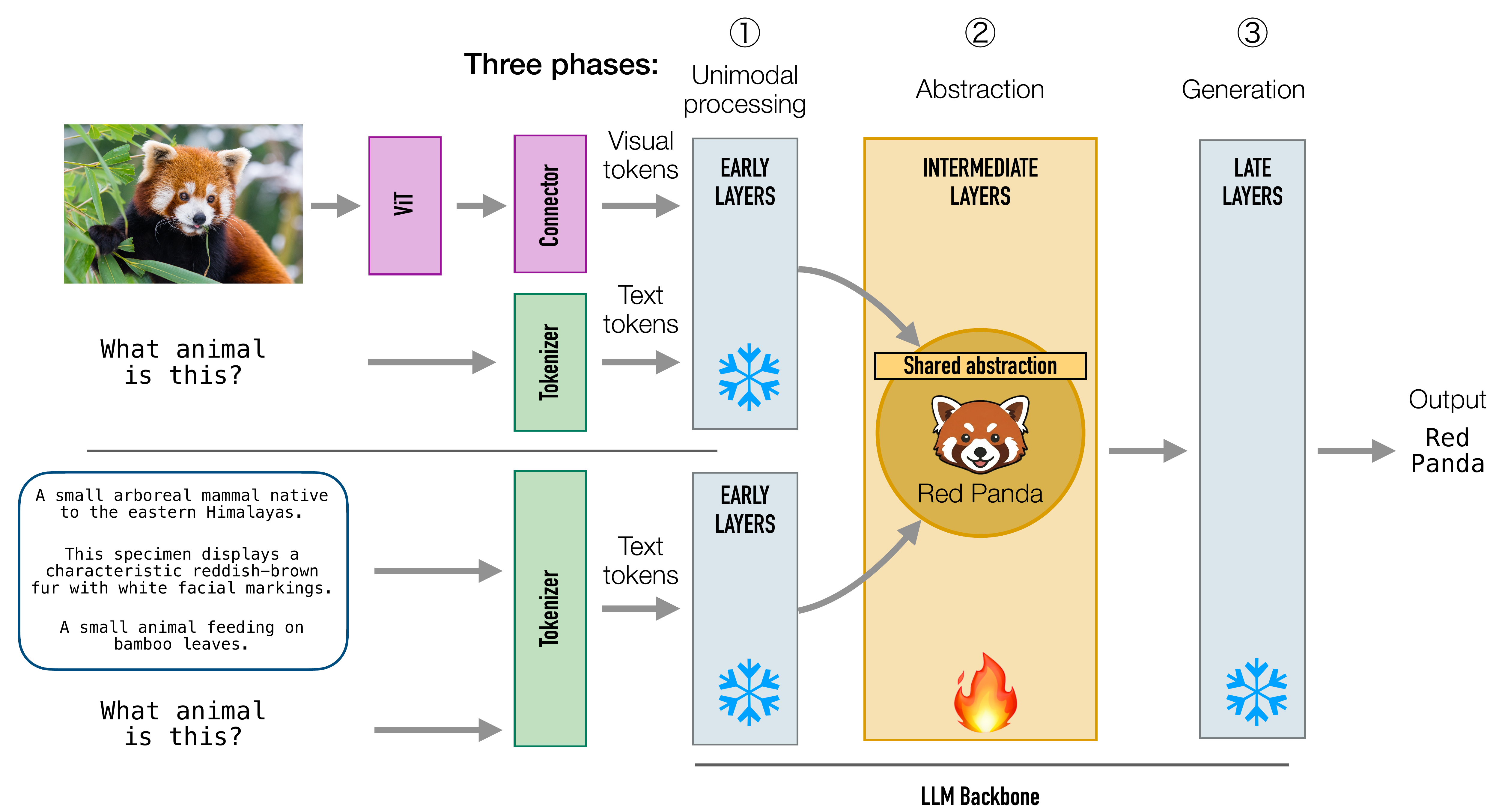}
\caption{
\textbf{Visual and textual abstractions align in the intermediate layers of VLMs.} 
%
\looseness=-1We compare two semantically equivalent inputs: (1) a multimodal prompt, where an image is encoded and injected into the LLM through a visual connector, and (2) a text-only prompt, where the image is replaced by a natural-language caption describing the same content.
For both inputs, we track the hidden representation of the final contextualized input token across the layers of the LLM.
The backbone can be divided into three functional regions: early layers that process modality-specific information, intermediate layers that form shared semantic abstractions, and late layers that specialize in answer generation.
Our results show that visual instruction tuning primarily aligns image-derived representations with pre-existing textual abstractions in the intermediate layers. Moreover, fine-tuning only this region is sufficient to recover nearly the full performance of standard vision-language fine-tuning.
}
\label{fig:cartoon}
\end{figure*}

We summarize our main contributions as follows:
\begin{enumerate} 
\item Through interventional analysis, we show that intermediate layers are \emph{necessary} for vision–language performance: across all layer ranges, their ablation causes the largest performance drop on a diverse set of tasks, identifying them as the main locus of cross-modal communication (\Cref{sec:ablation}); 
\item By probing the semantic content of hidden representations, we find that data abstractions emerge in these intermediate layers. We then compare pre-trained and fine-tuned representations of semantically equivalent visual and textual prompts, revealing a deep impact of instruction tuning in aligning visual and textual abstractions (\Cref{sec:probing,sec:geometry}). 
\item Localized instruction tuning shows that the abstraction layers are also \emph{sufficient} for multimodal integration: restricting fine-tuning to them closely matches full fine-tuning at lower training cost, with the largest gains on the most vision-centric tasks (\Cref{sec:localized_finetuning,sec:corr_vision_centric}). 
\end{enumerate}

\section{Methods}
\label{sec:methods}

\paragraph{Vision-Language Model Architectures.}
We focus our analysis on the following VLMs trained for image understanding tasks: LLaVA-1.5-7B/13B \cite{liu2024improved}, LLaVA-OneVision-1.5-4B/8B \cite{an2025llavaonevision15}, InternVL2-8B \cite{opengvlab2024internvl2}, and Cambrian-8B \cite{tong2024cambrian}. From this point on, for simplicity, we refer to LLaVA-1.5 as LLaVA and LLaVA-OneVision-1.5 as OneVision. These architectures include three main components: an LLM, a vision encoder, and a connector that maps visual embeddings into the LLM.
Multimodal fine-tuning consists of two phases. A Stage-1 phase trains only the connector parameters  to align the visual features with the LLM embedding space. 
A second visual-instruction tuning phase (Stage-2) also trains the LLM and, in some cases, the vision encoder on  an instruction-following dataset. 

We select a set of models to represent a diverse range of design choices in modern VLMs, including model scales, training data, and vision encoder architectures.
To verify that our findings generalize beyond the self-attention-based architecture, we also analyze Llama-3.2-Vision-90B \cite{meta2024llama3.2}, a VLM based on cross-attention. 
We provide more details on these architectures in the Appendix (\Cref{app:architectures}).

\paragraph{Datasets.}
In our experiments, we consider a broad range of vision and vision-language benchmarks, covering multiple-choice and open-ended question answering, captioning, classification, and compositional reasoning.
We also evaluate model performance on text-only benchmarks. Detailed information on the datasets we evaluated is in the Appendix (\Cref{app:datasets}). 

\paragraph{Causal interventions: attention knockout and layer-skipping.}
In \Cref{sec:ablation}, we use two intervention techniques to identify where visual information is processed and transferred within the language model: \emph{attention knockout} \cite{geva2023dissecting} and \emph{layer skipping}. Let $\mathbf{X}^{(l)}=\{\mathbf{x}^{(l)}_1,\ldots,\mathbf{x}^{(l)}_S\}$ denote the hidden states in the layer $l$, divided into text-token indices $\mathcal{T}$ and image-token indices $\mathcal{V}$. For attention head $h$, let $A^{(l,h)}$ be the \emph{pre-softmax} attention matrix. 
With attention knockout, we set $A^{(l,h)}_{ij}=-\infty$ for all text-token queries $i\in\mathcal{T}$ and image-token keys $j\in\mathcal{V}$. This intervention removes image-to-text information flow, but preserves within-modality interactions.
With layer skipping, we also bypass the computation done on visual tokens in a layer range $[l,l')$. In this case, in addition to the cross-modal attention knockout described above, we copy the image embeddings of layer $l$ to layer $l'$. 

\paragraph{Comparing the information content of semantically equivalent multimodal and textual prompts.}
To understand how visual instruction tuning aligns visual representations within the language model backbone, we compare hidden representations produced by prompts that convey the same semantic content via images or text.
Specifically, in \Cref{sec:geometry} we analyze the nearest-neighbor relationships, which are widely used to compare neural representations \cite{kornblith2019similarity, doimo2020hierarchical, glocal2023} and have been shown to reflect semantic similarity in deep layers \cite{vangansbeke2020scan, bergman2020deepnearestneighboranomaly, Boleda:2020}.

Since the transformer backbone is pre-trained on text, textual representations are expected to encode richer information about the prompts than visual ones. 
To capture this asymmetry, we use the \emph{information imbalance}, $\Delta$ \cite{glielmo2022ranking}, a nearest-neighbor-based measure of how well the local geometry of a neural representation predicts that of another \cite{chengemergence, camboulin2024understanding, acevedo2025quantitativeanalysissemanticinformation}.
We compare two sets of representations,
$X_{\mathrm{img}}=\{\mathbf{x}^{\mathrm{img}}_1,\ldots,\mathbf{x}^{\mathrm{img}}_N\}$ and
$X_{\mathrm{text}}=\{\mathbf{x}^{\mathrm{text}}_1,\ldots,\mathbf{x}^{\mathrm{text}}_N\}$,
where $\mathbf{x}^{\mathrm{img}}_i$ is the hidden representation of the final token in a visual question-answering prompt and $\mathbf{x}^{\mathrm{text}}_i$ is the corresponding representation for a text-only prompt in which the image is replaced by a caption conveying the same information.
Let $r^{X}_{ij}$ denote the rank of point $j$ among the neighbors of point $i$ in representation space $X$, computed using the Euclidean distance, and $n_i^{\mathrm{img}}$ denote the index of the nearest neighbor of sample $i$ in $X_{\mathrm{img}}$. 
The information imbalance from visual to textual representations is:
\begin{equation} 
\Delta_{\text{img}\rightarrow \text{text}} = \frac{2}{N^2}\sum_{i}r_{i\,n_i^{\text{img}}}^{X_{\text{text}}}. 
\label{eq:inf_imb} 
\end{equation} 
$\Delta_{\text{img}\rightarrow \text{text}}$ measures how much, on average, nearest-neighbors in $X_{\mathrm{img}}$ are preserved in $X_{\mathrm{text}}$. 
If the two data representations are statistically independent, the expected rank of a visual nearest neighbor in the textual space is $N/2$ and 
$
\mathbb{E}\!\left[\Delta_{\mathrm{img}\rightarrow\mathrm{text}}\right] = 1,
$
whereas a mutual geometric alignment implies that $\Delta_{\mathrm{img}\rightarrow\mathrm{text}}$ and $\Delta_{\mathrm{text}\rightarrow\mathrm{img}}$ are close to $0$.
Importantly, the metric is directional: 
$\Delta_{\mathrm{img}\rightarrow\mathrm{text}}
\neq
\Delta_{\mathrm{text}\rightarrow\mathrm{img}},
$
and allows us to determine whether one modality contains information that is not fully represented in the other. 

For completeness, we also report two measures of symmetric similarity, linear CKA \cite{kornblith2019similarity} and Neighborhood Overlap \cite{doimo2020hierarchical}, which quantify overall representational similarity but cannot capture asymmetries in information content.

\paragraph{Visual instruction tuning setup.} 
\looseness = -1 In \Cref{Sec:results_finetuning}, we perform the Stage-2 visual instruction tuning on LLaVA-7B and OneVision-4B, which are selected for their publicly available and fully reproducible training pipelines. 
Following their procedures, we train the MLP connector and the LLM for LLaVA, as well as the ViT for OneVision. 
We use officially supported training datasets for both models: the standard 665k instruction mixture for LLaVA, and the LLaVA-NeXT 780k data recipe \cite{liu2024llavanext} for OneVision. 
All training hyperparameters match the original ones \cite{liu2024improved, an2025llavaonevision15} (see Appendix, \Cref{sec:hyperparameters}).

\paragraph{Computational Resources.}We run most of our experiments on a NVIDIA H100 GPU with 80GB of VRAM.
We used 3 H100s for the analysis on Llama 3.2-90B and 4 H100s for fine-tuning runs.

\section{Multimodal Integration and Alignment in Intermediate Abstraction Layers}\label{sec:results_representations}
\begin{figure*}[t]
\centering
\includegraphics[width=\linewidth]{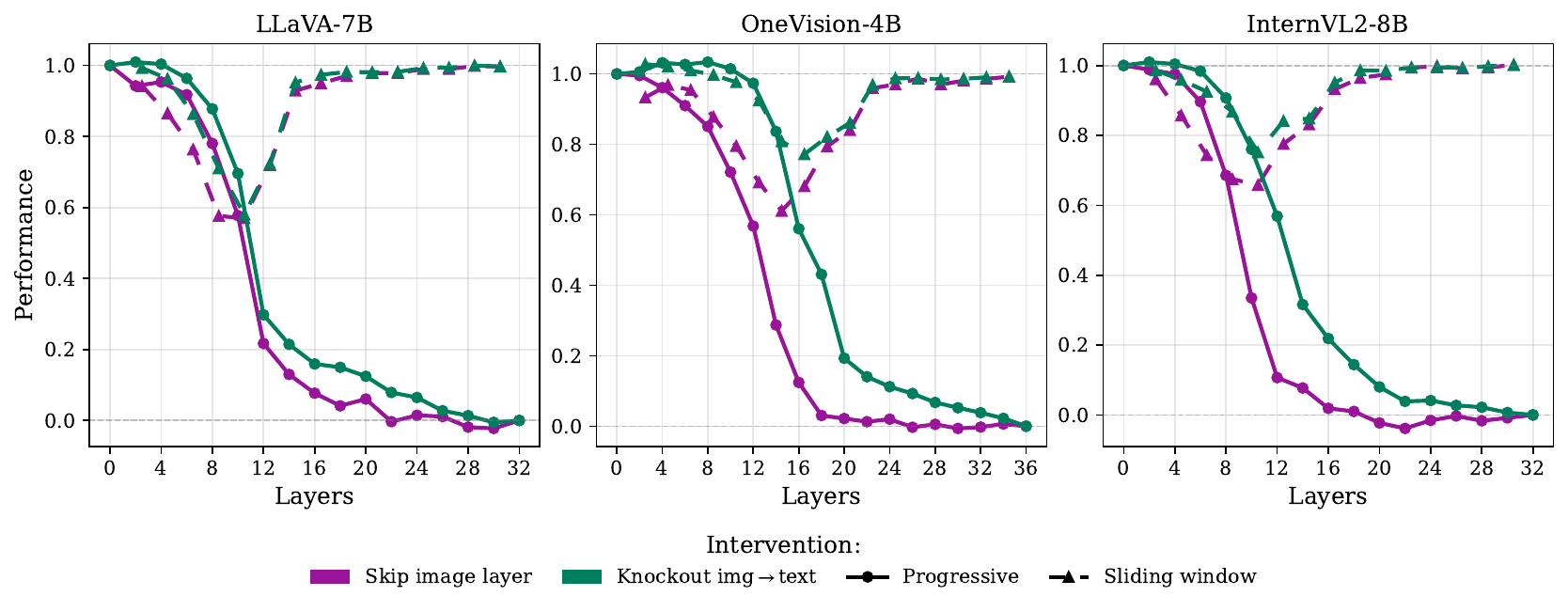}
\caption{
\textbf{Intermediate layers drive vision--language performance.}
Average performance over six vision--language tasks (GQA, MMBench-cn, MMBench-en, MME, SeedBench, and VQAv2; 2,000 samples each) for LLaVA-7B, OneVision-4B, and InternVL2-8B, normalized to the full model.
Purple curves skip image-token layers and green curves knock out image-to-text attention.
Solid lines show progressive interventions from early to late layers, while dashed lines show five-layer sliding windows.
}
\label{fig:interventions_average}
\end{figure*}
A primary objective of visual instruction tuning is to enable effective integration of visual and linguistic representations, which are inherently misaligned at the output of the vision encoder~\cite{mind_the_gap}.
Despite the convergence of many modern VLM proposals to a standard architecture, with a visual encoder and a pre-trained LLM bridged by a multimodal connector that handles the alignment, understanding whether the visual embeddings align with the textual ones in early  \cite{wybitul2026representationstextimagesalign} or middle layers \cite{zhang2025cross} remains a matter of debate. 
In this section, we show that multimodal fusion emerges predominantly in the intermediate layers, which are functionally critical for performance in multimodal tasks (\Cref{sec:ablation}), rich in high-level semantic information (\Cref{sec:probing}), and where the cross-modal alignment is maximal after visual instruction tuning (\Cref{sec:geometry}).

\subsection{Analyzing the functional role of the transformer blocks through causal interventions}
\label{sec:ablation}
We begin our analysis by examining whether and how much input visual features change in the early layers, and where cross-modal integration occurs.

To answer the first question, we apply the layer-skipping intervention to the image embeddings. In \Cref{fig:interventions_average}, we report in purple the impact on average performance on six vision-language tasks (see \Cref{app:ablation_studies}) as we skip an increasing number of layers from the input (full lines) and apply the intervention on a window of five consecutive layers (dashed profiles). 
Performance is normalized so that 1 corresponds to the unmodified model, and 0 to removing image embeddings from all layers. In some cases, normalized performance falls below 0 or exceeds 1 when intervening on a subset of layers degrades or improves performance more than removing or retaining the image information entirely.
For all models, removing the image features from the first six layers leaves the average performance above 90\% of that of the standard model, and only around half of the network, the performance drops to a random level. 
The sliding window ablation further shows that the groups of layers that cause the largest performance drop are around layer 8 for LLaVA (left), 16 for OneVision (center), and 12 for InternVL2 (right), but removing the image at the beginning or at the end of the network does not affect performance. 
The impact of layer-skipping for pure textual inputs instead follows a strikingly different trend (see \Cref{fig:app:text_layer_skip_progressive}). Removing only a few early layers causes performance on language tasks to collapse rapidly to near-random levels, and only the second half is robust to removing small layer windows (see \Cref{fig:app:text_layer_skip_sliding}). 
This different behavior can be partially explained by the previous observations that early layers ``de-tokenize'' raw tokens to construct concepts or word phrases from word pieces~\cite{elhage2022solu}. Such highly specific linguistic operations are largely bypassed by image embeddings, consistent with the ``mid-layer leap'' identified by \cite{krojer2026latentlens}.

Green profiles instead show the impact of attention knockout on the cross-modal attention weights. This experiment allows us to identify where cross-modal communication occurs, since with this intervention we block only the information flow from image to text tokens.
In LLaVA, image processing and cross-modal communication are simultaneous, while the rightward shift of the green profiles in OneVision shows that there is a range of early layers (6 to 12) with unimodal processing, since the performance is unaffected by the cross-modal attention knockout, but is greatly damaged if the visual embeddings are removed. Analogous results on additional models are reported in \Cref{fig:app:interventions_average}.

In summary, layer-skipping and attention knockout show that cross-modal integration is primarily localized to intermediate layers, whereas early layers focus solely on unimodal token processing.

\subsection{Probing the semantic content of intermediate representations on vision-language tasks} 
\label{sec:probing}
\begin{figure*}
\centering
\includegraphics[width=0.8\linewidth]{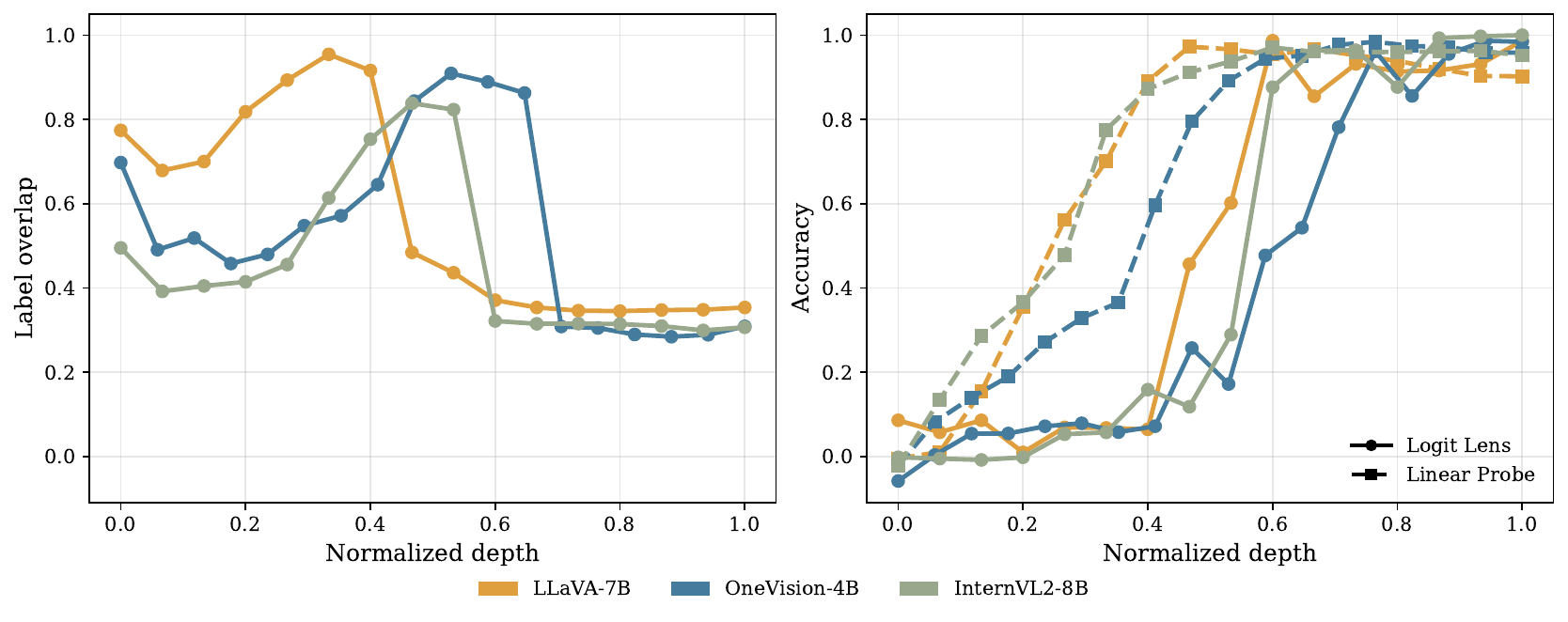}
\caption{
\textbf{Probing semantic information across layers.}
\looseness=-1
\textbf{Left:} Label overlap of last-token residual-stream representations on seven 10-option MCQ classification datasets.
\textbf{Right:} Chance-adjusted MCQ accuracy from the logit lens (solid) and a linear probe (dashed) on vision--language benchmarks (2,000 samples each).
Colors denote LLaVA-7B (orange), OneVision-4B (blue), and InternVL2-8B (green).
}
\label{fig:probing}
\end{figure*}
\Cref{fig:probing} reports two complementary experiments aimed at probing the semantic content of the representations of the last token of the prompt in LLaVA-7B (yellow), OneVision-4B (blue), and InternVL2-8B (green). The results of the other models are shown in \Cref{sec:app_probing} (Appendix).

\looseness=-1The left panel shows the average classification performance across seven datasets (see \Cref{app:sec:no}), each with at least 10 classes, framed as a 10-option multiple-choice VQA task. In this case, the partition into classes is an obvious semantic abstraction of the prompts.
We measure the performance with label overlap (LO) \cite{doimo2020hierarchical}, a $k$NN metric that quantifies the average fraction of $k$-nearest neighbors of a point sharing the same class. 
This allows us to probe at which layers the nearest neighbors encode the same high-level abstraction. Moreover, to further understand in which layers the neighborhood structure inherently represents the class abstraction independent of the specific prompt formulation, we assign the 10 classes to different letter options for each data point (e.g., \emph{"A: dog, B: cat"} for the first sample, \emph{"A: cat, B: dog"} for the second sample, etc.). 
The average LO increases in the first layers and reaches a peak of 0.9 in layer 12 (normalized depth $\sim$ 0.4) for LLaVA, 20 ($\sim$ 0.5) for OneVision, and 16 ($\sim$ 0.5) for InternVL2. 
These layer ranges are also the pivotal ones for the image integration and cross-modal communication, as we discussed in  \Cref{sec:ablation}. 
In later representations, the neighborhood structure rearranges to match the letter corresponding to the correct class, and LO falls below 0.4.

\looseness=-1The right panel confirms this by probing performance on multiple-choice visual question-answering benchmarks (ScienceQA, MMBench, MME, and POPE) for which we do not have explicit abstract labels that would allow us to group dataset instances into classes.
At the end of the abstraction phase, around layer 14 (depth $\sim$ 0.4) for LLaVA, 16 ($\sim$ 0.5) for InternVL2, and 20 ($\sim$ 0.6) for OneVision-4B, the performance of a logistic regression classifier reaches a plateau approximately matching the accuracy of the full model, while Logit Lens \cite{nostalgebraist2020interpreting} performance increases later, suggesting that mid-late layers serve to align already linearly separable answers with the output vocabulary to predict the correct letter.
In the Appendix, we show that a consistent identification of the abstraction layers extends to multimodal information retrieval (\Cref{app:sec:multimodal_retrieval_probing}) and to probing the semantic linguistic content of the representations (\Cref{app:sec:text_semantic_probing}).

Overall, these results demonstrate that in the early-mid layers, the representations accumulate the semantic properties of the inputs, while the last layers' function is to determine the correct letter-to-class assignment and align the representation with the unembedding directions for generation.  

\subsection{The geometric alignment of multimodal representations}
\label{sec:geometry}
Building on our observation that visual information is primarily processed in intermediate, semantically rich layers, we examine how this integration evolves during visual instruction tuning by comparing pairs of prompts that convey the same semantic information in either visual or purely textual form. 

\begin{figure}
\centering
\begin{minipage}{0.48\textwidth}
    \centering
    \includegraphics[width=\linewidth]{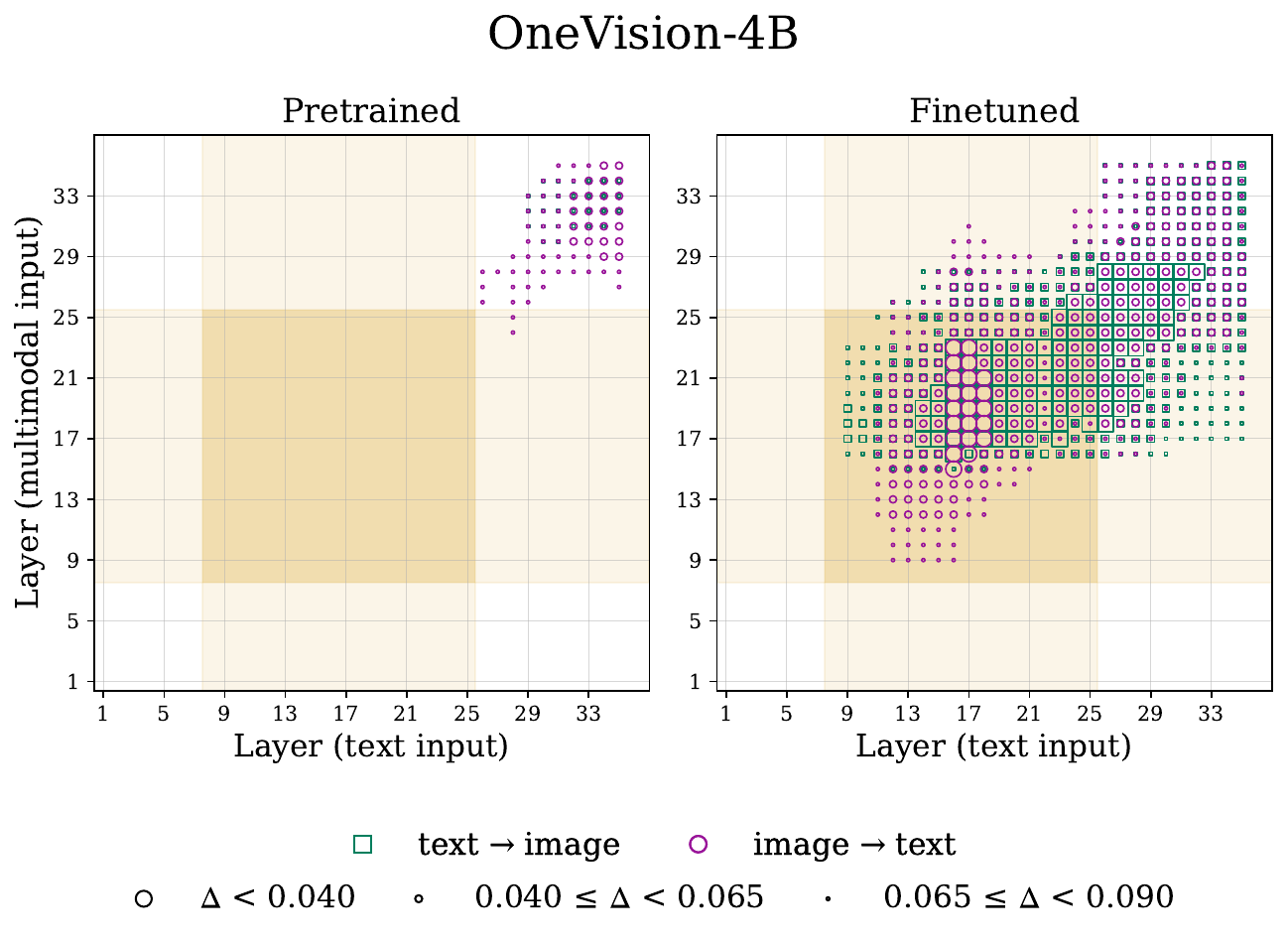}\hfill
\end{minipage}\hfill
\begin{minipage}{0.48\textwidth}
    \centering
    \includegraphics[width=\linewidth]{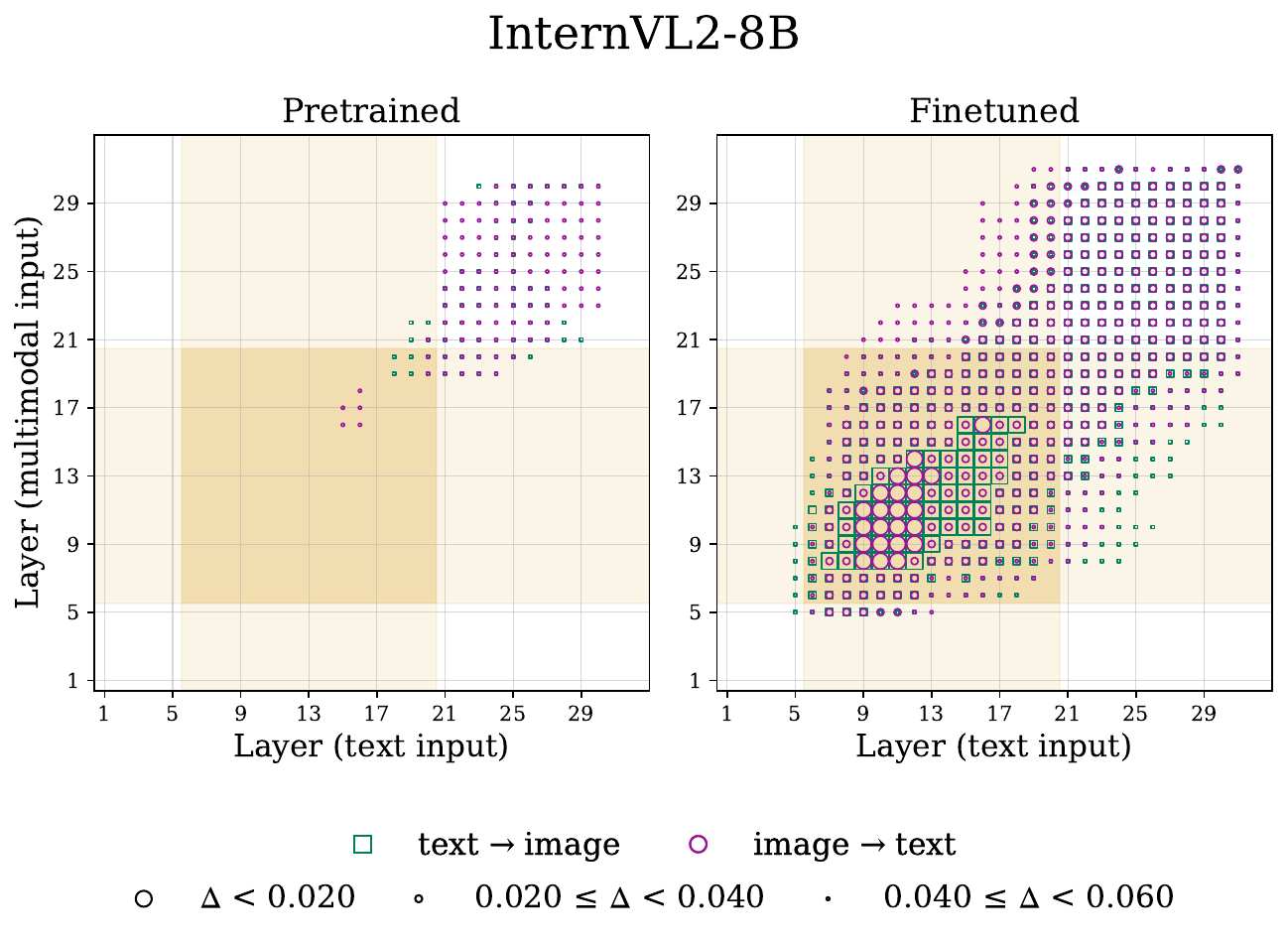}\hfill
\end{minipage}\hfill
\caption{
\textbf{Cross-modal alignment measured by information imbalance ($\Delta$).}
For OneVision-4B (left) and InternVL2-8B (right), we compare hidden representations from multimodal VQA prompts (y-axis) with equivalent text-only prompts where the image is replaced by a caption (x-axis), before and after instruction tuning.
Green squares show text-to-image predictability, and purple circles the opposite direction.
Larger markers indicate lower $\Delta$ and stronger alignment.
The yellow band marks the abstraction region.
Values are averaged over VQAv2, COCO-QA, GQA, SEED, and MMBench (3,000 prompts per dataset; the full MMBench).
}
\label{fig:information_imbalance}
\end{figure}

Concretely, we compare the last-token hidden representations of visual question-answering prompts from VQAv2, COCO-QA, GQA, SEED, and MMBench-EN with those of equivalent prompts in which the images are replaced with captions.
For VQAv2 and COCO-QA, we use human-annotated captions from COCO, and for the other datasets, we prompt Qwen3-VL \cite{bai2025qwen3vltechnicalreport} to generate synthetic descriptions.
In \Cref{fig:information_imbalance}, we show the relative information content of visual and textual representations, measured with the information imbalance, $\Delta$  (see \Cref{sec:methods}).
Purple markers indicate how well visual representations predict the information content of the equivalent matching textual ones, while green markers measure the predictive power in the opposite direction.
Larger marker size corresponds to lower imbalance and stronger alignment.

We report imbalance values for OneVision-4B (left) and InternVL2-8B (right) in their pre-trained and instruction-tuned variants; the corresponding plots for the other models are in \Cref{app:sec:imbalances}. 
In all the cases, after instruction tuning, semantically equivalent visual and textual inputs develop strongly mutually informative representations in mid-network layers consistent with those that develop explicit semantic abstractions of the data (see \Cref{fig:probing}). 
In OneVision-4B, these are around layer 16-22; in InternVL2, between layers 8 and 13 (for LLaVA-7B from 6 to 13, see \Cref{fig:app:information_imbalance_llava} of the Appendix)

Moreover, the values of the information imbalance show, more manifestly for OneVision, that the textual representations are, on average, more informative than the visual representations, especially after the abstraction phase, as indicated by larger green markers from layer 17 to 27. 
This suggests that the text modality, on which the LLM backbone was originally pre-trained, remains more informative after visual instruction tuning. 
At the same time, visual representations continue to retain predictive power over later textual representations, although to a lesser extent.
This indicates that the conceptual abstractions formed from visual inputs persist in later layers and that, once multimodal alignment is established, the originating modality becomes less important.
A similar trend is observed with symmetric representation similarity measures such as linear CKA and neighborhood overlap (see Appendix, \Cref{fig:app:linear_cka_llava,fig:app:neighborhood_overlap_llava}), even though these are inherently unable to capture the directional asymmetry between textual and visual representations.%

In summary, visual instruction tuning substantially increases the alignment between semantically equivalent visual and textual representations. This effect is concentrated in the intermediate abstraction layers previously identified by interventions and probing, while later layers preserve a residual asymmetry in which textual representations remain more informative than visual ones.

\section{The impact of fine-tuning on different VLM layers}\label{Sec:results_finetuning}

Our previous analysis suggests that the primary effect of visual instruction tuning is the expansion and the cross-modal alignment of a specific abstraction region within the LLM backbone. 
To further validate the functional role of these layers, we evaluate a localized fine-tuning strategy. By restricting instruction tuning to the weights of the identified abstraction layers, we will show that multimodal integration can be achieved through a targeted adaptation of the network parameters.

\begin{figure}[t]
\centering
\begin{minipage}[t]{0.48\textwidth}
    \vspace{0pt}
    \centering
    \fontsize{8pt}{9pt}\selectfont
    \setlength{\tabcolsep}{3pt}
    \begin{tabular}{lcccc}
    \toprule
     & \multicolumn{2}{c}{LLaVA-7B} & \multicolumn{2}{c}{OneVision-4B} \\
    \cmidrule(lr){2-3} \cmidrule(lr){4-5}
    Trained layers & $[6, 16]$ & $[6, 16]^C$ & $[8, 25]$ & $[8, 25]^C$ \\
    \midrule
    MC VQA $\uparrow$        & 98.2 & 92.5 & 101.2 & 98.5  \\
    OE VQA $\uparrow$        & 100.2 & 95.0 & 99.1  & 95.0  \\
    Compositional $\uparrow$ & 101.1 & 95.1 & 100.3 & 100.5 \\
    Captioning $\uparrow$    & 98.5 & 97.1 & 98.3  & 99.0  \\
    \midrule
    Overall $\uparrow$       & 98.9 & 93.7 & 100.4 & 98.0  \\
    Train time $\downarrow$  & 86.1 & 92.4 & 75.7  & 79.0  \\
    \bottomrule
    \end{tabular}

\end{minipage}
\hfill
\begin{minipage}[t]{0.48\textwidth}
    \vspace{0pt}
    \centering
    \includegraphics[width=\linewidth]{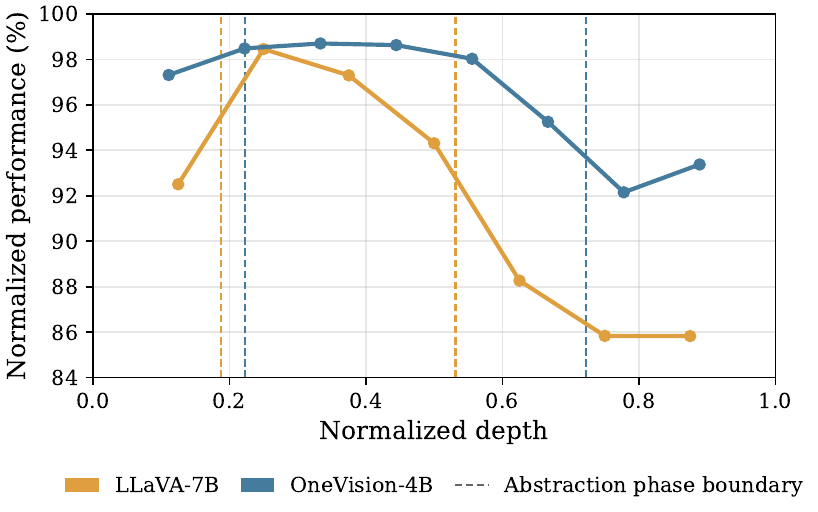}
\end{minipage}%

\caption{
    \textbf{Evaluation of localized fine-tuning strategies on LLaVA-7B and OneVision-4B.}
    \textbf{Left:} Comparison between the selected abstraction layers and their complementary window. Results are normalized to the corresponding fully fine-tuned model.
    MC VQA averages 12 multiple-choice tasks; OE VQA averages 4 open-ended tasks; Compositional averages ARO and SugarCrepe; Captioning averages COCO and Flickr30k.
    Overall averages all tasks.
    Train time refers to visual instruction tuning time on 4 NVIDIA H100 GPUs. Detailed results for all benchmarks are reported in \Cref{fig:app:radar_localized_ft}.
    \textbf{Right:} Comparison between equally sized windows (8 consecutive layers), located at variable depths within the models. The marker is placed at the center of the window, while the vertical dashed lines represent the start and end of the identified abstraction regions.
    }
    \label{tab:performance_ft}
\end{figure}

\subsection{Localized Visual Instruction Tuning}
\label{sec:localized_finetuning}
We compare the effects of fine-tuning the intermediate abstraction phase and its complement in LLaVA-7B and OneVision-4B, since their training code and data are publicly available and their fine-tuning costs are manageable on our current computational infrastructure. We use layers $[6, 16]$ for LLaVA and $[8, 25]$ for OneVision. We do not aim to identify precise layer boundaries: windows are chosen generously to cover the abstraction region identified in \Cref{sec:results_representations}, and we expect small shifts to have negligible effect. A quantitative criterion yields nearly identical windows: selecting the layers in which the information imbalance between semantically equivalent visual and textual representations (see \Cref{sec:geometry}) decreases by at least $85\%$ after instruction tuning, relative to the pretrained checkpoint, gives $[6, 16]$ for LLaVA and $[9, 25]$ for OneVision.
Following the original training procedures, we train the encoder only for OneVision, while we train the multimodal connector for both models. 
For the complementary layer windows, which include the early and late layers of the LLMs, we additionally train the embedding and unembedding parameters.

In \Cref{tab:performance_ft} (\textit{left}), we report the performance of models fine-tuned according to these strategies, normalized with respect to the fully-trained baselines (available in the Appendix, \Cref{tab:performance}) and averaged across four task categories:
multiple-choice visual question answering (\textit{MC VQA}) comprising CVBench, MMBench-Chinese, MMBench-English, MME, MMStar, MMMU, MMMU-Pro, Naturalbench, POPE, RealWorldQA, SEEDBench, and ScienceQA;
open-ended visual question answering (\textit{OE VQA}) including ChartQA, CountBench, GQA, and VQAv2;
\textit{Compositional} including ARO and SugarCrepe; and %
\textit{Captioning} with COCO and Flickr30k. 
The results show that fine-tuning the abstraction region successfully recovers the performance of the fully trained baselines.
For LLaVA, despite training only one-third of the LLM layers, we can achieve an overall mean performance of $98.9\%$. Similarly, for OneVision, training half the layers matches the base performance ($100.4\%$).
Localized configurations are consistently more effective than their complements $[6, 16]^C=[0,5]\cup[17,31]$ and $[8, 25]^C=[0,7]\cup[26,35]$. The complementary windows fail to recover the same performance despite having an identical number of layers for OneVision and 10 more trainable layers for LLaVA, lagging behind the abstraction window performance by 2$\%$ and $5\%$, respectively. We report the complete results on all benchmarks in the Appendix (\Cref{fig:app:radar_localized_ft}). 
Beyond maintaining accuracy, this localized strategy improves training efficiency. By updating only the layers in the abstraction region, the models achieve performance comparable to full fine-tuning while reducing training time by $14\%$ for LLaVA and $24\%$ for OneVision. 
As we further show in \Cref{tab:performance_ft} (\textit{right}), the position of the layer window, and not its size, is the main driver of performance. To isolate the effect of position, we fine-tune windows of 8 consecutive layers placed at different depths of the network, keeping the number of trainable layers fixed. On both models, these windows are maximally effective in terms of average performance when placed in the early-to-intermediate layers of the network, where they substantially overlap with the chosen abstraction regions.

In summary, fine-tuning only the middle abstraction layers of LLaVA and OneVision matches or slightly improves the performance of training the entire model, whereas training the outer layers results in a clear drop in accuracy. As a practical advantage, this targeted approach improves efficiency, cutting training time by up to $24\%$ without sacrificing benchmark results.
\subsection{Fine-tuning the intermediate layers is beneficial for vision-centric tasks}
\label{sec:corr_vision_centric}
\begin{figure*}
\centering
\includegraphics[width=\textwidth]{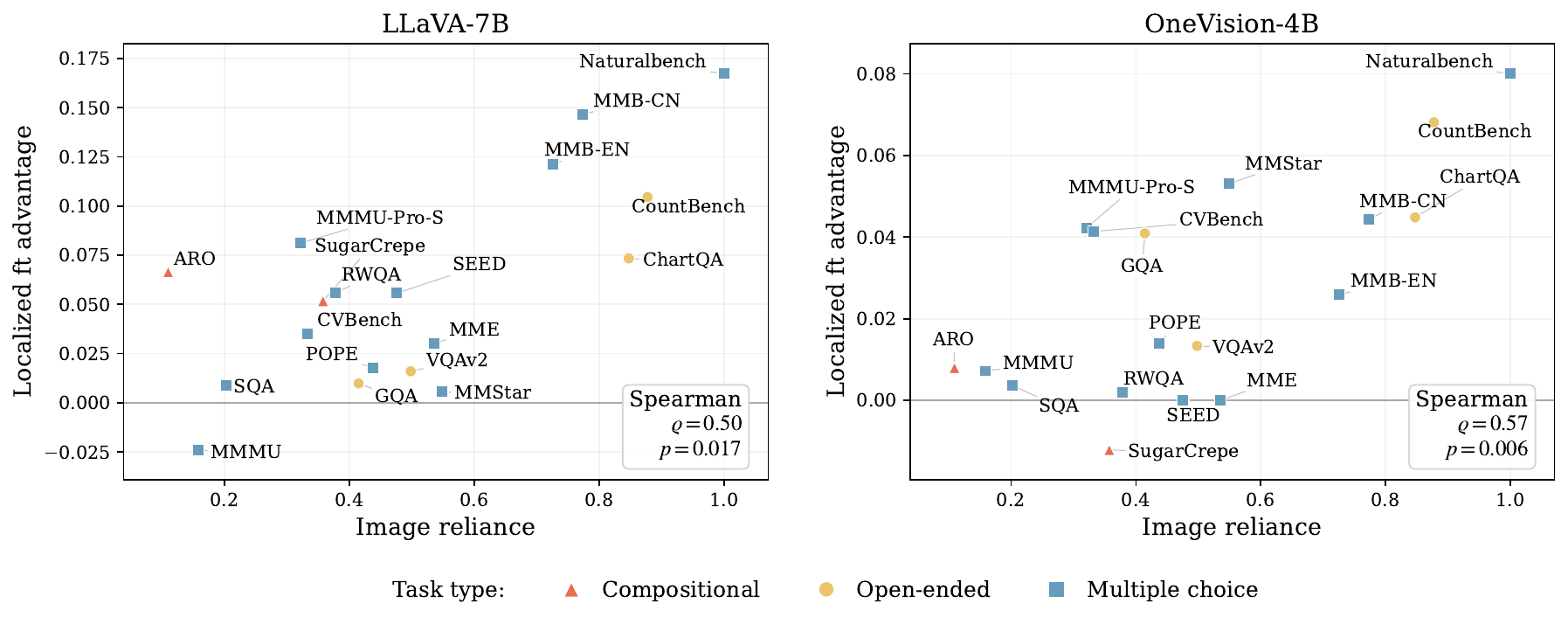}
\caption{
\textbf{Localized fine-tuning helps most when the image matters.}
Each point is a benchmark. 
The x-axis measures image reliance: the normalized gap between image-conditioned and text-only performance, averaged over four reference VLMs (up to 5,000 samples per benchmark). 
The y-axis measures the gain from fine-tuning the abstraction layers instead of their complement, for LLaVA-7B (left) and OneVision-4B (right).
Benchmarks that rely more on visual input benefit more from localized tuning.
}
\label{fig:visual_reliance_correlation}
\end{figure*}
In this section, we further analyze the performance disparity between the identified abstraction layers and their functional complement. 
As we report in detail in \Cref{fig:app:radar_localized_ft} (Appendix), our fine-tuning runs localized to the identified abstraction layers generally outperform training the complementary set of layers, even though the latter contains an equal or larger number of layers. In 14 of 20 total benchmarks evaluated, our localized window outperforms its complement for both LLaVA and OneVision; on the remaining six, results are mixed, and no benchmark favors the complement in both.
Upon closer examination of these performance gains, we observe that they are particularly notable in a specific subset of tasks, such as MMBench (both English and Chinese), ChartQA, CountBench, and NaturalBench. 
We hypothesize that this advantage stems from the visual complexity of these benchmarks. Success on these tasks requires substantial multimodal integration, preventing the model from simply relying on text-based assumptions or unimodal biases. 
To quantify this relationship, we introduce an Image Reliance (IR) metric, which is defined as the normalized performance drop of a given VLM $m$ when visual input is withheld:
\begin{equation}\label{eq:ir}
    \text{IR($m$, dataset)}=\frac{\text{score($m$, image-text)-score($m$, text only)}}{\text{score($m$, image-text)}.}
\end{equation}
Then, since we aim to determine a model-agnostic measure of image-centricity of a dataset, we average this metric over a range of four high-performing VLMs (InternVL2-8B \cite{opengvlab2024internvl2}, OneVision-8B \cite{an2025llavaonevision15}, Cambrian-8B \cite{tong2024cambrian}, Pixtral-12B \cite{pixtral12b}).
In \Cref{fig:visual_reliance_correlation}, we plot the normalized performance gap between the localized fine-tuned models ($[6,16]$ for LLaVA \textit{(left)} and $[8,25]$ for OneVision \textit{(right)}) and their complements against this average reliance metric for all 18 benchmarks. We exclude image captioning tasks from this analysis as these are trivially impossible to solve without relying on the image.
Our analysis reveals a clear positive association between the two variables across both models, supported by a Spearman correlation of approximately $\varrho = 0.5$ with high statistical significance ($p < 0.05$). Image-centric tasks such as MMBench, CountBench, and Naturalbench consistently appear in the upper right corner of the plots, confirming both their high reliance on visual information and their performance gains when the abstraction phase is tuned. 
Conversely, on less vision-centric tasks, where the model can rely more heavily on textual context (such as ScienceQA or MMMU), the performance gap between the two tuning strategies narrows or even slightly inverts. To verify generalization across multimodal fusion strategies, we evaluate Llama-3.2-Vision-90B in the Appendix (\Cref{fig:app:llama90b_visual_reliance}). Instead of performing localized fine-tuning, we selectively activate its cross-attention layers, and we confirm that the performance advantage of intermediate abstraction layers over their complement strongly correlates with image reliance.

Overall, the performance advantage of fine-tuning only the middle abstraction layers is particularly marked on vision-heavy tasks. This confirms that these specific layers bear primary responsibility for successfully integrating visual cues into textual processing.

\section{Related Work and Discussion}
Our work shows that visual instruction tuning does not uniformly modify the representation space within the LLM backbone, but rather focuses on an intermediate group of semantically rich layers. 
We establish this result across seven VLM architectures using causal interventions, semantic probing, and geometric similarity analysis of semantically equivalent visual and textual representations. 
We finally validate the functional role of this intermediate region through localized fine-tuning: updating only the intermediate layers approximately preserves full fine-tuning performance, especially on vision-centric tasks.

\paragraph{Semantic abstraction in intermediate layers of unimodal transformers.} \looseness=-1The presence of an intermediate abstraction phase in autoregressive transformers processing unimodal inputs was previously observed
in the visual \cite{valeriani2023} and textual \cite{chengemergence, skeanlayer} domains. 
Our findings are consistent with those results and further show that after instruction tuning, VLMs preserve the hierarchical organization of abstraction of the underlying LLM backbone, while assigning multimodal understanding to the most semantically rich layers.
The evidence that visual instruction tuning strengthens and expands cross-modal alignment in intermediate layers also extends work on unimodal representation similarity, where independently trained image and text models have been shown to develop convergent abstract representations \cite{pmlr-v235-huh24a,acevedo2025quantitativeanalysissemanticinformation}. 
While those studies compare models trained separately, we examine how visual and textual abstractions align within a single multimodal backbone during instruction tuning. 

\paragraph{Connection with vision-language interpretability studies.} This question is also central to recent interpretability work on VLMs, which has offered contrasting accounts of where multimodal integration occurs. 
Work employing logit lens \cite{nostalgebraist2020interpreting} and transfer-based analyses argue for a shared intermediate semantic hub \cite{wu2025the}, and analyses using sparse autoencoders \cite{venhoff2025visualrepresentationsmaplanguage}, circuit tracing \cite{nikankin2025same}, find progressive alignment in late representations. Conversely, \cite{zhang2025cross, wybitul2026representationstextimagesalign} find alignment between visual and textual concepts in the early layers of VLMs. Other recent works have shown the prominent role of specific attention heads in intermediate-late transformer blocks for multimodal processing \cite{basile2025head,ortu-etal-2026-seeing}. 
Our results go beyond these earlier studies by combining a broader methodological framework with more extensive experimental validation to reveal a new link between semantic abstraction and multimodal integration. While \cite{wu2025the} provides correlational evidence for the existence of a transmodal hub across pre-trained models, we present causal evidence that this hub is the functional core of multimodal integration during visual instruction tuning. Likewise, while 
\cite{nikankin2025same,venhoff2025visualrepresentationsmaplanguage,zhang2025cross, serra2025the} characterize cross-modal information flow only through attention ablation or circuit-level mechanisms within a limited set of architectures, we show that semantic abstraction and localization of cross-modal communication are fundamentally linked during visual instruction tuning across seven distinct architectures, rather than being artifacts of any specific model.
By combining intervention and probing experiments, geometric alignment analyses, and localized fine-tuning interventions, our work reconciles these views and provides causal evidence for where multimodal integration occurs.

\paragraph{Limitations.}

\looseness=-1Although we evaluated a broad set of models representative of the current state of the art, our analysis is inherently limited to fully open architectures. Specifically, our approach necessitates access to both the pre-trained and fine-tuned checkpoints and, for the localized fine-tuning, the original training code and data. In addition, while our choice of trainable layers is grounded in extensive empirical evidence (see \Cref{sec:results_representations} and \Cref{fig:visual_reliance_correlation}), our primary objective was mechanistic characterization rather than performance maximization. As such, our configuration may not represent the global optimum, and specific applications might benefit from a more exhaustive hyperparameter search.

\paragraph{Future work.} Our present study focuses on vision–language models, and a natural next step is to examine whether these alignment mechanisms persist across other modalities, such as audio or video, when integrated into a pre-trained language backbone. It would also be informative to compare our findings with alternative multimodal integration strategies, such as native end-to-end training on multimodal data (e.g., \cite{team2024chameleon, wang2024emu3nexttokenpredictionneed}). Such comparisons would help disentangle which aspects of the alignment of abstractions in intermediate layers are intrinsic to language-centric backbones and which depend on specific architectural or training choices.
\section*{Contributions}
The paper originates as the Master’s thesis work of Luis Palacios, hosted at the Laboratory of Data Engineering (LADE) of Area Science Park, Trieste, Italy. Diego Doimo and Alberto Cazzaniga jointly supervised the project, providing technical guidance and research direction, with Lorenzo Basile as co-supervisor.
The core experiments were designed by Lorenzo Basile, Diego Doimo, and Alberto Cazzaniga. Following this design, Luis Palacios, Lorenzo Basile, and Diego Doimo collaboratively developed the codebase and executed the experimental pipeline. All authors contributed to the data analysis and the writing of this paper.

\section*{Acknowledgments}
\looseness=-1The authors acknowledge the Area Science Park supercomputing platform ORFEO made available for conducting the research reported in this paper, and the technical support of the LADE staff.
LB was supported by the European Union – NextGenerationEU within the PNRR project ``Finanziamento di progetti presentati da giovani ricercatori" – Mission 4 Component 2 Investment 1.2, CUP: J93C25000440001. DD and AC were supported by the project ``Supporto alla diagnosi di malattie rare tramite l’intelligenza artificiale" CUP: F53C22001770002, and ``Valutazione automatica delle immagini diagnostiche tramite l’intelligenza artificiale", CUP: F53C22001780002. AC was supported by the European Union – NextGenerationEU within the PNRR project ``PRP@CERIC" IR0000028 – Mission 4 Component 2 Investment 3.1 Action 3.1.1 and by the PRIN-PNRR project ``SCOLORINA", Mission 4, Component 2, Investment 1.1, and Call for tender no. 1409 published on 14 September 2022 by the Italian Ministry of University and Research (MUR), funded by the European Union – NextGenerationEU – CUP J53D23015070001.
{
    \small
    \bibliographystyle{unsrt}
    \bibliography{references}

@inproceedings{ortu-etal-2026-seeing,
    title = "When Seeing Overrides Knowing: Disentangling Knowledge Conflicts in Vision-Language Models",
    author = "Ortu, Francesco  and
      Jin, Zhijing  and
      Doimo, Diego  and
      Cazzaniga, Alberto",
    editor = "Liakata, Maria  and
      Moreira, Viviane P.  and
      Zhang, Jiajun  and
      Jurgens, David",
    booktitle = "Proceedings of the 64th Annual Meeting of the {A}ssociation for {C}omputational {L}inguistics (Volume 1: Long Papers)",
    month = jul,
    year = "2026",
    address = "San Diego, California, United States",
    publisher = "Association for Computational Linguistics",
    url = "https://aclanthology.org/2026.acl-long.642/",
    doi = "10.18653/v1/2026.acl-long.642",
    pages = "14109--14130",
    ISBN = "979-8-89176-390-6"
}

@article{doimo2024representation,
  title={The representation landscape of few-shot learning and fine-tuning in large language models},
  author={Doimo, Diego and Serra, Alessandro and Ansuini, Alessio and Cazzaniga, Alberto},
  journal={Advances in Neural Information Processing Systems},
  volume={37},
  pages={18122--18165},
  year={2024}
}

@inproceedings{
basile2025head,
title={Head Pursuit: Probing Attention Specialization in Multimodal Transformers},
author={Lorenzo Basile and Valentino Maiorca and Diego Doimo and Francesco Locatello and Alberto Cazzaniga},
booktitle={The Thirty-ninth Annual Conference on Neural Information Processing Systems},
year={2025},
url={https://openreview.net/forum?id=WQ9rnkaUWm}
}

@inproceedings{
camboulin2024understanding,
title={Understanding Variational Autoencoders with Intrinsic Dimension and Information Imbalance},
author={Charles Camboulin and Diego Doimo and Aldo Glielmo},
booktitle={UniReps: 2nd Edition of the Workshop on Unifying Representations in Neural Models},
year={2024},
url={https://openreview.net/forum?id=uViw1lr2mZ}
}

@misc{bergman2020deepnearestneighboranomaly,
      title={Deep Nearest Neighbor Anomaly Detection}, 
      author={Liron Bergman and Niv Cohen and Yedid Hoshen},
      year={2020},
      eprint={2002.10445},
      archivePrefix={arXiv},
      primaryClass={cs.LG},
      url={https://arxiv.org/abs/2002.10445}, 
}

@article{Boleda:2020,
  author = 	 {Gemma Boleda},
  title = 	 {Distributional Semantics and Linguistic Theory},
  journal = 	 {Annual Review of Linguistics},
  year = 	 {2020},
  pages = {213--234},
  volume = 6,
}

@inproceedings{glocal2023,
 author = {Muttenthaler, Lukas and Linhardt, Lorenz and Dippel, Jonas and Vandermeulen, Robert A and Hermann, Katherine and Lampinen, Andrew and Kornblith, Simon},
 booktitle = {Advances in Neural Information Processing Systems},
 editor = {A. Oh and T. Naumann and A. Globerson and K. Saenko and M. Hardt and S. Levine},
 pages = {50978--51007},
 publisher = {Curran Associates, Inc.},
 title = {Improving neural network representations using human similarity judgments},
 url = {https://proceedings.neurips.cc/paper_files/paper/2023/file/9febda1c8344cc5f2d51713964864e93-Paper-Conference.pdf},
 volume = {36},
 year = {2023}
}

@inproceedings{vangansbeke2020scan,
  title={Scan: Learning to classify images without labels},
  author={Van Gansbeke, Wouter and Vandenhende, Simon and Georgoulis, Stamatios and Proesmans, Marc and Van Gool, Luc},
  booktitle={Proceedings of the European Conference on Computer Vision},
  year={2020}
}

@inproceedings{mind_the_gap,
 author = {Liang, Victor Weixin and Zhang, Yuhui and Kwon, Yongchan and Yeung, Serena and Zou, James},
 booktitle = {Advances in Neural Information Processing Systems},
 editor = {S. Koyejo and S. Mohamed and A. Agarwal and D. Belgrave and K. Cho and A. Oh},
 pages = {17612--17625},
 publisher = {Curran Associates, Inc.},
 title = {Mind the Gap: Understanding the Modality Gap in Multi-modal Contrastive Representation Learning},
 url = {https://proceedings.neurips.cc/paper_files/paper/2022/file/702f4db7543a7432431df588d57bc7c9-Paper-Conference.pdf},
 volume = {35},
 year = {2022}
}

@inproceedings{xie2025region,
  title={Region-based cluster discrimination for visual representation learning},
  author={Xie, Yin and Yang, Kaicheng and An, Xiang and Wu, Kun and Zhao, Yongle and Deng, Weimo and Ran, Zimin and Wang, Yumeng and Feng, Ziyong and Miles, Roy and others},
  booktitle={Proceedings of the IEEE/CVF International Conference on Computer Vision},
  pages={1793--1803},
  year={2025}
}

@misc{bai2025qwen3vltechnicalreport,
      title={Qwen3-VL Technical Report}, 
      author={Shuai Bai and Yuxuan Cai and Ruizhe Chen and Keqin Chen and Xionghui Chen and Zesen Cheng and Lianghao Deng and Wei Ding and Chang Gao and Chunjiang Ge and Wenbin Ge and Zhifang Guo and Qidong Huang and Jie Huang and Fei Huang and Binyuan Hui and Shutong Jiang and Zhaohai Li and Mingsheng Li and Mei Li and Kaixin Li and Zicheng Lin and Junyang Lin and Xuejing Liu and Jiawei Liu and Chenglong Liu and Yang Liu and Dayiheng Liu and Shixuan Liu and Dunjie Lu and Ruilin Luo and Chenxu Lv and Rui Men and Lingchen Meng and Xuancheng Ren and Xingzhang Ren and Sibo Song and Yuchong Sun and Jun Tang and Jianhong Tu and Jianqiang Wan and Peng Wang and Pengfei Wang and Qiuyue Wang and Yuxuan Wang and Tianbao Xie and Yiheng Xu and Haiyang Xu and Jin Xu and Zhibo Yang and Mingkun Yang and Jianxin Yang and An Yang and Bowen Yu and Fei Zhang and Hang Zhang and Xi Zhang and Bo Zheng and Humen Zhong and Jingren Zhou and Fan Zhou and Jing Zhou and Yuanzhi Zhu and Ke Zhu},
      year={2025},
      eprint={2511.21631},
      archivePrefix={arXiv},
      primaryClass={cs.CV},
      url={https://arxiv.org/abs/2511.21631}, 
}

@misc{wybitul2026representationstextimagesalign,
      title={Representations of Text and Images Align From Layer One}, 
      author={Evžen Wybitul and Javier Rando and Florian Tramèr and Stanislav Fort},
      year={2026},
      eprint={2601.08017},
      archivePrefix={arXiv},
      primaryClass={cs.CV},
      url={https://arxiv.org/abs/2601.08017}, 
}

@inproceedings{geva2023dissecting,
  title={Dissecting recall of factual associations in auto-regressive language models},
  author={Geva, Mor and Bastings, Jasmijn and Filippova, Katja and Globerson, Amir},
  booktitle={Proceedings of the 2023 Conference on Empirical Methods in Natural Language Processing},
  pages={12216--12235},
  year={2023}
}

@inproceedings{harada-etal-2025-massive,
    title = "Massive Supervised Fine-tuning Experiments Reveal How Data, Layer, and Training Factors Shape {LLM} Alignment Quality",
    author = "Harada, Yuto  and
      Yamauchi, Yusuke  and
      Oda, Yusuke  and
      Oseki, Yohei  and
      Miyao, Yusuke  and
      Takagi, Yu",
    editor = "Christodoulopoulos, Christos  and
      Chakraborty, Tanmoy  and
      Rose, Carolyn  and
      Peng, Violet",
    booktitle = "Proceedings of the 2025 Conference on Empirical Methods in Natural Language Processing",
    month = nov,
    year = "2025",
    address = "Suzhou, China",
    publisher = "Association for Computational Linguistics",
    url = "https://aclanthology.org/2025.emnlp-main.1138/",
    doi = "10.18653/v1/2025.emnlp-main.1138",
    pages = "22360--22381",
    ISBN = "979-8-89176-332-6"
}

@inproceedings{liu-niehues-2025-middle,
    title = "Middle-Layer Representation Alignment for Cross-Lingual Transfer in Fine-Tuned {LLM}s",
    author = "Liu, Danni  and
      Niehues, Jan",
    editor = "Che, Wanxiang  and
      Nabende, Joyce  and
      Shutova, Ekaterina  and
      Pilehvar, Mohammad Taher",
    booktitle = "Proceedings of the 63rd Annual Meeting of the Association for Computational Linguistics (Volume 1: Long Papers)",
    month = jul,
    year = "2025",
    address = "Vienna, Austria",
    publisher = "Association for Computational Linguistics",
    url = "https://aclanthology.org/2025.acl-long.778/",
    doi = "10.18653/v1/2025.acl-long.778",
    pages = "15979--15996",
    ISBN = "979-8-89176-251-0"
}

@inproceedings{conneau-etal-2018-cram,
    title = "What you can cram into a single {\$}{\&}!{\#}* vector: Probing sentence embeddings for linguistic properties",
    author = {Conneau, Alexis  and
      Kruszewski, German  and
      Lample, Guillaume  and
      Barrault, Lo{\"i}c  and
      Baroni, Marco},
    editor = "Gurevych, Iryna  and
      Miyao, Yusuke",
    booktitle = "Proceedings of the 56th Annual Meeting of the Association for Computational Linguistics (Volume 1: Long Papers)",
    month = jul,
    year = "2018",
    address = "Melbourne, Australia",
    publisher = "Association for Computational Linguistics",
    url = "https://aclanthology.org/P18-1198/",
    doi = "10.18653/v1/P18-1198",
    pages = "2126--2136"
}

@article{elhage2022solu,
   title={Softmax Linear Units},
   author={Elhage, Nelson and Hume, Tristan and Olsson, Catherine and Nanda, Neel and Henighan, Tom and Johnston, Scott and ElShowk, Sheer and Joseph, Nicholas and DasSarma, Nova and Mann, Ben and Hernandez, Danny and Askell, Amanda and Ndousse, Kamal and Jones, Andy and Drain, Dawn and Chen, Anna and Bai, Yuntao and Ganguli, Deep and Lovitt, Liane and Hatfield-Dodds, Zac and Kernion, Jackson and Conerly, Tom and Kravec, Shauna and Fort, Stanislav and Kadavath, Saurav and Jacobson, Josh and Tran-Johnson, Eli and Kaplan, Jared and Clark, Jack and Brown, Tom and McCandlish, Sam and Amodei, Dario and Olah, Christopher},
   year={2022},
   journal={Transformer Circuits Thread},
   note={https://transformer-circuits.pub/2022/solu/index.html}
}

@misc{liu2024llavanext,
    title={LLaVA-NeXT: Improved reasoning, OCR, and world knowledge},
    url={https://llava-vl.github.io/blog/2024-01-30-llava-next/},
    author={Liu, Haotian and Li, Chunyuan and Li, Yuheng and Li, Bo and Zhang, Yuanhan and Shen, Sheng and Lee, Yong Jae},
    month={January},
    year={2024}
}

@article{Dubey2024TheL3,
  author  = {Dubey, Abhimanyu and others},
  title   = {The Llama 3 Herd of Models},
  journal = {arXiv preprint arXiv:2407.21783},
  year    = {2024},
  url     = {https://arxiv.org/abs/2407.21783}
}

@article{meta2024llama3.2,
  title        = {Llama 3.2: Revolutionizing edge AI and vision with open, customizable models},
  author       = {{Meta AI}},
  year         = {2024},
  month        = {Sep 25},
  journal      = {Meta AI Blog},
}

@misc{meta2024llama32visioncard,
  title        = {Llama 3.2 Vision Model Card},
  author       = {{Meta AI}},
  year         = {2024},
  howpublished = {\url{https://github.com/meta-llama/llama-models/blob/main/models/llama3_2/MODEL_CARD_VISION.md}},
  note         = {Official model card},
}

@misc{opengvlab2024internvl2,
  title        = {InternVL2: Better than the Best---Expanding Performance Boundaries of Open-Source Multimodal Models with the Progressive Scaling Strategy},
  author       = {{OpenGVLab Team}},
  year         = {2024},
  month        = {Jul},
  url          = {https://internvl.github.io/blog/2024-07-02-InternVL-2.0/},
}

@article{an2025llavaonevision15,
  title        = {LLaVA-OneVision-1.5: Fully Open Framework for Democratized Multimodal Training},
  author       = {An, Xiang and Xie, Yin and Yang, Kaicheng and Zhang, Wenkang and Zhao, Xiuwei and Cheng, Zheng and Wang, Yirui and Xu, Songcen and Chen, Changrui and Zhu, Didi and others},
  journal      = {arXiv preprint arXiv:2509.23661},
  year         = {2025},
  url          = {https://arxiv.org/abs/2509.23661},
}

@misc{wang2024emu3nexttokenpredictionneed,
      title={Emu3: Next-Token Prediction is All You Need}, 
      author={Xinlong Wang and Xiaosong Zhang and Zhengxiong Luo and Quan Sun and Yufeng Cui and Jinsheng Wang and Fan Zhang and Yueze Wang and Zhen Li and Qiying Yu and Yingli Zhao and Yulong Ao and Xuebin Min and Tao Li and Boya Wu and Bo Zhao and Bowen Zhang and Liangdong Wang and Guang Liu and Zheqi He and Xi Yang and Jingjing Liu and Yonghua Lin and Tiejun Huang and Zhongyuan Wang},
      year={2024},
      eprint={2409.18869},
      archivePrefix={arXiv},
      primaryClass={cs.CV},
      url={https://arxiv.org/abs/2409.18869}, 
}

@article{team2024chameleon,
  title={Chameleon: Mixed-modal early-fusion foundation models},
  author={Team, Chameleon},
  journal={arXiv preprint arXiv:2405.09818},
  year={2024}
}

@article{arc,
  title={Think you have Solved Question Answering? Try ARC, the AI2 Reasoning Challenge},
  author={Peter Clark and Isaac Cowhey and Oren Etzioni and Tushar Khot and Ashish Sabharwal and Carissa Schoenick and Oyvind Tafjord},
  journal={ArXiv},
  year={2018},
  volume={abs/1803.05457}
}

@article{mmlu,
  title={Measuring Massive Multitask Language Understanding},
  author={Dan Hendrycks and Collin Burns and Steven Basart and Andy Zou and Mantas Mazeika and Dawn Song and Jacob Steinhardt},
  journal={Proceedings of the International Conference on Learning Representations (ICLR)},
  year={2021}
}

@article{winogrande,
    title={WinoGrande: An Adversarial Winograd Schema Challenge at Scale},
    author={Sakaguchi, Keisuke and Bras, Ronan Le and Bhagavatula, Chandra and Choi, Yejin},
    journal={arXiv preprint arXiv:1907.10641},
    year={2019}
}

@inproceedings{hellaswag,
    title={HellaSwag: Can a Machine Really Finish Your Sentence?},
    author={Zellers, Rowan and Holtzman, Ari and Bisk, Yonatan and Farhadi, Ali and Choi, Yejin},
    booktitle ={Proceedings of the 57th Annual Meeting of the Association for Computational Linguistics},
    year={2019}
}

@inproceedings{commonsenseqa,
    title = "{C}ommonsense{QA}: A Question Answering Challenge Targeting Commonsense Knowledge",
    author = "Talmor, Alon  and
      Herzig, Jonathan  and
      Lourie, Nicholas  and
      Berant, Jonathan",
    booktitle = "Proceedings of the 2019 Conference of the North {A}merican Chapter of the Association for Computational Linguistics: Human Language Technologies, Volume 1 (Long and Short Papers)",
    month = jun,
    year = "2019",
    address = "Minneapolis, Minnesota",
    publisher = "Association for Computational Linguistics",
    url = "https://aclanthology.org/N19-1421",
    doi = "10.18653/v1/N19-1421",
    pages = "4149--4158",
    archivePrefix = "arXiv",
    eprint        = "1811.00937",
    primaryClass  = "cs",
}

@inproceedings{piqa,
    author = {Yonatan Bisk and Rowan Zellers and
            Ronan Le Bras and Jianfeng Gao
            and Yejin Choi},
    title = {PIQA: Reasoning about Physical Commonsense in
           Natural Language},
    booktitle = {Thirty-Fourth AAAI Conference on
               Artificial Intelligence},
    year = {2020},
}

@misc{pixtral12b,
      title={Pixtral 12B}, 
      author={Pixtral Team},
      year={2024},
      eprint={2410.07073},
      archivePrefix={arXiv},
      primaryClass={cs.CV},
      url={https://arxiv.org/abs/2410.07073}, 
}

@article{team2025gemma,
  title={Gemma 3 technical report},
  author={Team, Gemma and Kamath, Aishwarya and Ferret, Johan and Pathak, Shreya and Vieillard, Nino and Merhej, Ramona and Perrin, Sarah and Matejovicova, Tatiana and Ram{\'e}, Alexandre and Rivi{\`e}re, Morgane and others},
  journal={arXiv preprint arXiv:2503.19786},
  year={2025}
}

@InProceedings{pmlr-v235-huh24a,
  title = 	 {Position: The Platonic Representation Hypothesis},
  author =       {Huh, Minyoung and Cheung, Brian and Wang, Tongzhou and Isola, Phillip},
  booktitle = 	 {Proceedings of the 41st International Conference on Machine Learning},
  pages = 	 {20617--20642},
  year = 	 {2024},
  editor = 	 {Salakhutdinov, Ruslan and Kolter, Zico and Heller, Katherine and Weller, Adrian and Oliver, Nuria and Scarlett, Jonathan and Berkenkamp, Felix},
  volume = 	 {235},
  series = 	 {Proceedings of Machine Learning Research},
  month = 	 {21--27 Jul},
  publisher =    {PMLR},
  url = 	 {https://proceedings.mlr.press/v235/huh24a.html}
}

@misc{acevedo2025quantitativeanalysissemanticinformation,
      title={A quantitative analysis of semantic information in deep representations of text and images}, 
      author={Santiago Acevedo and Andrea Mascaretti and Riccardo Rende and Matéo Mahaut and Marco Baroni and Alessandro Laio},
      year={2025},
      eprint={2505.17101},
      archivePrefix={arXiv},
      primaryClass={cs.CL},
      url={https://arxiv.org/abs/2505.17101}, 
}

@inproceedings{
wu2025the,
title={The Semantic Hub Hypothesis: Language Models Share Semantic Representations Across Languages and Modalities},
author={Zhaofeng Wu and Xinyan Velocity Yu and Dani Yogatama and Jiasen Lu and Yoon Kim},
booktitle={The Thirteenth International Conference on Learning Representations},
year={2025},
url={https://openreview.net/forum?id=FrFQpAgnGE}
}

@misc{venhoff2025visualrepresentationsmaplanguage,
      title={How Visual Representations Map to Language Feature Space in Multimodal LLMs}, 
      author={Constantin Venhoff and Ashkan Khakzar and Sonia Joseph and Philip Torr and Neel Nanda},
      year={2025},
      eprint={2506.11976},
      archivePrefix={arXiv},
      primaryClass={cs.CV},
      url={https://arxiv.org/abs/2506.11976}, 
}

@inproceedings{zhang2025cross,
  title={Cross-modal information flow in multimodal large language models},
  author={Zhang, Zhi and Yadav, Srishti and Han, Fengze and Shutova, Ekaterina},
  booktitle={Proceedings of the Computer Vision and Pattern Recognition Conference},
  pages={19781--19791},
  year={2025}
}

@article{glielmo2022ranking,
  title={Ranking the information content of distance measures},
  author={Glielmo, Aldo and Zeni, Claudio and Cheng, Bingqing and Cs{\'a}nyi, G{\'a}bor and Laio, Alessandro},
  journal={PNAS nexus},
  volume={1},
  number={2},
  pages={pgac039},
  year={2022},
  publisher={Oxford University Press}
}

@inproceedings{skeanlayer,
  title={Layer by Layer: Uncovering Hidden Representations in Language Models},
  author={Skean, Oscar and Arefin, Md Rifat and Zhao, Dan and Patel, Niket Nikul and Naghiyev, Jalal and LeCun, Yann and Shwartz-Ziv, Ravid},
  booktitle={Forty-second International Conference on Machine Learning},
  year={2025}
}

@inproceedings{chengemergence,
  title={Emergence of a High-Dimensional Abstraction Phase in Language Transformers},
  author={Cheng, Emily and Doimo, Diego and Kervadec, Corentin and Macocco, Iuri and Yu, Lei and Laio, Alessandro and Baroni, Marco},
  booktitle={The Thirteenth International Conference on Learning Representations},
  year={2025}
}

@inproceedings{valeriani2023,
 author = {Valeriani, Lucrezia and Doimo, Diego and Cuturello, Francesca and Laio, Alessandro and Ansuini, Alessio and Cazzaniga, Alberto},
 booktitle = {Advances in Neural Information Processing Systems},
 editor = {A. Oh and T. Naumann and A. Globerson and K. Saenko and M. Hardt and S. Levine},
 pages = {51234--51252},
 publisher = {Curran Associates, Inc.},
 title = {The geometry of hidden representations of large transformer models},
 volume = {36},
 year = {2023}
}

@inproceedings{kornblith2019similarity,
  title={Similarity of neural network representations revisited},
  author={Kornblith, Simon and Norouzi, Mohammad and Lee, Honglak and Hinton, Geoffrey},
  booktitle={International Conference on Machine Learning},
  pages={3519--3529},
  year={2019},
  organization={PMLR}
}

@article{doimo2020hierarchical,
  title={Hierarchical nucleation in deep neural networks},
  author={Doimo, Diego and Glielmo, Aldo and Ansuini, Alessio and Laio, Alessandro},
  journal={Advances in Neural Information Processing Systems},
  volume={33},
  year={2020}
}

@inproceedings{li2024seed,
  title={Seed-bench: Benchmarking multimodal large language models},
  author={Li, Bohao and Ge, Yuying and Ge, Yixiao and Wang, Guangzhi and Wang, Rui and Zhang, Ruimao and Shan, Ying},
  booktitle={Proceedings of the IEEE/CVF Conference on Computer Vision and Pattern Recognition},
  pages={13299--13308},
  year={2024}
}

@inproceedings{liu2024mmbench,
  title={Mmbench: Is your multi-modal model an all-around player?},
  author={Liu, Yuan and Duan, Haodong and Zhang, Yuanhan and Li, Bo and Zhang, Songyang and Zhao, Wangbo and Yuan, Yike and Wang, Jiaqi and He, Conghui and Liu, Ziwei and others},
  booktitle={European conference on computer vision},
  pages={216--233},
  year={2024},
  organization={Springer}
}

@inproceedings{
fu2025mme,
title={{MME}: A Comprehensive Evaluation Benchmark for Multimodal Large Language Models},
author={Chaoyou Fu and Peixian Chen and Yunhang Shen and Yulei Qin and Mengdan Zhang and Xu Lin and Jinrui Yang and Xiawu Zheng and Ke Li and Xing Sun and Yunsheng Wu and Rongrong Ji and Caifeng Shan and Ran He},
booktitle={The Thirty-ninth Annual Conference on Neural Information Processing Systems Datasets and Benchmarks Track},
year={2025},
url={https://openreview.net/forum?id=DgH9YCsqWm}
}

@inproceedings{li2023evaluating,
  title={Evaluating Object Hallucination in Large Vision-Language Models},
  author={Li, Yifan and Du, Yifan and Zhou, Kun and Wang, Jinpeng and Zhao, Wayne Xin and Wen, Ji-Rong},
  booktitle={Proceedings of the 2023 Conference on Empirical Methods in Natural Language Processing},
  pages={292--305},
  year={2023}
}

@inproceedings{goyal2017making,
  title={Making the v in vqa matter: Elevating the role of image understanding in visual question answering},
  author={Goyal, Yash and Khot, Tejas and Summers-Stay, Douglas and Batra, Dhruv and Parikh, Devi},
  booktitle={Proceedings of the IEEE conference on computer vision and pattern recognition},
  pages={6904--6913},
  year={2017}
}

@inproceedings{hudson2019gqa,
  title={Gqa: A new dataset for real-world visual reasoning and compositional question answering},
  author={Hudson, Drew A and Manning, Christopher D},
  booktitle={Proceedings of the IEEE/CVF conference on computer vision and pattern recognition},
  pages={6700--6709},
  year={2019}
}

@article{lu2022learn,
  title={Learn to explain: Multimodal reasoning via thought chains for science question answering},
  author={Lu, Pan and Mishra, Swaroop and Xia, Tanglin and Qiu, Liang and Chang, Kai-Wei and Zhu, Song-Chun and Tafjord, Oyvind and Clark, Peter and Kalyan, Ashwin},
  journal={Advances in Neural Information Processing Systems},
  volume={35},
  pages={2507--2521},
  year={2022}
}

@article{ren2015exploring,
  title={Exploring models and data for image question answering},
  author={Ren, Mengye and Kiros, Ryan and Zemel, Richard},
  journal={Advances in neural information processing systems},
  volume={28},
  year={2015}
}

@inproceedings{lin2014microsoft,
  title={Microsoft coco: Common objects in context},
  author={Lin, Tsung-Yi and Maire, Michael and Belongie, Serge and Hays, James and Perona, Pietro and Ramanan, Deva and Doll{\'a}r, Piotr and Zitnick, C Lawrence},
  booktitle={European conference on computer vision},
  pages={740--755},
  year={2014},
  organization={Springer}
}

@article{young2014image,
  title={From image descriptions to visual denotations: New similarity metrics for semantic inference over event descriptions},
  author={Young, Peter and Lai, Alice and Hodosh, Micah and Hockenmaier, Julia},
  journal={Transactions of the association for computational linguistics},
  volume={2},
  pages={67--78},
  year={2014},
  publisher={MIT Press One Rogers Street, Cambridge, MA 02142-1209, USA journals-info~…}
}

@inproceedings{
serra2025the,
title={The Narrow Gate: Localized Image-Text Communication in Native Multimodal Models},
author={Alessandro Pietro Serra and Francesco Ortu and Emanuele Panizon and Lucrezia Valeriani and Lorenzo Basile and Alessio Ansuini and Diego Doimo and Alberto Cazzaniga},
booktitle={The Thirty-ninth Annual Conference on Neural Information Processing Systems},
year={2025},
url={https://openreview.net/forum?id=n0I0IvdfB3}
}

@inproceedings{
neo2025towards,
title={Towards Interpreting Visual Information Processing in Vision-Language Models},
author={Clement Neo and Luke Ong and Philip Torr and Mor Geva and David Krueger and Fazl Barez},
booktitle={The Thirteenth International Conference on Learning Representations},
year={2025},
url={https://openreview.net/forum?id=chanJGoa7f}
}

@inproceedings{
nikankin2025same,
title={Same Task, Different Circuits: Disentangling Modality-Specific Mechanisms in {VLM}s},
author={Yaniv Nikankin and Dana Arad and Yossi Gandelsman and Yonatan Belinkov},
booktitle={The Thirty-ninth Annual Conference on Neural Information Processing Systems},
year={2025},
url={https://openreview.net/forum?id=AD8ksC9bw1}
}

@inproceedings{liu2024improved,
  title={Improved baselines with visual instruction tuning},
  author={Liu, Haotian and Li, Chunyuan and Li, Yuheng and Lee, Yong Jae},
  booktitle={Proceedings of the IEEE/CVF conference on computer vision and pattern recognition},
  pages={26296--26306},
  year={2024}
}

@article{liu2023visual,
  title={Visual instruction tuning},
  author={Liu, Haotian and Li, Chunyuan and Wu, Qingyang and Lee, Yong Jae},
  journal={Advances in neural information processing systems},
  volume={36},
  pages={34892--34916},
  year={2023}
}

@article{tong2024cambrian,
  title={Cambrian-1: A fully open, vision-centric exploration of multimodal llms},
  author={Tong, Peter and Brown, Ellis and Wu, Penghao and Woo, Sanghyun and IYER, Adithya Jairam Vedagiri and Akula, Sai Charitha and Yang, Shusheng and Yang, Jihan and Middepogu, Manoj and Wang, Ziteng and others},
  journal={Advances in Neural Information Processing Systems},
  volume={37},
  pages={87310--87356},
  year={2024}
}

@article{nostalgebraist2020interpreting,
    title={interpreting GPT: the logit lens},
    author={nostalgebraist},
    year={2020},
    journal={LessWrong},
    url={https://www.lesswrong.com/posts/AcKRB8wDpdaN6v6ru/interpreting-gpt-the-logit-lens}
}

@article{alayrac2022flamingo,
  title={Flamingo: a visual language model for few-shot learning},
  author={Alayrac, Jean-Baptiste and Donahue, Jeff and Luc, Pauline and Miech, Antoine and Barr, Iain and Hasson, Yana and Lenc, Karel and Mensch, Arthur and Millican, Katherine and Reynolds, Malcolm and others},
  journal={Advances in neural information processing systems},
  volume={35},
  pages={23716--23736},
  year={2022}
}

@article{Qwen-VL,
  title={Qwen-VL: A Versatile Vision-Language Model for Understanding, Localization, Text Reading, and Beyond},
  author={Bai, Jinze and Bai, Shuai and Yang, Shusheng and Wang, Shijie and Tan, Sinan and Wang, Peng and Lin, Junyang and Zhou, Chang and Zhou, Jingren},
  journal={arXiv preprint arXiv:2308.12966},
  year={2023}
}

@inproceedings{radford2021learning,
  title={Learning transferable visual models from natural language supervision},
  author={Radford, Alec and Kim, Jong Wook and Hallacy, Chris and Ramesh, Aditya and Goh, Gabriel and Agarwal, Sandhini and Sastry, Girish and Askell, Amanda and Mishkin, Pamela and Clark, Jack and others},
  booktitle={International conference on machine learning},
  pages={8748--8763},
  year={2021},
  organization={PmLR}
}

@inproceedings{masry2022chartqa,
  title={Chartqa: A benchmark for question answering about charts with visual and logical reasoning},
  author={Masry, Ahmed and Do, Xuan Long and Tan, Jia Qing and Joty, Shafiq and Hoque, Enamul},
  booktitle={Findings of the association for computational linguistics: ACL 2022},
  pages={2263--2279},
  year={2022}
}

@inproceedings{yue2024mmmu,
  title={{MMMU}: A Massive Multi-discipline Multimodal Understanding and Reasoning Benchmark for Expert {AGI}},
  author={Yue, Xiang and Ni, Yuansheng and Zhang, Kai and Zheng, Tianyu and Liu, Ruoqi and Zhang, Ge and Stevens, Samuel and Jiang, Dongfu and Ren, Weiming and Sun, Yuxuan and Wei, Cong and Yu, Botao and Yuan, Ruibin and Sun, Renliang and Yin, Ming and Zheng, Boyuan and Yang, Zhenzhu and Liu, Yibo and Huang, Wenhao and Sun, Huan and Su, Yu and Chen, Wenhu},
  booktitle={Proceedings of the IEEE/CVF Conference on Computer Vision and Pattern Recognition},
  year={2024}
}

@inproceedings{yue2024mmmupro,
  title={{MMMU-Pro}: A More Robust Multi-discipline Multimodal Understanding Benchmark},
  author={Yue, Xiang and Zheng, Tianyu and Ni, Yuansheng and Wang, Yubo and Zhang, Kai and Tong, Shengbang and Sun, Yuxuan and Yu, Botao and Zhang, Ge and Sun, Huan and Su, Yu and Chen, Wenhu and Neubig, Graham},
  booktitle={Proceedings of the 63rd Annual Meeting of the Association for Computational Linguistics},
  year={2025},
  url={https://arxiv.org/abs/2409.02813}
}

@article{chen2024mmstar,
  title={Are We on the Right Way for Evaluating Large Vision-Language Models?},
  author={Chen, Lin and Li, Jinsong and Dong, Xiaoyi and Zhang, Pan and Zang, Yuhang and Chen, Zehui and Duan, Haodong and Wang, Jiaqi and Qiao, Yu and Lin, Dahua and Feng, Zhao},
  journal={arXiv preprint arXiv:2403.20330},
  year={2024}
}

@inproceedings{li2024naturalbench,
  title={{NaturalBench}: Evaluating Vision-Language Models on Natural Adversarial Samples},
  author={Li, Baiqi and Lin, Zhiqiu and Peng, Wenxuan and Nyandwi, Jean de Dieu and Jiang, Daniel and Ma, Zixian and Khanuja, Simran and Krishna, Ranjay and Neubig, Graham and Ramanan, Deva},
  booktitle={Advances in Neural Information Processing Systems},
  year={2024}
}

@misc{xai2024grok15v,
  title={Grok-1.5 Vision Preview},
  author={{xAI}},
  year={2024},
  howpublished={\url{https://x.ai/news/grok-1.5v}}
}

@inproceedings{paiss2023teaching,
  title={Teaching {CLIP} to Count to Ten},
  author={Paiss, Roni and Ephrat, Ariel and Tov, Omer and Zada, Shiran and Mosseri, Inbar and Irani, Michal and Dekel, Tali},
  booktitle={Proceedings of the IEEE/CVF International Conference on Computer Vision},
  pages={3147--3157},
  year={2023}
}

@inproceedings{yuksekgonul2023when,
  title={When and Why Vision-Language Models Behave like Bags-of-Words, and What to Do About It?},
  author={Yuksekgonul, Mert and Bianchi, Federico and Kalluri, Pratyusha and Jurafsky, Dan and Zou, James},
  booktitle={International Conference on Learning Representations},
  year={2023}
}

@inproceedings{hsieh2023sugarcrepe,
  title={{SugarCrepe}: Fixing Hackable Benchmarks for Vision-Language Compositionality},
  author={Hsieh, Cheng-Yu and Zhang, Jieyu and Ma, Zixian and Kembhavi, Aniruddha and Krishna, Ranjay},
  booktitle={Advances in Neural Information Processing Systems},
  year={2023}
}

@inproceedings{Ravi2016OptimizationAA,
  title={Optimization as a Model for Few-Shot Learning},
  author={Sachin Ravi and H. Larochelle},
  booktitle={International Conference on Learning Representations},
  year={2016},
  url={https://api.semanticscholar.org/CorpusID:67413369}
}

@article{Xiao2010SUNDL,
  title={SUN database: Large-scale scene recognition from abbey to zoo},
  author={Jianxiong Xiao and James Hays and Krista A. Ehinger and Aude Oliva and Antonio Torralba},
  journal={2010 IEEE Computer Society Conference on Computer Vision and Pattern Recognition},
  year={2010},
  pages={3485-3492},
  url={https://api.semanticscholar.org/CorpusID:1309931}
}

@article{Cimpoi2013DescribingTI,
  title={Describing Textures in the Wild},
  author={Mircea Cimpoi and Subhransu Maji and Iasonas Kokkinos and Sammy Mohamed and Andrea Vedaldi},
  journal={2014 IEEE Conference on Computer Vision and Pattern Recognition},
  year={2013},
  pages={3606-3613},
  url={https://api.semanticscholar.org/CorpusID:4309276}
}

@article{Zhou2018PlacesA1,
  title={Places: A 10 Million Image Database for Scene Recognition},
  author={Bolei Zhou and {\`A}gata Lapedriza and Aditya Khosla and Aude Oliva and Antonio Torralba},
  journal={IEEE Transactions on Pattern Analysis and Machine Intelligence},
  year={2018},
  volume={40},
  pages={1452-1464},
  url={https://api.semanticscholar.org/CorpusID:2608922}
}

@inproceedings{Bossard2014Food101M,
  title={Food-101 - Mining Discriminative Components with Random Forests},
  author={Lukas Bossard and Matthieu Guillaumin and Luc Van Gool},
  booktitle={European Conference on Computer Vision},
  year={2014},
  url={https://api.semanticscholar.org/CorpusID:12726540}
}

@article{FeiFei2004LearningGV,
  title={Learning Generative Visual Models from Few Training Examples: An Incremental Bayesian Approach Tested on 101 Object Categories},
  author={Li Fei-Fei and Rob Fergus and Pietro Perona},
  journal={2004 Conference on Computer Vision and Pattern Recognition Workshop},
  year={2004},
  pages={178-178},
  url={https://api.semanticscholar.org/CorpusID:2156851}
}

@article{Nilsback2008AutomatedFC,
  title={Automated Flower Classification over a Large Number of Classes},
  author={Maria-Elena Nilsback and Andrew Zisserman},
  journal={2008 Sixth Indian Conference on Computer Vision, Graphics \& Image Processing},
  year={2008},
  pages={722-729},
  url={https://api.semanticscholar.org/CorpusID:15193013}
}

@inproceedings{Zhang2024LMMsEvalRC,
  title={LMMs-Eval: Reality Check on the Evaluation of Large Multimodal Models},
  author={Kaichen Zhang and Bo Li and Peiyuan Zhang and Fanyi Pu and Joshua Adrian Cahyono and Kairui Hu and Yuhao Dong and Shuai Liu and Yuanhan Zhang and Jingkang Yang and Chunyuan Li and Ziwei Liu},
  booktitle={North American Chapter of the Association for Computational Linguistics},
  year={2024},
  url={https://api.semanticscholar.org/CorpusID:271244782}
}

@inproceedings{
krojer2026latentlens,
title={LatentLens: Revealing Highly Interpretable Visual Tokens in {LLM}s},
author={Benno Krojer and Shravan Nayak and Oscar Ma{\~n}as and Vaibhav Adlakha and Desmond Elliott and Siva Reddy and Marius Mosbach},
booktitle={Forty-third International Conference on Machine Learning},
year={2026},
url={https://openreview.net/forum?id=qJFatzEGQc}
}
}

\clearpage

\appendix

\section*{Appendix}

\section{Model architectures and training paradigms}
\label{app:architectures}
All the VLMs we study share the same coarse structure: a pre-trained vision encoder extracts visual features and a connector maps them into the embedding space of a pre-trained LLM. They also share the same training progression: a modality alignment phase that trains the connector on image--text pairs while the LLM is frozen, followed by a visual instruction tuning phase on instruction-following data. Within this template, the models differ in nearly every choice that could affect where visual information is integrated: the vision encoder, the connector mechanism, the language backbone and its scale, the amount of multimodal data, and which components are updated during instruction tuning. Analyzing all of them lets us verify that the localization of multimodal integration in the intermediate layers is not tied to one specific design.

\textbf{LLaVA-1.5} \cite{liu2023visual,liu2024improved} is the minimal instance of the template: a CLIP ViT-L/336 encoder, a two-layer MLP projector, and a Vicuna backbone (7B and 13B). Stage 1 trains the projector on 558K image--text pairs and Stage 2 trains the projector and the LLM on a 665K instruction mixture, with the vision encoder frozen throughout.

\textbf{InternVL2-8B} \cite{opengvlab2024internvl2} keeps the ViT--MLP--LLM interface but scales the vision side with an InternViT encoder and dynamic high-resolution tiling, and uses an InternLM chat backbone; its Stage 2 updates the encoder, the projector, and the LLM on a broad bilingual mixture that includes OCR, multi-image, and video data.

\textbf{LLaVA-OneVision-1.5} \cite{an2025llavaonevision15} (4B and 8B) replaces the visual tower with the region-aware RICE-ViT \cite{xie2025region} and the backbone with Qwen3, and inserts a full-parameter mid-training stage on 85M image--text pairs before full-parameter instruction tuning on 22M samples, placing it at the opposite end of the data-scale spectrum from LLaVA-1.5.

\textbf{Cambrian-8B} \cite{tong2024cambrian} is a vision-centric design in which several frozen vision encoders feed a Spatial Vision Aggregator, a connector based on learnable latent queries and cross-attention that compresses the visual features before a LLaMA-style backbone; Stage 1 trains the aggregator on 2.5M alignment pairs and Stage 2 trains the aggregator and the LLM on the Cambrian-7M mixture.

\textbf{Llama-3.2-Vision-90B} \cite{Dubey2024TheL3,meta2024llama3.2,meta2024llama32visioncard} departs from the template in the integration mechanism: instead of inserting visual tokens into the prompt, cross-attention layers placed after every fourth self-attention block read the image-encoder features. Multimodal training never updates the language weights: vision pre-training trains the encoder and the adapter against a frozen Llama-3.1 backbone, and post-training swaps in the instruction-tuned chat backbone and again updates only the vision pathway. It tests whether our conclusions hold when visual information enters through cross-attention, the LLM is never changed, and the scale is one order of magnitude larger.

The exact checkpoints used in our analyses are listed in \Cref{tab:checkpoint_provenance}.

\begin{table}[H]
    \centering
    \fontsize{7pt}{8pt}\selectfont
    \resizebox{\textwidth}{!}{%
    \begin{tabular}{llll}
    \toprule
    Model & Base/pre-trained checkpoint & Instruction-tuned checkpoint \\ 
    \midrule
    LLaVA-1.5 &
    \makecell[l]{Can be assembled from pre-trained connector + LLM backbone:\\
    7B: \href{https://huggingface.co/liuhaotian/llava-v1.5-mlp2x-336px-pretrain-vicuna-7b-v1.5}{\texttt{liuhaotian/llava-v1.5-mlp2x-336px-pretrain-vicuna-7b-v1.5}}\\
    \quad + \href{https://huggingface.co/lmsys/vicuna-7b-v1.5}{\texttt{lmsys/vicuna-7b-v1.5}}\\
    13B: \href{https://huggingface.co/liuhaotian/llava-v1.5-mlp2x-336px-pretrain-vicuna-13b-v1.5}{\texttt{liuhaotian/llava-v1.5-mlp2x-336px-pretrain-vicuna-13b-v1.5}}\\
    \quad + \href{https://huggingface.co/lmsys/vicuna-13b-v1.5}{\texttt{lmsys/vicuna-13b-v1.5}}} &
    \makecell[l]{\href{https://huggingface.co/llava-hf/llava-1.5-7b-hf}{\texttt{llava-hf/llava-1.5-7b-hf}}\\
    \href{https://huggingface.co/llava-hf/llava-1.5-13b-hf}{\texttt{llava-hf/llava-1.5-13b-hf}}} &
    \\ 
    InternVL2 &
    \href{https://huggingface.co/OpenGVLab/InternVL2-Pretrain-Models/tree/main/InternVL2-8B-Pretrain}{\texttt{OpenGVLab/InternVL2-Pretrain-Models/InternVL2-8B-Pretrain}} &
    \href{https://huggingface.co/OpenGVLab/InternVL2-8B}{\texttt{OpenGVLab/InternVL2-8B}} &
    \\ 
    LLaVA-OneVision-1.5 &
    \makecell[l]{\href{https://huggingface.co/lmms-lab/LLaVA-OneVision-1.5-4B-Base}{\texttt{lmms-lab/LLaVA-OneVision-1.5-4B-Base}}\\
    \href{https://huggingface.co/lmms-lab/LLaVA-OneVision-1.5-8B-Base}{\texttt{lmms-lab/LLaVA-OneVision-1.5-8B-Base}}} &
    \makecell[l]{\href{https://huggingface.co/lmms-lab/LLaVA-OneVision-1.5-4B-Instruct}{\texttt{lmms-lab/LLaVA-OneVision-1.5-4B-Instruct}}\\
    \href{https://huggingface.co/lmms-lab/LLaVA-OneVision-1.5-8B-Instruct}{\texttt{lmms-lab/LLaVA-OneVision-1.5-8B-Instruct}}} &
    \\ 
    Cambrian-1 &
    \href{https://huggingface.co/nyu-visionx/cambrian-8b-pretrain}{\texttt{nyu-visionx/cambrian-8b-pretrain}} &
    \href{https://huggingface.co/nyu-visionx/cambrian-8b}{\texttt{nyu-visionx/cambrian-8b}} &
    \\ 
    Llama-3.2-Vision & -- & 
    \href{https://huggingface.co/meta-llama/Llama-3.2-90B-Vision-Instruct}{\texttt{meta-llama/Llama-3.2-90B-Vision-Instruct}} &
    \\ 
    \bottomrule
    \end{tabular}
    }
    \caption{\textbf{Checkpoint provenance.} Base/pre-trained and instruction-tuned checkpoints studied.}
    \label{tab:checkpoint_provenance}
\end{table}

\section{Datasets}
\label{app:datasets}

\begin{table}[H]
    \centering
    \fontsize{7pt}{8pt}\selectfont
    \begin{tabular}{>{\raggedright\arraybackslash}p{0.19\textwidth}>{\raggedright\arraybackslash}p{0.33\textwidth}>{\raggedright\arraybackslash}p{0.40\textwidth}}
    \toprule
    Evaluation role & Datasets & Source \\
    \midrule
    General multimodal multiple-choice questions &
    MMBench \cite{liu2024mmbench}, MME \cite{fu2025mme}, SEED-Bench \cite{li2024seed}, MMStar \cite{chen2024mmstar} &
    Hugging Face: \texttt{lmms-lab/MMBench}, \texttt{lmms-lab/MME}, \texttt{lmms-lab/SEED-Bench}, \texttt{Lin-Chen/MMStar} (lmms-eval task definitions) \\
    Expert and scientific reasoning &
    ScienceQA \cite{lu2022learn}, MMMU \cite{yue2024mmmu}, MMMU-Pro \cite{yue2024mmmupro} &
    Hugging Face: \texttt{lmms-lab/ScienceQA}, \texttt{lmms-lab/MMMU}, \texttt{MMMU/MMMU\_Pro} (lmms-eval) \\
    Grounding, spatial, and robustness probes &
    POPE \cite{li2023evaluating}, CV-Bench \cite{tong2024cambrian}, RealWorldQA \cite{xai2024grok15v}, NaturalBench \cite{li2024naturalbench}, CountBench \cite{paiss2023teaching} &
    Hugging Face: \texttt{lmms-lab/POPE}, \texttt{nyu-visionx/CV-Bench}, \texttt{lmms-lab/RealWorldQA}, \texttt{BaiqiL/NaturalBench-lmms-eval}, \texttt{vikhyatk/CountBenchQA} (lmms-eval) \\
    Open-ended VQA and chart QA &
    VQAv2 \cite{goyal2017making}, GQA \cite{hudson2019gqa}, COCO-QA \cite{ren2015exploring}, ChartQA \cite{masry2022chartqa} &
    Hugging Face: \texttt{lmms-lab/VQAv2}, \texttt{lmms-lab/GQA}, \texttt{lmms-lab/ChartQA} (lmms-eval); COCO-QA in the captioned prompts below \\
    Compositional image-text matching &
    ARO \cite{yuksekgonul2023when}, SugarCrepe \cite{hsieh2023sugarcrepe} &
    ARO: annotations and Visual Genome images from the \href{https://github.com/mertyg/vision-language-models-are-bows}{ARO repository}, COCO val2014 images (\href{https://cocodataset.org}{cocodataset.org}), Flickr30k images (\texttt{nlphuji/flickr30k}). SugarCrepe: \href{https://github.com/RAIVNLab/sugar-crepe}{SugarCrepe repository}, COCO val2017 images \\
    Captioning &
    COCO \cite{lin2014microsoft}, Flickr30k \cite{young2014image} &
    Hugging Face: \texttt{lmms-lab/COCO-Caption2017}, \texttt{lmms-lab/flickr30k} (lmms-eval) \\
    Text-only controls &
    ARC-Easy \cite{arc}, CommonsenseQA \cite{commonsenseqa}, HellaSwag \cite{hellaswag}, MMLU \cite{mmlu}, PIQA \cite{piqa}, WinoGrande \cite{winogrande} &
    Hugging Face: \texttt{allenai/ai2\_arc}, \texttt{tau/commonsense\_qa}, \texttt{Rowan/hellaswag}, \texttt{cais/mmlu}, \texttt{baber/piqa}, \texttt{allenai/winogrande} (lm-evaluation-harness task definitions) \\
    \midrule
    Image classification: object recognition &
    mini-ImageNet \cite{Ravi2016OptimizationAA} (100 classes), Caltech101 \cite{FeiFei2004LearningGV} (101 classes) &
    Hugging Face: \texttt{timm/mini-imagenet}, \texttt{flwrlabs/caltech101} \\
    Image classification: fine-grained recognition &
    Food101 \cite{Bossard2014Food101M} (101 classes), Flowers102 \cite{Nilsback2008AutomatedFC} (102 classes) &
    Hugging Face: \texttt{ethz/food101}, \texttt{dpdl-benchmark/oxford\_flowers102} \\
    Image classification: scene recognition &
    SUN397 \cite{Xiao2010SUNDL} (397 classes), Places365 \cite{Zhou2018PlacesA1} (365 classes) &
    Hugging Face: \texttt{tanganke/sun397}, \texttt{torch-uncertainty/Places365} \\
    Image classification: texture recognition &
    DTD \cite{Cimpoi2013DescribingTI} (47 classes) &
    Hugging Face: \texttt{tanganke/dtd} \\
    Captioned VQA prompts (interventions, probing, geometry) &
    VQAv2, COCO-QA, GQA, SEED-Bench, MMBench, MME with human or Qwen3-VL captions &
    Hugging Face: \texttt{l11p/VQAv2-COCO-restval-ICQ}, \texttt{l11p/COCOQA-restval-ICQA}, \texttt{l11p/GQA-val-balanced-noyesno-captioned}, \texttt{l11p/SEED-Bench-images-captioned}, \texttt{l11p/MMBench\_EN-dev-captioned}, \texttt{l11p/MME-test-captioned} \\
    Image--caption retrieval &
    COCO detailed captions &
    \texttt{liuhaotian/LLaVA-Instruct-150K} (\texttt{detail\_23k}); COCO train2014 images \\
    Linguistic probing &
    CoordInv, SOMO \cite{conneau-etal-2018-cram} &
    \href{https://github.com/facebookresearch/SentEval}{SentEval} probing tasks \\
    \bottomrule
    \end{tabular}
    \caption{\textbf{Datasets and their sources.} }
    \label{tab:dataset_taxonomy}
\end{table}

\begin{table}[H]
    \centering
    \fontsize{7pt}{8pt}\selectfont
    \begin{tabular}{>{\raggedright\arraybackslash}p{0.26\textwidth}>{\raggedright\arraybackslash}p{0.30\textwidth}>{\raggedright\arraybackslash}p{0.36\textwidth}}
    \toprule
    Experiment & Datasets & Instances and selection \\
    \midrule
    Multimodal interventions (\Cref{fig:interventions_average,fig:app:interventions_average,fig:app:mm_layer_skip_progressive,fig:app:mm_layer_skip_sliding,fig:app:mm_attn_knockout_progressive,fig:app:mm_attn_knockout_sliding}) &
    GQA, MMBench-EN, MMBench-CN, MME, SEED-Bench, VQAv2 & 2,000 per benchmark: the first 2,000 instances  \\
    Text-only interventions (\Cref{fig:app:text_layer_skip_progressive,fig:app:text_layer_skip_sliding}) &
    ARC-Easy, CommonsenseQA, HellaSwag, MMLU, PIQA, WinoGrande & full evaluation sets  \\
    Label overlap (\Cref{fig:probing} left, \Cref{fig:app:label_overlap_average,fig:app:label_overlap_by_dataset}) &
    the seven classification datasets of \Cref{tab:dataset_taxonomy} & 10 classes per dataset (see above), all their images: 500 to 2,500 per dataset  \\
    Logit lens and linear probes (\Cref{fig:probing} right, \Cref{fig:app:linear_probe_mc,fig:app:logit_lens_mc,fig:app:logit_lens_oe}) &
    ScienceQA, MMBench, MME, POPE; COCO-QA, CountBench, GQA, VQAv2 & 2,000 per benchmark: random subset with a fixed seed  \\
    COCO retrieval (\Cref{fig:app:retrieval_coco}) & LLaVA-Instruct detailed captions & full set  \\
    Linguistic probing (\Cref{fig:app:linguistic_probing}) & CoordInv, SOMO & 5,000 sentences per task: class-balanced random subsets with a fixed seed, two repetitions  \\
    Information imbalance planes (\Cref{fig:information_imbalance,fig:app:information_imbalance_llava,fig:app:information_imbalance_onevision,fig:app:information_imbalance_internvl_cambrian}) and \Cref{tab:imbalance_last_image_tokens} &
    VQAv2, COCO-QA, GQA, SEED-Bench, MMBench (captioned) & 3,000 per dataset: the first 3,000 prompts (one per image); all 1,132 for MMBench  \\
    Linear CKA and neighborhood overlap (\Cref{app:cka_overlap}) & same datasets & 3,000 per dataset: the first 3,000 prompts (one per image); all 1,132 for MMBench  \\
    Benchmark tables and localized fine-tuning (\Cref{tab:performance_ft,tab:performance,fig:app:radar_localized_ft,fig:visual_reliance_correlation}) &
    the 20 benchmarks of \Cref{tab:dataset_taxonomy} & up to 5,000 per benchmark: the first 5,000 instances; ARO: random 5,000 per sub-task with a fixed seed (Flickr30k-Order: all 4,885); SugarCrepe: full  \\
    Llama-3.2-Vision-90B interventions and visual reliance (\Cref{fig:app:llama90b_ablations,tab:performance_ft_llama90b,fig:app:llama90b_visual_reliance}) &
    as above & 5,000 per benchmark: the first 5,000 instances  \\
    Llama-3.2-Vision-90B information imbalance (\Cref{fig:app:llama90b_information_imbalance}) & VQAv2, GQA, MMBench (captioned) & full datasets \\ \\
    \bottomrule
    \end{tabular}
    \caption{\textbf{Number of instances per experiment.}}
    \label{tab:sample_sizes}
\end{table}

\paragraph{Evaluation benchmarks.} Our evaluation suite, introduced in \Cref{sec:methods}, covers multiple-choice and open-ended visual question answering, captioning, image classification, and compositional reasoning, together with text-only controls. \Cref{tab:dataset_taxonomy} groups the benchmarks by the capability they probe. The categories are descriptive, since several datasets mix perception, language understanding, and reasoning.

The general multiple-choice benchmarks (MMBench, MME, SEED-Bench) give broad coverage of perception and reasoning, and MMStar restricts the samples to those that require visual context. ScienceQA, MMMU, and MMMU-Pro shift the evaluation toward knowledge-heavy, college-level reasoning, with MMMU-Pro filtering out text-only answerable questions and adding a vision-only setting. The grounding and robustness probes isolate failure modes that aggregate scores can hide: object hallucination (POPE), 2D/3D and real-world spatial understanding (CV-Bench, RealWorldQA), blind answering on paired natural images (NaturalBench), and counting (CountBench). VQAv2, COCO-QA, GQA, and ChartQA cover open-ended answering on natural images (COCO-QA with automatically generated questions), compositional scenes, and charts, and COCO and Flickr30k cover free-form captioning. ARO and SugarCrepe test compositional image--text matching with hard negative captions that differ from the correct one in attributes, relations, or word order; since both were designed for contrastive encoders, we adapt them to a multiple-choice format in which the model selects the correct caption from a small candidate set. Finally, the text-only benchmarks measure the knowledge-based (ARC-Easy, MMLU) and commonsense (CommonsenseQA, HellaSwag, PIQA, WinoGrande) reasoning that visual instruction tuning is expected to preserve. 

All multimodal benchmarks except the compositional ones are evaluated with the \href{https://github.com/evolvinglmms-lab/lmms-eval}{\texttt{lmms-eval}} library \cite{Zhang2024LMMsEvalRC} (v0.7.1), and the text-only benchmarks with \href{https://github.com/EleutherAI/lm-evaluation-harness}{\texttt{lm-evaluation-harness}} (v0.4.9). \Cref{tab:dataset_taxonomy} gives the source of every dataset and \Cref{tab:sample_sizes} the number of instances used in each experiment.

\paragraph{Probing the semantics of hidden representations.}  The label overlap experiments probing the semantic content of intermediate representations use the seven image classification datasets in \Cref{tab:dataset_taxonomy}, chosen to span different kinds of visual category: everyday object identity (mini-ImageNet, Caltech101), fine-grained distinctions among visually similar dishes and flower species (Food101, Flowers102), scene categorization that relies on global context rather than on a single object (SUN397, Places365), and texture attributes without object or scene semantics (DTD). Each dataset is restricted to 10 classes, the 10 most populated ones for Caltech101 and Flowers102 and 10 classes drawn with a fixed seed for the others, keeping all their images: 500 (SUN397) to 2,500 (Food101) per dataset. Exact-match accuracy on these datasets is above chance for all models, always above 50\% and above 90\% for some model--dataset combinations.

\section{Visual instruction tuning hyperparameters}\label{sec:hyperparameters}
We run localized fine-tuning on models with fully reproducible training pipelines and with publicly available data, focusing on the following model variants: LLaVA-1.5-7B, LLaVA-OneVision-1.5-4B (\Cref{Sec:results_finetuning}). The training hyperparameters were selected according to the official implementations, and they are reported in \Cref{tab:finetune_hyperparams}. 
\begin{table}[H]
\centering
\begin{tabular}{lcc}
\toprule
Hyperparameter & LLaVA-1.5 (7B) & LLaVA-OneVision-1.5 (4B) \\
\midrule
Batch size & 128 & 224 \\
Learning rate & \texttt{2e-5} & \texttt{1e-5} \\
Min learning rate & 0 & \texttt{1e-6} \\
LR schedule & cosine decay & cosine decay \\
Warmup ratio & 0.03 & 0.002 \\
Weight decay & 0 & 0 \\
Training length & 1 epoch & 1 epoch \\
Optimizer & AdamW & Adam \\
Optimizer betas & (0.9, 0.999) & (0.9, 0.99) \\
\bottomrule
\end{tabular}
\caption{\textbf{Visual instruction tuning hyperparameters.} LLaVA-1.5-7B values follow Table~9 of \cite{liu2024improved}; LLaVA-OneVision-1.5-4B values follow the released instruction-tuning scripts \cite{an2025llavaonevision15}.}
\label{tab:finetune_hyperparams}
\end{table}

\section{Task performances}

\subsection{Multimodal benchmarks.}

\begin{table}[H]
\centering
\resizebox{\columnwidth}{!}{
\begin{tabular}{l l|cccccccc}
\toprule
Task & Metric & LLaVA-7B & LLaVA-13B & OneVision-4B-LN & OneVision-4B & OneVision-8B & InternVL2-8B & Cambrian-8B & Llama-90B \\
\midrule
\multicolumn{10}{l}{\textbf{Multiple-choice}} \\
ScienceQA        & Exact Match (\%)       & 67.8 & 71.2 & 81.6 & 98.5 & 98.8 & 97.1 & 80.3 & 91.4 \\
MMBench-EN       & GPT Score              & 64.5 & 67.9 & 79.8 & 83.7 & 85.8 & 81.7 & 75.8 & 81.4 \\
MMBench-CN       & GPT Score              & 52.7 & 56.6 & 77.7 & 61.1 & 77.2 & 79.4 & 64.4 & 77.1 \\
MME              & Normalized Score (\%)  & 62.3 & 62.4 & 74.5 & 77.3 & 79.0 & 78.7 & 68.4 & 65.7 \\
POPE             & F1 Score (\%)          & 81.6 & 82.3 & 86.9 & 87.4 & 87.3 & 85.7 & 86.7 & 0.0 \\
SEED             & Accuracy (\%)          & 64.9 & 68.2 & 77.0 & 76.4 & 77.3 & 76.2 & 74.0 & 78.0 \\
MMMU             & Accuracy (\%)          & 35.7 & 34.7 & 48.8 & 52.0 & 55.6 & 48.3 & 41.6 & 49.2 \\
MMMU-Pro         & Accuracy (\%)          & 18.1 & 20.6 & 30.4 & 34.3 & 36.6 & 32.2 & 24.0 & 32.3 \\
MMStar           & Mean Acc. (\%)         & 33.7 & 36.5 & 55.8 & 65.2 & 68.1 & 59.4 & 47.5 & 52.7 \\
NaturalBench     & Group Acc. (\%)        & 14.4 & 14.7 & 27.7 & 27.6 & 29.2 & 24.5 & 20.4 & 32.5 \\
CVBench          & Accuracy (\%)          & 54.1 & 54.9 & 71.0 & 76.8 & 79.3 & 71.3 & 8.5 & 69.9 \\
RealWorldQA      & Exact Match (\%)       & 54.8 & 56.6 & 64.1 & 68.0 & 68.5 & 64.3 & 64.2 & 66.0 \\
\midrule
\multicolumn{10}{l}{\textbf{Open-ended}} \\
GQA              & Exact Match (\%)       & 61.9 & 62.3 & 61.6 & 62.4 & 62.5 & 63.0 & 65.6 & 53.8 \\
VQAv2            & Exact Match (\%)       & 76.6 & 77.6 & 80.1 & 81.1 & 82.0 & 79.0 & 82.8 & 73.6 \\
CountBench       & Accuracy (\%)          & 45.6 & 48.3 & 75.6 & 83.7 & 85.3 & 58.7 & 14.9 & 78.0 \\
ChartQA          & Relaxed Acc. (\%)      & 16.6 & 18.8 & 79.3 & 87.4 & 86.3 & 82.6 & 73.4 & 25.7 \\
\midrule
\multicolumn{10}{l}{\textbf{Compositionality}} \\
ARO              & Mean Acc. (\%)         & 74.2 & 83.8 & 87.5 & 90.0 & 90.0 & 88.7 & 89.7 & 85.44 \\
SugarCrepe       & Mean Acc. (\%)         & 79.6 & 87.9 & 89.6 & 94.4 & 94.1 & 95.2 & 92.6 & 94.71 \\
\midrule
\multicolumn{10}{l}{\textbf{Captioning}} \\
COCO             & CIDEr ($\times 100$)   & 108.8 & 112.7 & 135.6 & 118.8 & 135.7 & 90.3 & 9.3 & 51.2 \\
Flickr30k        & CIDEr ($\times 100$)   & 83.3 & 86.7 & 84.7 & 87.0 & 88.4 & 74.7 & 14.8 & 70.8 \\
\midrule
Overall Mean     &                        & 57.6 & 60.2 & 73.5 & 75.7 & 78.4 & 71.6 & 54.9 & 61.5 \\
\bottomrule
\end{tabular}
}
\caption{\textbf{Multimodal task performance.} We report each model on multiple-choice, open-ended, compositional, and captioning benchmarks using the metric specified in the second column; each benchmark is evaluated on up to 5,000 samples (ARO: 5,000 per sub-task; SugarCrepe: the full dataset). The Overall Mean summarizes the available entries in this multimodal evaluation suite. OneVision4B-LN is the fully fine-tuned model on the LLaVA-NeXT dataset, which we use as a reference in \Cref{sec:localized_finetuning}, while OneVision4B is the original instruction tuned checkpoint. Note: the performance of Cambrian-8B on CVBench and CountBench is driven by wrong answer formatting. We exclude these entries from the computation of the Image Reliance metric in \Cref{sec:corr_vision_centric}.}
\label{tab:performance}
\end{table}

\begin{table}[H]
\centering
\resizebox{\columnwidth}{!}{
\begin{tabular}{l l|ccccccc}
\toprule
Task & Metric & LLaVA-7B & LLaVA-13B & OneVision-4B-LN & OneVision-4B & OneVision-8B & InternVL2-8B & Cambrian-8B \\
\midrule
\multicolumn{9}{l}{\textbf{Multiple-choice}} \\
ScienceQA        & Exact Match (\%)       & 64.3 & 66.1 & 71.3 & 74.0 & 73.3 & 74.4 & 69.2 \\
MMBench-EN       & GPT Score              & 19.5 & 22.2 & 27.6 & 24.2 & 23.4 & 21.0 & 22.7 \\
MMBench-CN       & GPT Score              & 15.4 & 19.0 & 23.4 & 16.8 & 16.4 & 18.7 & 14.8 \\
MME              & Normalized Score (\%)  & 32.7 & 30.0 & 30.1 & 31.2 & 33.9 & 40.9 & 33.0 \\
POPE             & F1 Score (\%)          & 10.5 & 13.0 & 49.8 & 13.9 & 19.4 & 17.2 & 0.0 \\
SEED             & Accuracy (\%)          & 36.4 & 37.6 & 40.0 & 39.4 & 40.3 & 40.5 & 39.4 \\
MMMU             & Accuracy (\%)          & 31.5 & 32.6 & 41.6 & 43.2 & 45.1 & 42.0 & 34.7 \\
MMMU-Pro         & Accuracy (\%)          & 14.6 & 16.6 & 22.7 & 23.0 & 25.9 & 21.1 & 17.1 \\
MMStar           & Mean Acc. (\%)         & 24.2 & 21.9 & 28.3 & 27.7 & 31.8 & 27.8 & 20.5 \\
NaturalBench     & Group Acc. (\%)        & 0.0 & 0.0 & 0.0 & 0.0 & 0.0 & 0.1 & 0.0 \\
CVBench          & Accuracy (\%)          & 42.4 & 42.8 & 48.3 & 43.5 & 50.6 & 52.3 & 33.7 \\
RealWorldQA      & Exact Match (\%)       & 42.9 & 42.7 & 46.4 & 42.7 & 43.5 & 35.3 & 41.2 \\
\midrule
\multicolumn{9}{l}{\textbf{Open-ended}} \\
GQA              & Exact Match (\%)       & 39.1 & 39.6 & 36.8 & 36.6 & 35.6 & 33.7 & 39.4 \\
VQAv2            & Exact Match (\%)       & 42.4 & 43.3 & 40.6 & 40.1 & 40.4 & 38.9 & 42.4 \\
CountBench       & Accuracy (\%)          & 2.4 & 0.2 & 9.8 & 10.4 & 9.4 & 7.9 & 0.2 \\
ChartQA          & Relaxed Acc. (\%)      & 11.3 & 12.4 & 12.4 & 12.6 & 13.8 & 12.0 & 11.2 \\
\midrule
\multicolumn{9}{l}{\textbf{Compositionality}} \\
ARO              & Mean Acc. (\%)         & 64.9 & 67.0 & 79.0 & 78.8 & 80.3 & 80.1 & 80.6 \\
SugarCrepe       & Mean Acc. (\%)         & 52.2 & 50.2 & 49.8 & 56.1 & 59.4 & 61.6 & 59.2 \\
\midrule
\multicolumn{9}{l}{\textbf{Captioning}} \\
COCO             & CIDEr ($\times 100$)   & 0.6 & 3.9 & 8.4 & 8.2 & 7.6 & 9.1 & 4.0 \\
Flickr30k        & CIDEr ($\times 100$)   & 0.9 & 3.0 & 5.2 & 5.1 & 6.1 & 6.2 & 3.8 \\
\midrule
Overall Mean     &                        & 27.4 & 28.2 & 33.6 & 31.4 & 32.8 & 32.0 & 28.4 \\
\bottomrule
\end{tabular}
}
\caption{\textbf{Multimodal benchmark performance without the image.} Same evaluation as \Cref{tab:performance} with the image removed from the prompt; these are the text-only scores used in \Cref{eq:ir}.}
\end{table}

\subsection{Multimodal performance after localized fine-tuning.}

\begin{figure}[H]
\centering
\includegraphics[width=\textwidth]{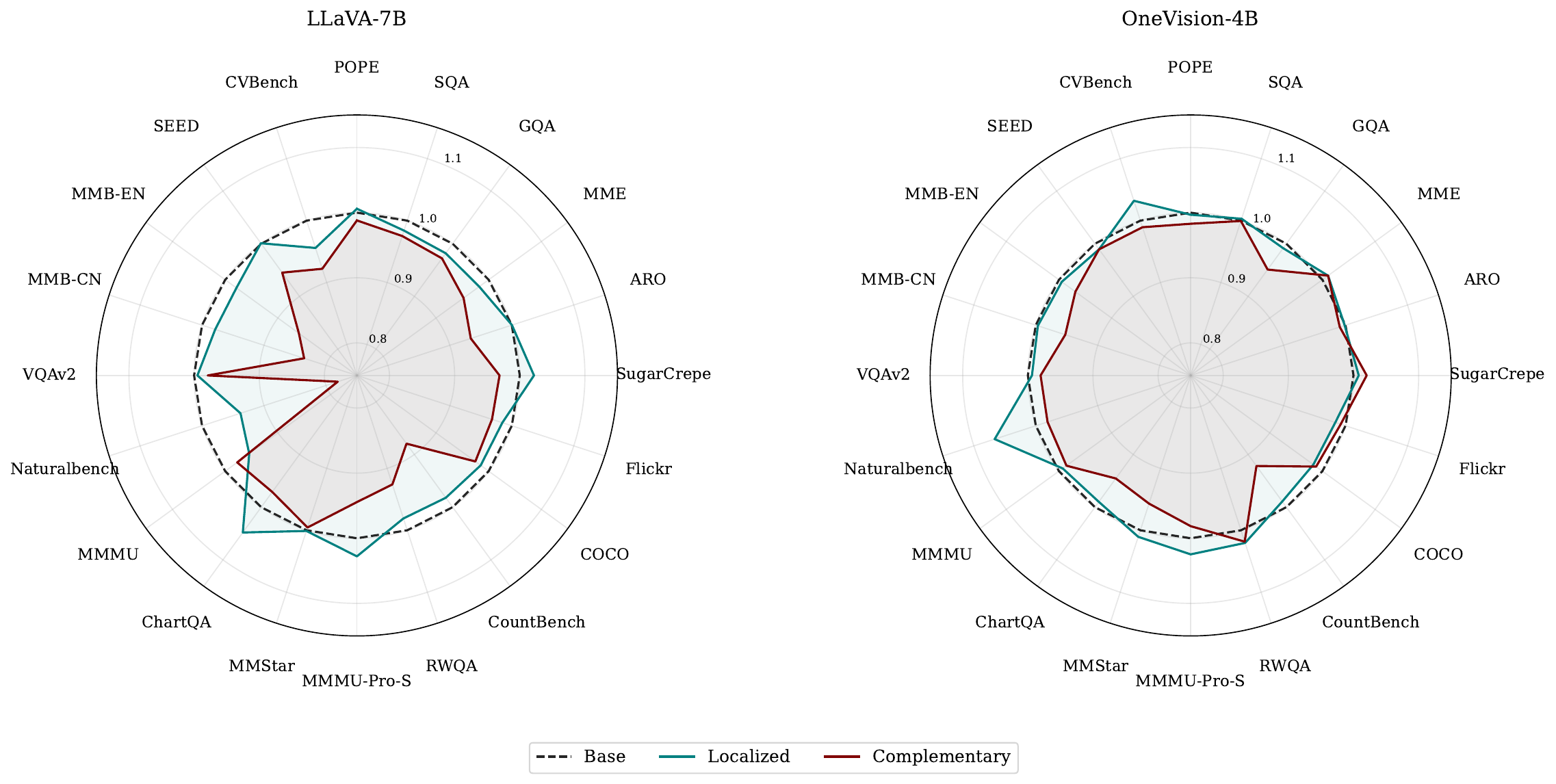}\caption{\textbf{Per-benchmark results of localized fine-tuning.} LLaVA-7B and OneVision-4B fine-tuned on the abstraction layers (localized) or on their complement, normalized to full fine-tuning. The compositional benchmarks (ARO and SugarCrepe) are reported only in \Cref{tab:performance_ft}.}\label{fig:app:radar_localized_ft}
\end{figure}

\subsection{Text benchmarks.}

\begin{table}[H]
\centering
\small
\resizebox{\columnwidth}{!}{
\begin{tabular}{ll|cccccc}
\toprule
Task & Metric & LLaVA-7B & LLaVA-13B & Cambrian-8B & InternVL2-8B & LLaVA-OV-1.5-4B & LLaVA-OV-1.5-8B \\
\midrule
ARC-Easy        & acc\_norm    & 60.40 & 63.38 & 69.78 & 67.13 & 73.11 & 77.82 \\
CommonsenseQA   & acc          & 60.52 & 52.50 & 77.89 & 83.13 & 80.59 & 83.37 \\
HellaSwag       & acc\_norm    & 67.58 & 71.17 & 68.42 & 75.38 & 67.26 & 72.14 \\
MMLU            & acc          & 41.63 & 46.92 & 55.91 & 71.11 & 72.65 & 75.40 \\
PIQA            & acc\_norm    & 76.71 & 77.86 & 76.93 & 81.01 & 77.31 & 78.94 \\
Winogrande      & acc          & 67.25 & 68.90 & 67.01 & 76.01 & 68.90 & 70.88 \\
\bottomrule
\end{tabular}
}
\caption{\textbf{Text-only benchmark performance.} Zero-shot, on the full evaluation sets.}
\label{tab:text_benchmarks}
\end{table}

\subsection{Text benchmarks after localized fine-tuning.}

\begin{table}[H]
\centering
\small
\begin{tabular}{ll|ccc|ccc}
\toprule
Task & Metric & \multicolumn{3}{c|}{LLaVA-7B} & \multicolumn{3}{c}{OneVision-4B} \\
 & Trained layers & all & $[6, 16]$ & $[6, 16]^C$ & all & $[8, 25]$ & $[8, 25]^C$ \\
\midrule
ARC-Easy        & acc\_norm    & 60.40 & 59.89 & 59.13 & 61.49 & 61.45 & 59.93 \\
CommonsenseQA   & acc          & 60.52 & 61.43 & 41.93 & 77.72 & 77.07 & 78.38 \\
HellaSwag       & acc\_norm    & 67.58 & 66.92 & 68.04 & 65.54 & 65.65 & 65.63 \\
MMLU            & acc          & 41.63 & 41.35 & 35.48 & 69.65 & 69.83 & 69.57 \\
PIQA            & acc\_norm    & 76.71 & 76.17 & 76.50 & 76.17 & 76.17 & 76.17 \\
Winogrande      & acc          & 67.25 & 65.90 & 65.67 & 65.11 & 65.35 & 66.46 \\
\midrule
Mean            &              & 62.35 & 61.94 & 57.79 & 69.28 & 69.25 & 69.36 \\
Relative to all layers (\%) &  & 100.0 & 99.4  & 91.7  & 100.0 & 100.0 & 100.1 \\
\bottomrule
\end{tabular}
\caption{\textbf{Text-only benchmark performance after localized fine-tuning.} Same evaluation as \Cref{tab:text_benchmarks} (zero-shot, full evaluation sets) for LLaVA-7B and OneVision-4B fine-tuned on all layers, on the abstraction layers, and on their complement, as in \Cref{tab:performance_ft}. For OneVision-4B, \emph{all} is OneVision-4B-LN (\Cref{tab:performance}). The last row averages, over the six tasks, the score divided by that of the model fine-tuned on all layers.}
\label{tab:text_benchmarks_ft}
\end{table}

\clearpage 

\section{Interventions: additional results}

\subsection{Interventions: dataset average for LLaVA-13B, LLaVA-OneVision-8B, Cambrian-8B}
\begin{figure}[H]
\centering
\includegraphics[width=\linewidth]{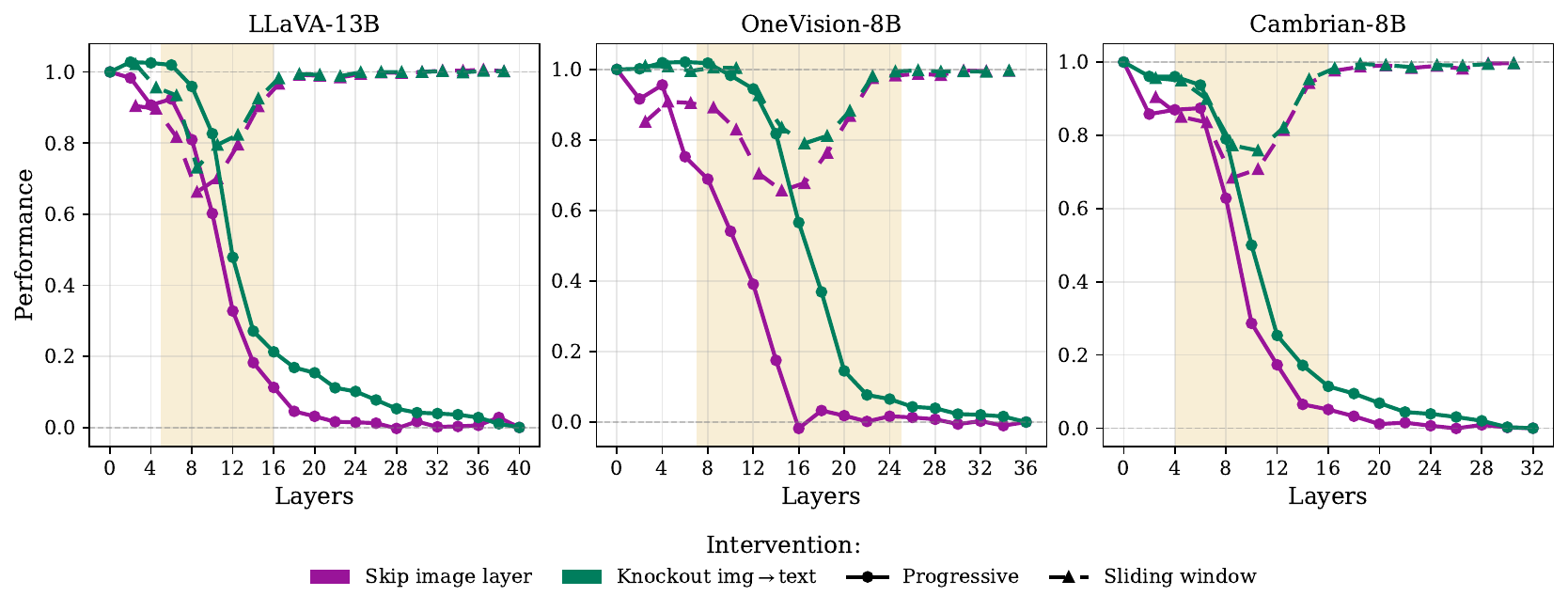}
\caption{\textbf{Interventions on LLaVA-13B, OneVision-8B, and Cambrian-8B.} Same setup as \Cref{fig:interventions_average}: layer skipping and cross-modal attention knockout, applied progressively from the input or on a sliding window of five layers.}
\label{fig:app:interventions_average}
\end{figure}

\subsection{Interventions: all datasets (multimodal benchmarks)}
\subsubsection*{Progressive layer-skipping}
\label{app:ablation_studies}
\begin{figure}[H]
\centering
\includegraphics[width=0.9\linewidth]{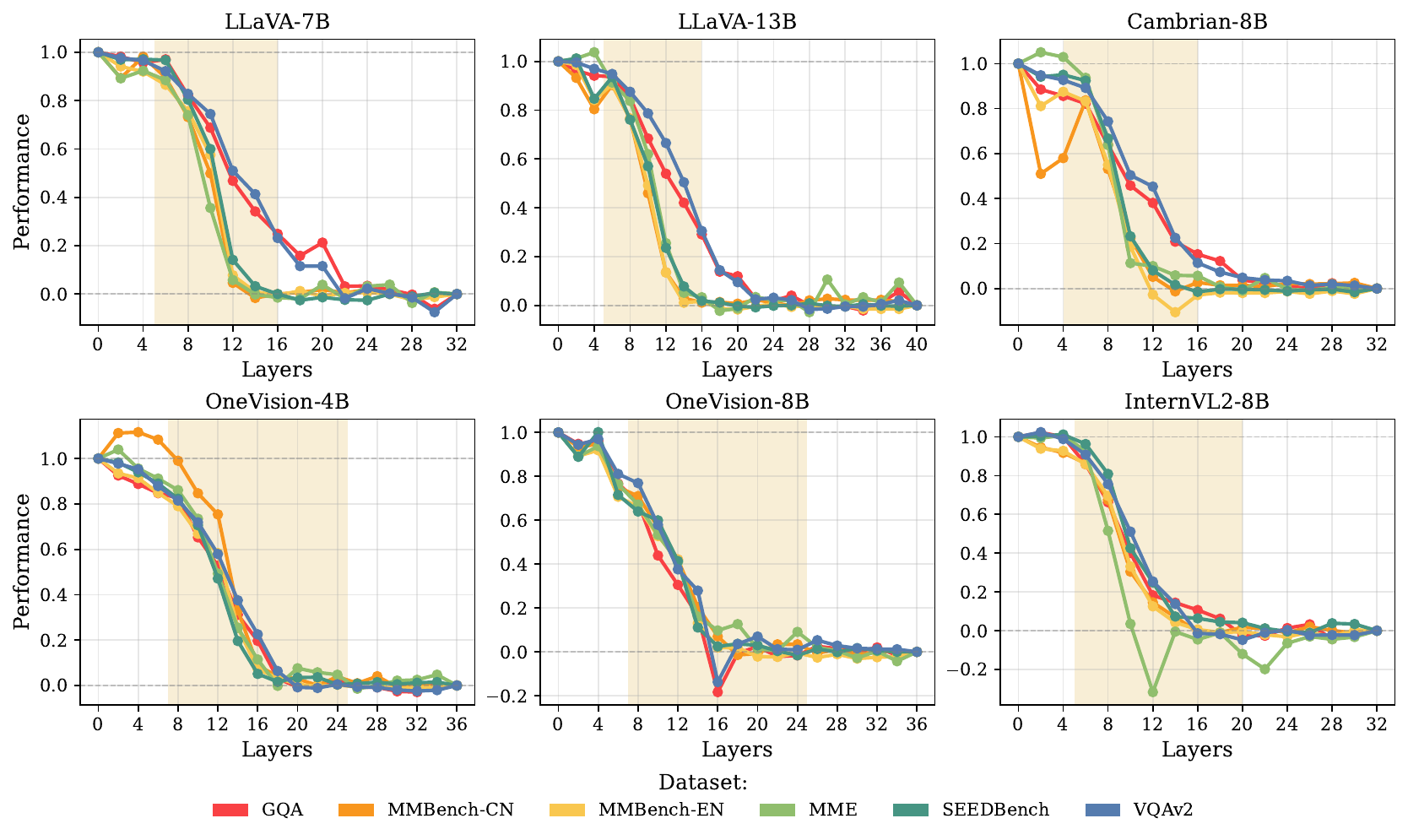}
\caption{\textbf{Progressive layer skipping on multimodal benchmarks.} Normalized performance as image-token layers are skipped cumulatively from the input, for each model and dataset (2,000 samples per dataset).}
\label{fig:app:mm_layer_skip_progressive}
\end{figure}

\subsubsection*{Sliding window layer-skipping}

\begin{figure}[H]
\centering
\includegraphics[width=0.9\linewidth]{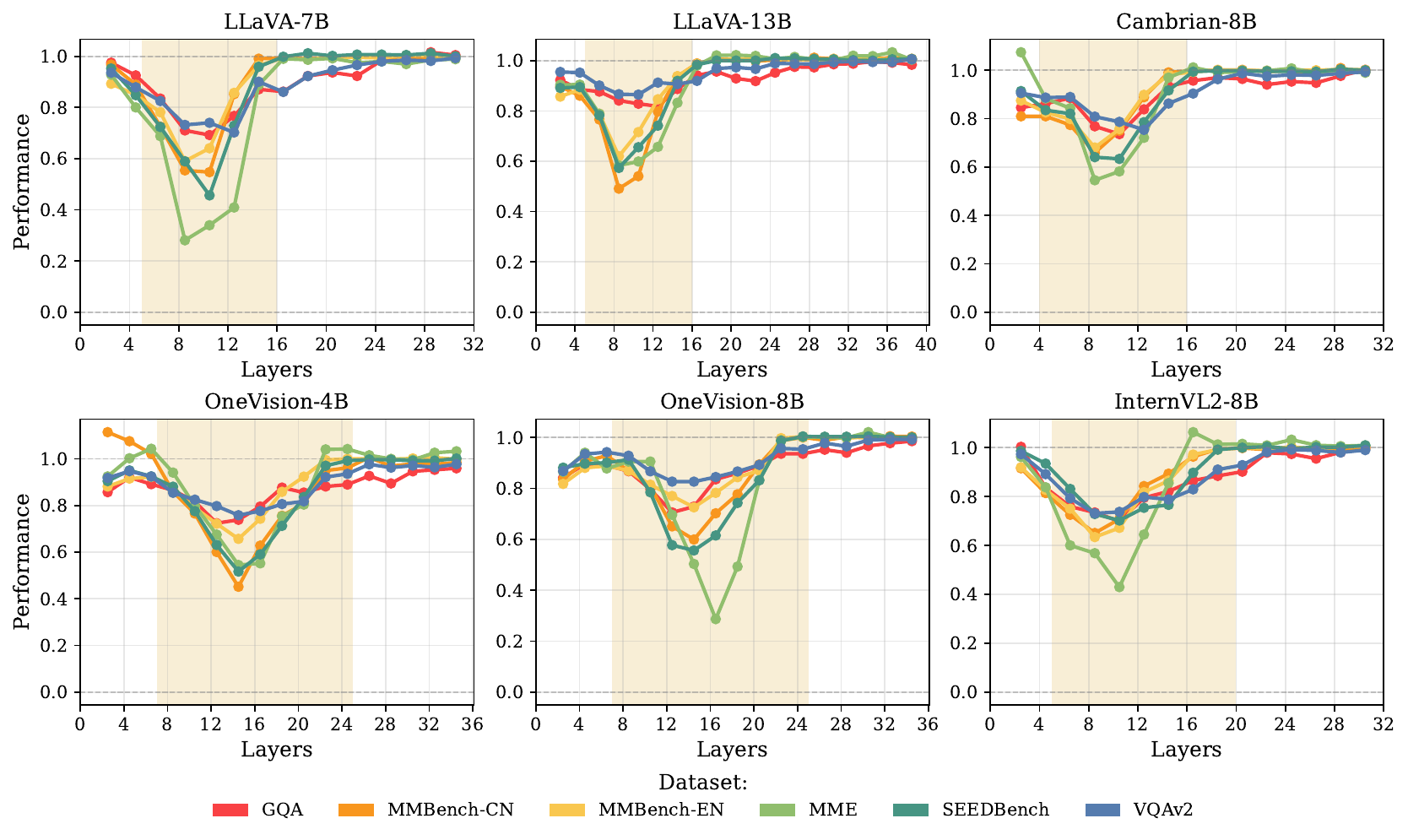}
\caption{\textbf{Sliding-window layer skipping on multimodal benchmarks.} Normalized performance when a window of five consecutive image-token layers is skipped.}
\label{fig:app:mm_layer_skip_sliding}
\end{figure}

\subsubsection*{Progressive attention knockout}

\begin{figure}[H]
\centering
\includegraphics[width=0.9\linewidth]{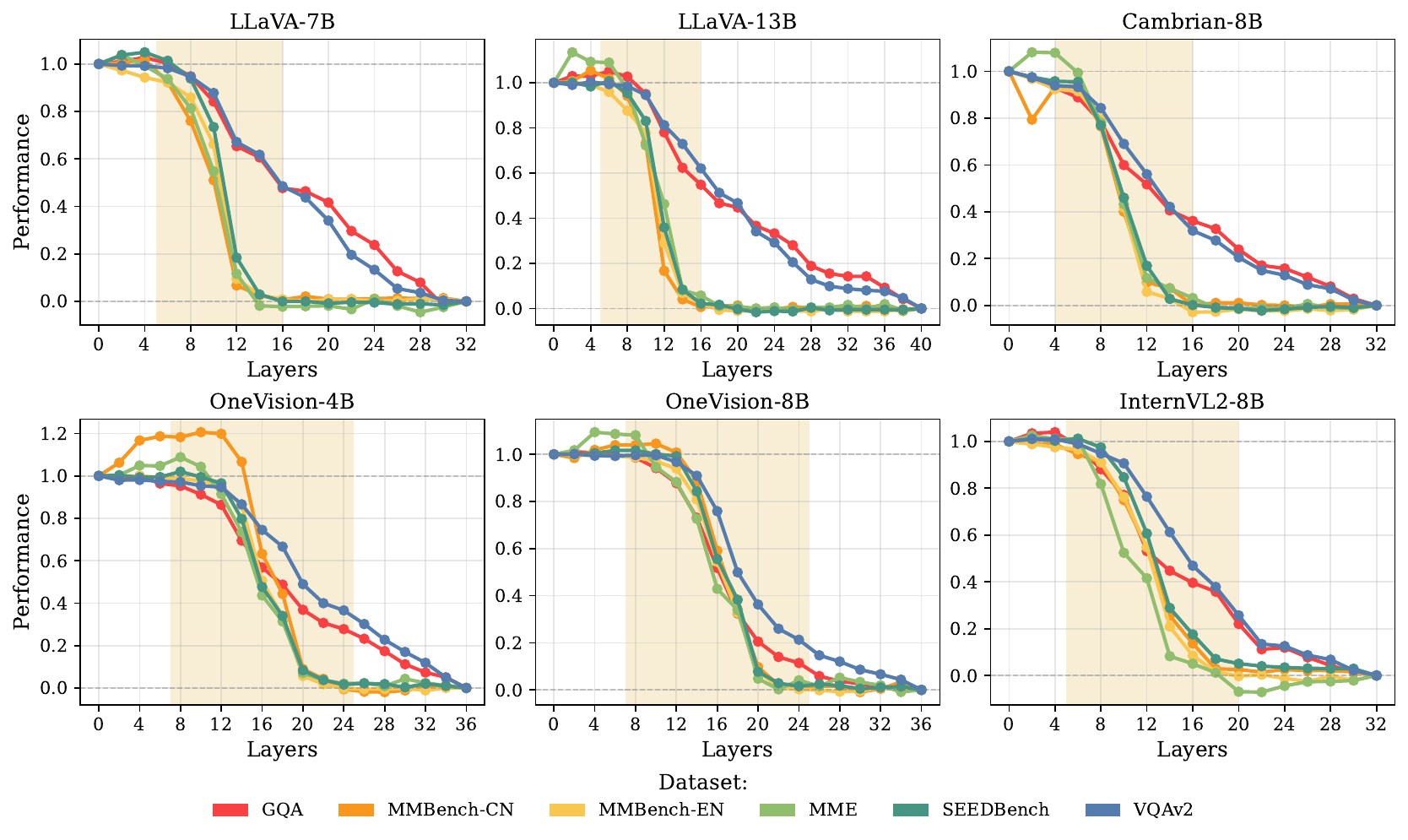}
\caption{\textbf{Progressive attention knockout on multimodal benchmarks.} Normalized performance as image-to-text attention is removed cumulatively from the input.}
\label{fig:app:mm_attn_knockout_progressive}
\end{figure}

\subsubsection*{Sliding window attention knockout}

\begin{figure}[H]
\centering
\includegraphics[width=0.9\linewidth]{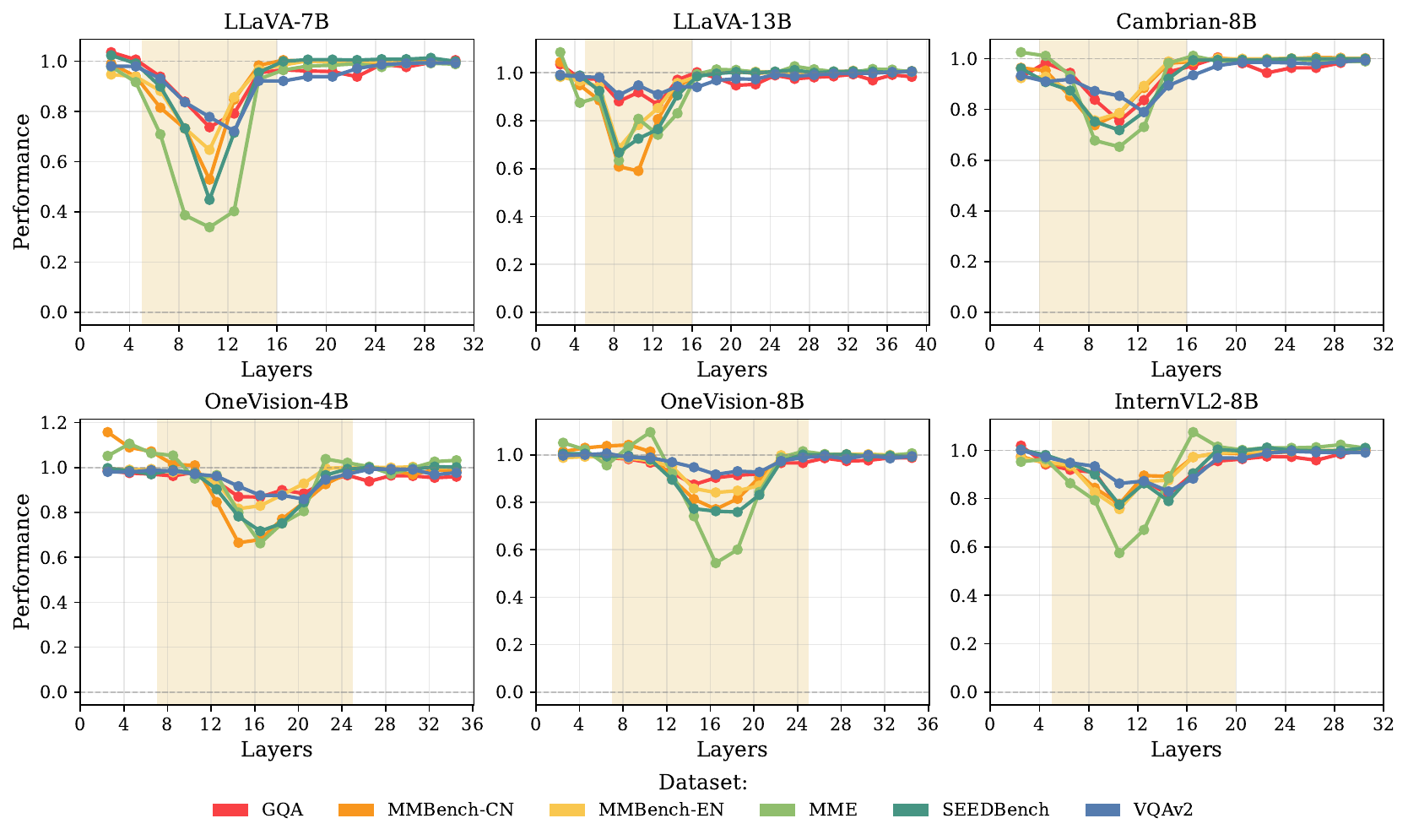}
\caption{\textbf{Sliding-window attention knockout on multimodal benchmarks.} Normalized performance when image-to-text attention is removed in a window of five consecutive layers.}
\label{fig:app:mm_attn_knockout_sliding}
\end{figure}

\subsection{Interventions: all datasets (text benchmarks)}

\subsubsection*{Progressive layer-skipping (all datasets)}

\begin{figure}[H]
\centering
\includegraphics[width=0.9\linewidth]{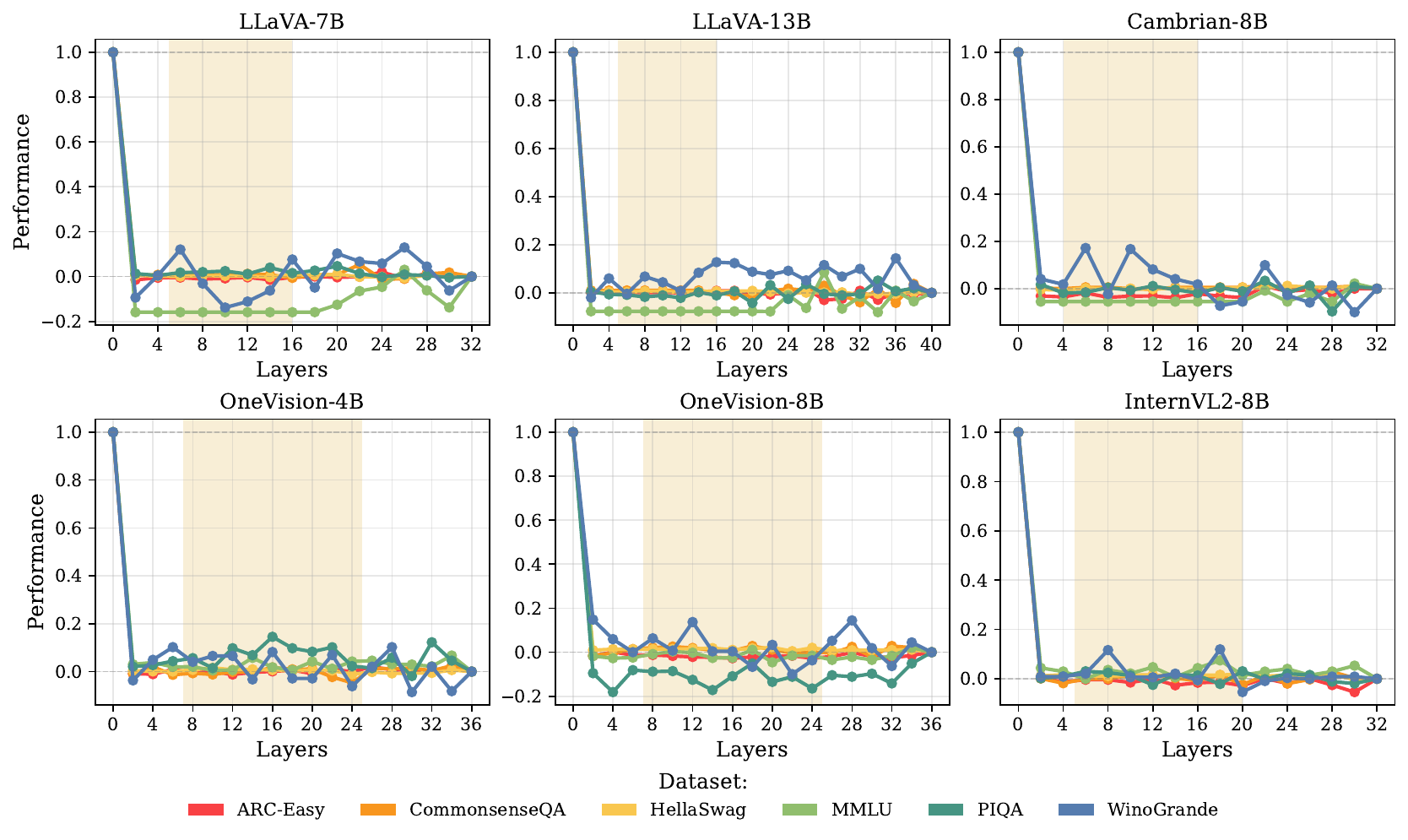}
\caption{\textbf{Progressive layer skipping on text-only benchmarks.} Normalized performance as layers are skipped cumulatively from the input, on the full evaluation sets of the six text benchmarks.}
\label{fig:app:text_layer_skip_progressive}
\end{figure}

\subsubsection*{Sliding window layer-skipping (all datasets)}

\begin{figure}[H]
\centering
\includegraphics[width=0.9\linewidth]{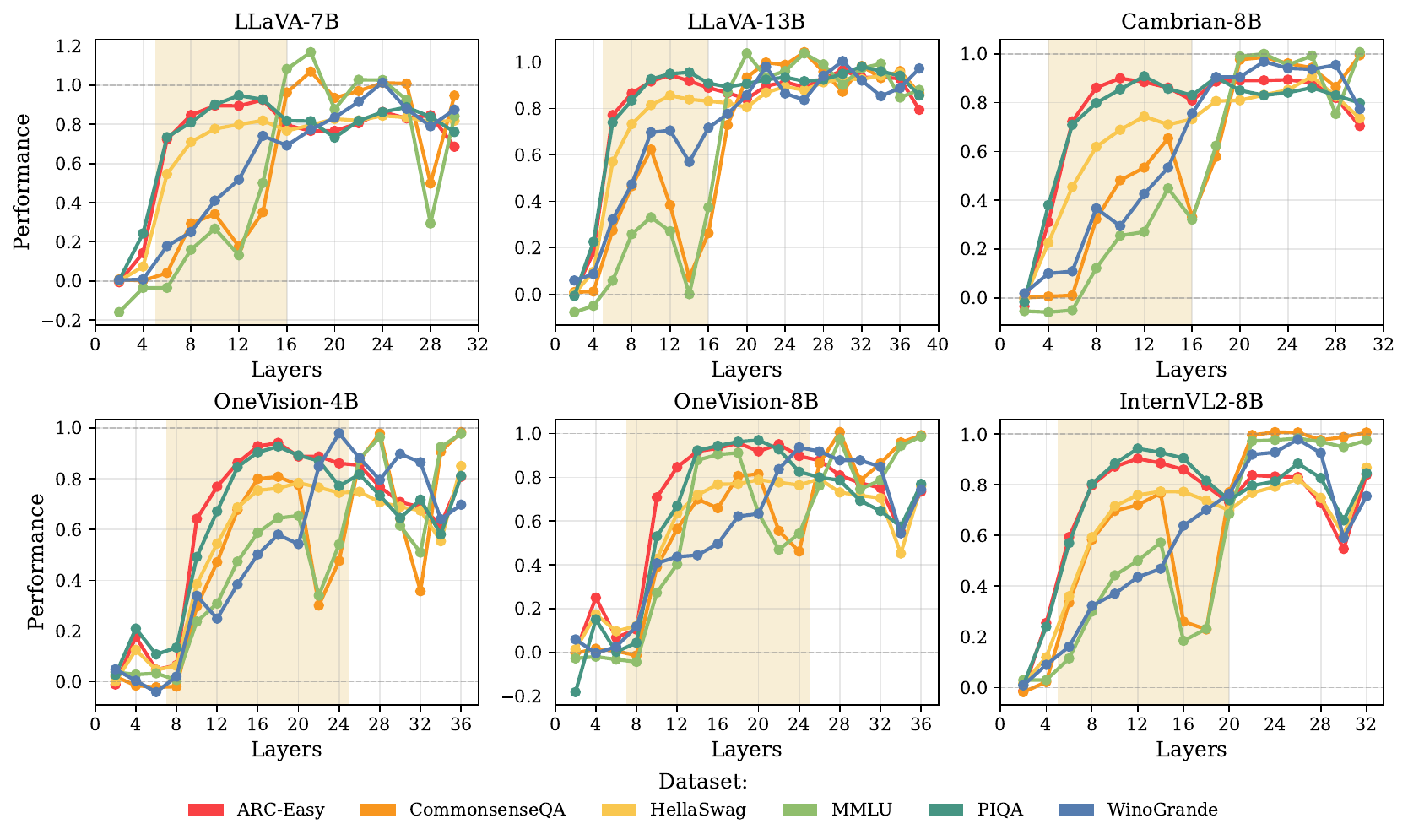}
\caption{\textbf{Sliding-window layer skipping on text-only benchmarks.} Normalized performance when a window of four consecutive layers is skipped.}
\label{fig:app:text_layer_skip_sliding}
\end{figure}

\section{Probing the semantic content of intermediate layers}\label{sec:app_probing}


\subsection{Averaged Label Overlap across Models}

\paragraph{Probing the semantic information with Label Overlap.}
\label{app:sec:no}
To evaluate how well the residual stream encodes an abstract property of the image, like the identity of the main object represented (e.g., ImageNet) 
we use the \emph{neighborhood overlap} \cite{doimo2020hierarchical}. 
The neighborhood overlap measures the consistency between the $k$-nearest neighbors of a data point $i$ in the residual stream at a layer $l$, denoted by $\mathcal{N}_k^{l}(i)$,  and the nearest neighbors in a reference data representation, $\mathcal{N}_k^{gt}(i)$, which encodes the abstract property of interest. 
In the case of classification datasets, all the points belonging to the same class are considered nearest neighbors for the purpose of computing the reference $\mathcal{N}_k^{gt}(i)$.
The \emph{neighborhood overlap}, denoted by $\chi_k^{l,gt}$, can be then written as:
\begin{equation}
\label{eq:NO}
    \chi_k^{l,gt}= \frac{1}{nk} \sum_{i=1}^n
    \left| \mathcal{N}_k^{l}(i) \cap \mathcal{N}_k^{gt}(i) \right|
\end{equation} 
where $|\cdot|$ denotes the cardinality of a set, and $n$ the dataset size. In our experiments, we set $k=30$.

\begin{figure}[H]
\centering
\includegraphics[width=0.5\textwidth]{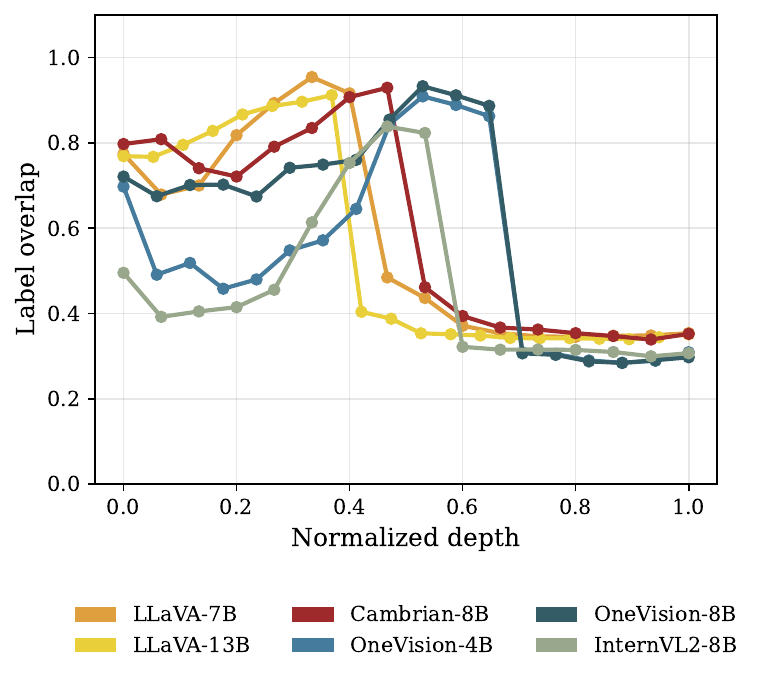}
\caption{\textbf{Average label overlap across depth.} Label overlap of last-token representations on 10-option classification prompts, averaged over the seven datasets of \Cref{tab:dataset_taxonomy}; one curve per model.}
\label{fig:app:label_overlap_average}
\end{figure}

\subsection{Label Overlap Profiles across Models and Datasets}\label{sec:lo_models_and_datasets}

\begin{figure}[H]
\centering
\includegraphics[width=0.9\textwidth]{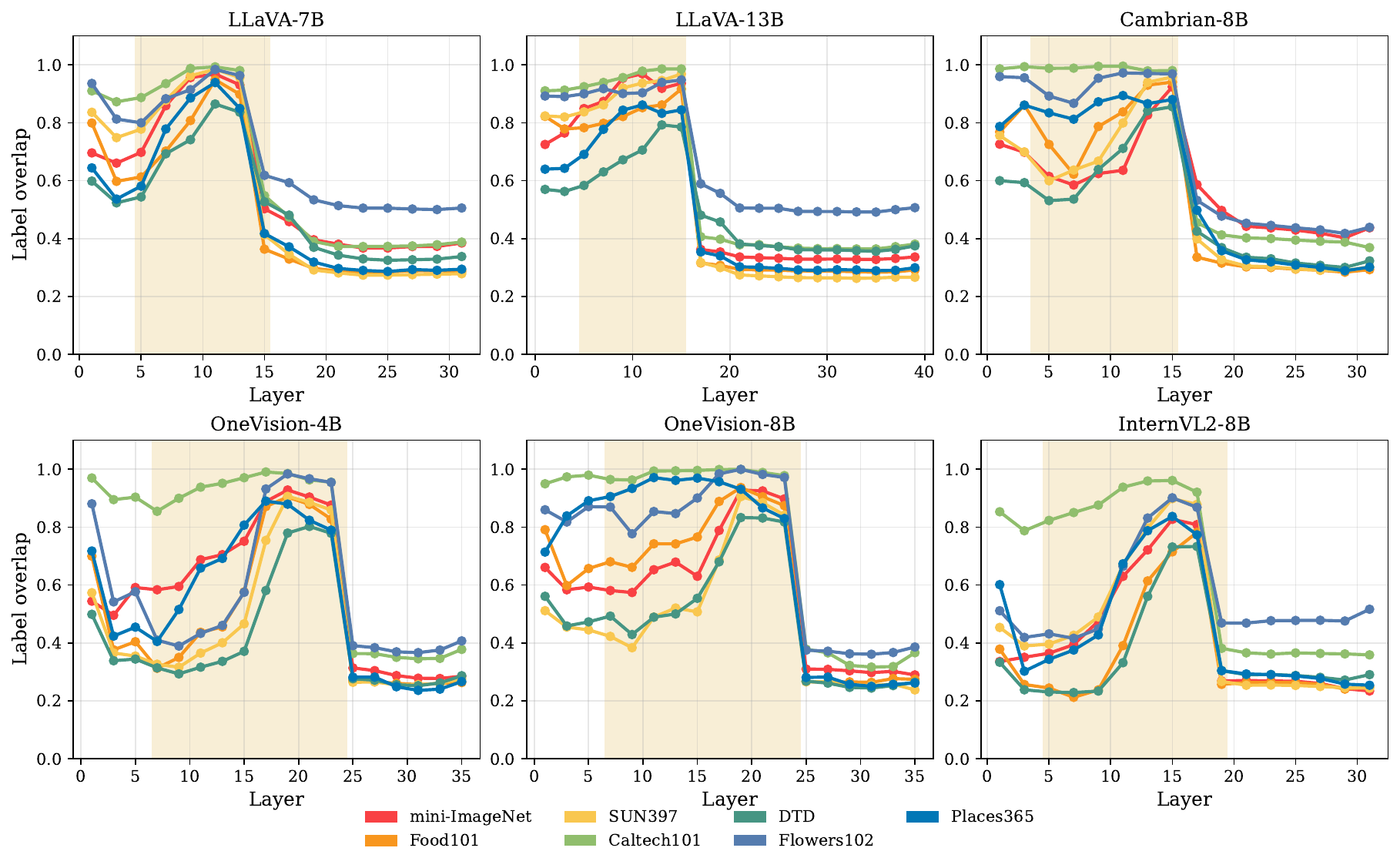}
\caption{\textbf{Label overlap per dataset.} Same setup as \Cref{fig:app:label_overlap_average}, shown separately for each classification dataset and model.}
\label{fig:app:label_overlap_by_dataset}
\end{figure}
\subsection{Linear Probing multiple-choice tasks}
\begin{figure}[H]
\centering
\includegraphics[width=0.9\textwidth]{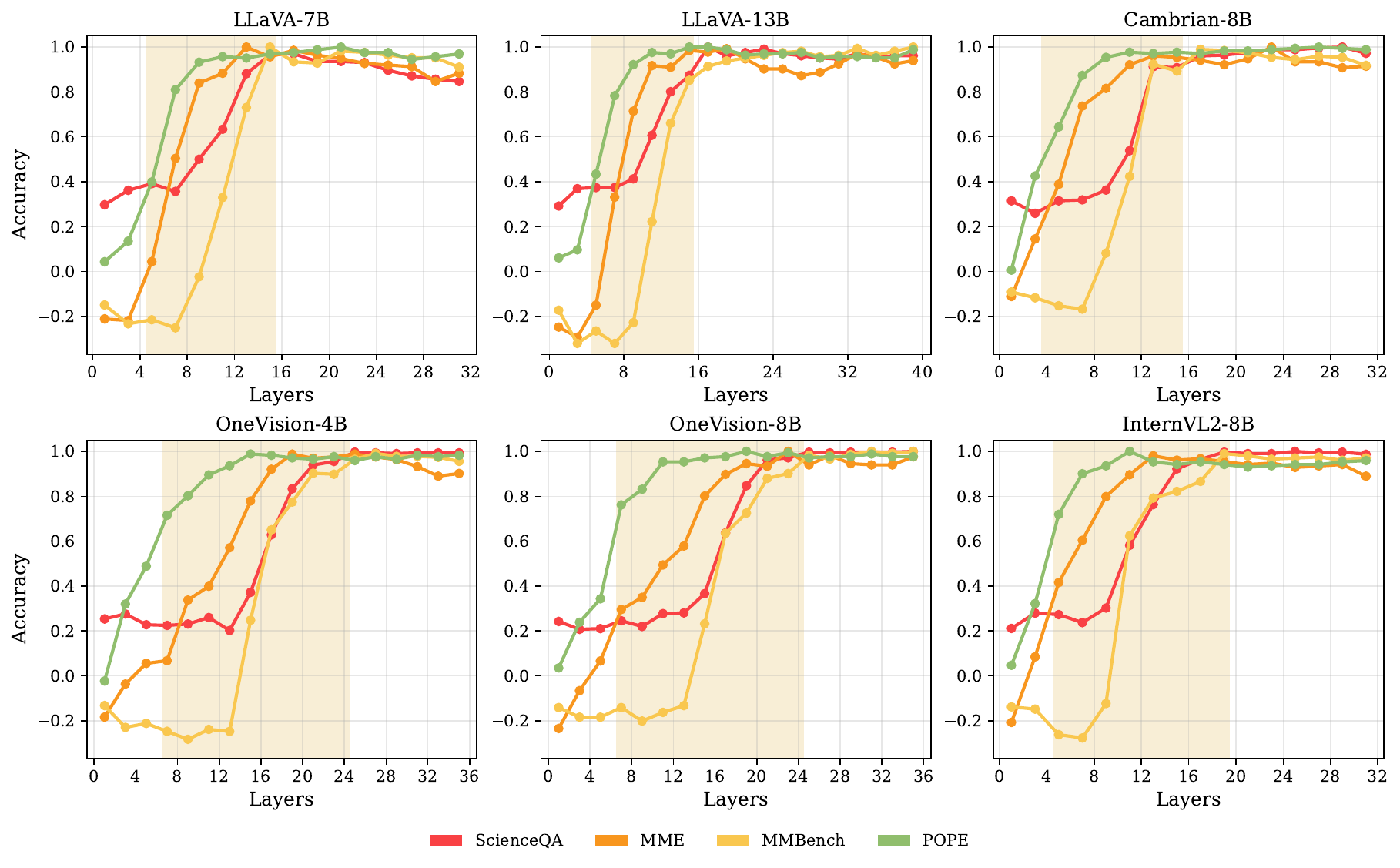}
\caption{\textbf{Linear probing on multiple-choice VQA tasks.} Layer-wise accuracy of a logistic-regression probe on the last-token representation, normalized against chance (2,000 samples per task).}
\label{fig:app:linear_probe_mc}
\end{figure}

\subsection{Logit Lens on multiple-choice tasks}
\begin{figure}[H]
\centering
\includegraphics[width=0.9\textwidth]{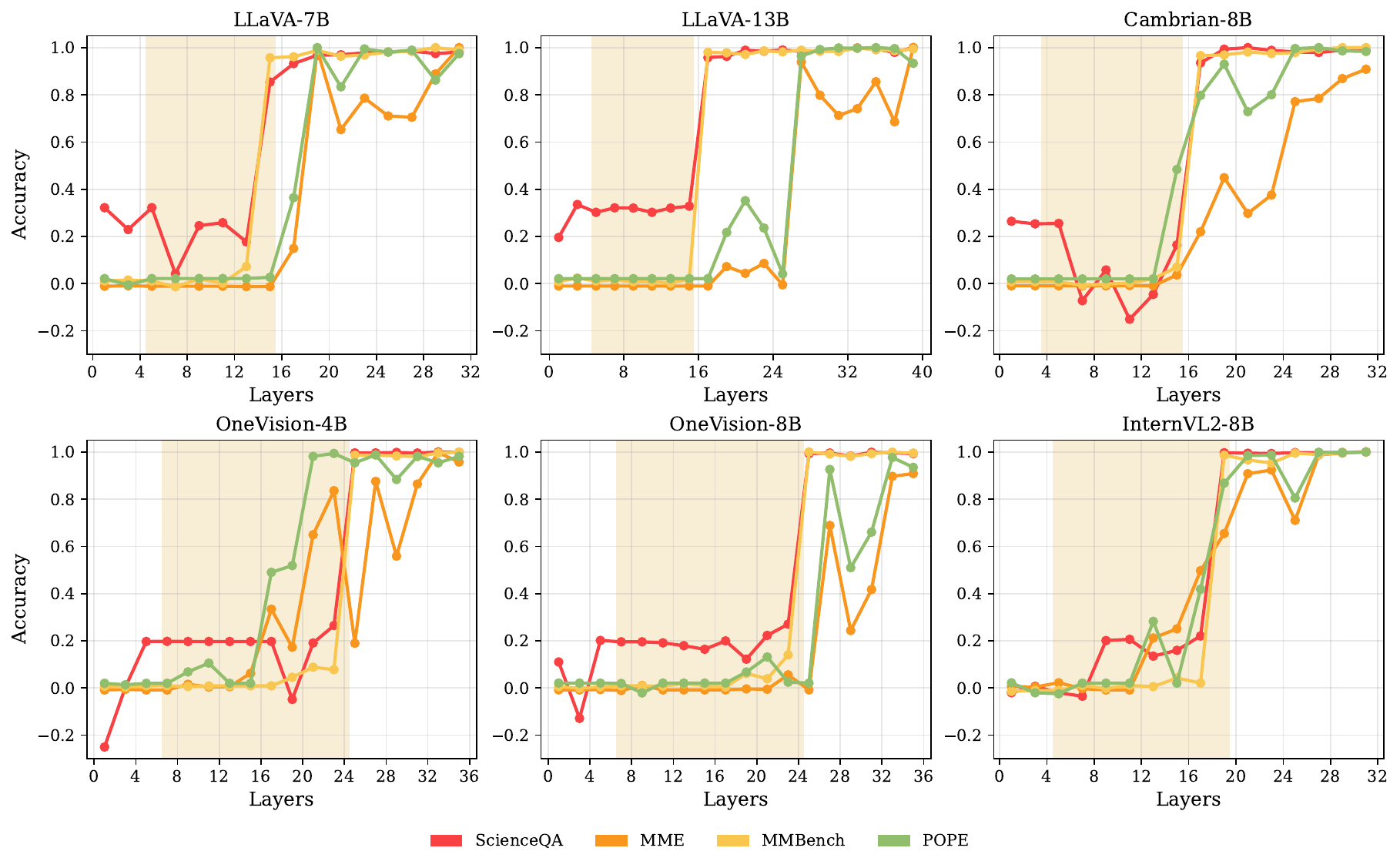}
\caption{\textbf{Logit lens on multiple-choice VQA tasks.} Layer-wise logit-lens accuracy, normalized against chance, on the same tasks as \Cref{fig:app:linear_probe_mc}.}
\label{fig:app:logit_lens_mc}
\end{figure}
\subsection{Logit Lens on open-ended tasks}
\begin{figure}[H]
\centering
\includegraphics[width=0.9\textwidth]{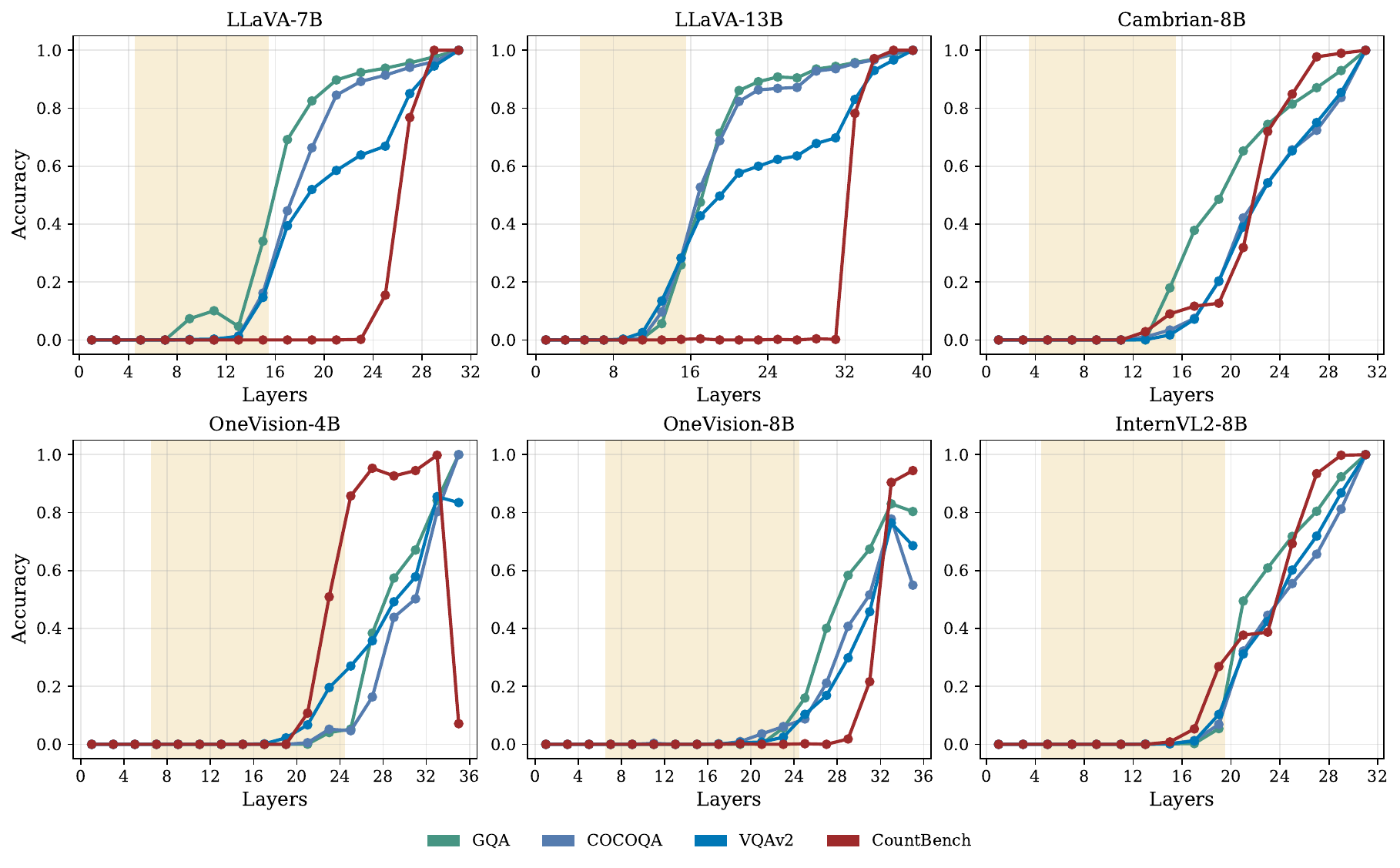}
\caption{\textbf{Logit lens on open-ended VQA tasks.} Top-10 token accuracy of the logit lens on tasks whose answers are not restricted to a fixed set of options (2,000 samples per task).}
\label{fig:app:logit_lens_oe}
\end{figure}

\subsection{Multimodal retrieval metrics.}
\label{app:sec:multimodal_retrieval_probing}

\begin{figure}[H]
\centering
\begin{minipage}{\textwidth}
\centering
\includegraphics[width=0.9\linewidth]{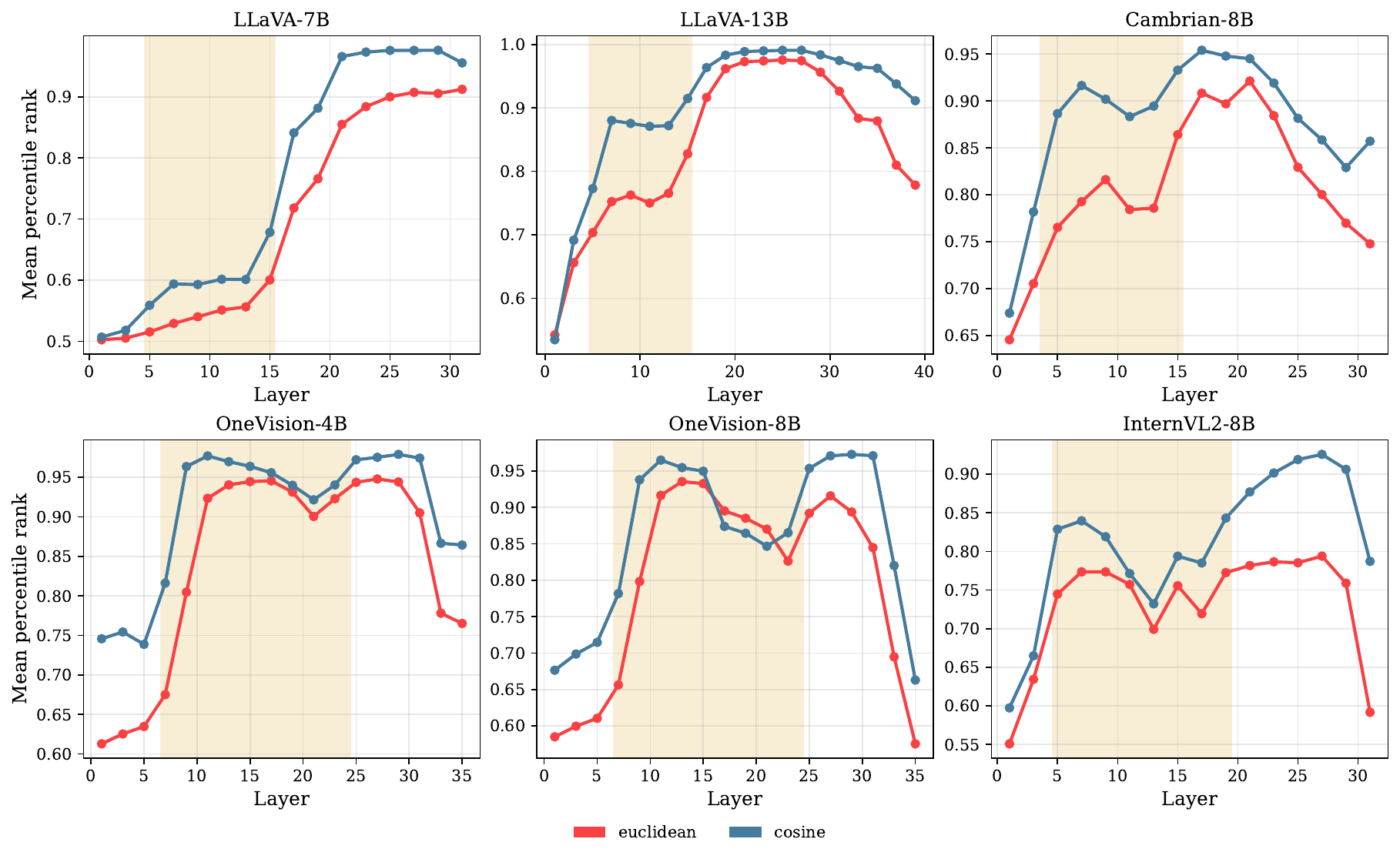}
\end{minipage}
\caption{\textbf{COCO image--caption retrieval (full LLaVA-Instruct detail-23k caption set).} At each layer, the hidden representations of an image and of its paired caption are averaged over tokens into one embedding each; curves show the mean percentile rank, averaged over the image-to-text and text-to-image directions, under Euclidean and cosine distance.}
\label{fig:app:retrieval_coco}
\end{figure}

\subsection{Probing semantic linguistic properties of hidden layers.}
\label{app:sec:text_semantic_probing}

\begin{figure}[H]
\centering
\begin{minipage}{0.9\textwidth}
\centering
\includegraphics[width=\linewidth]{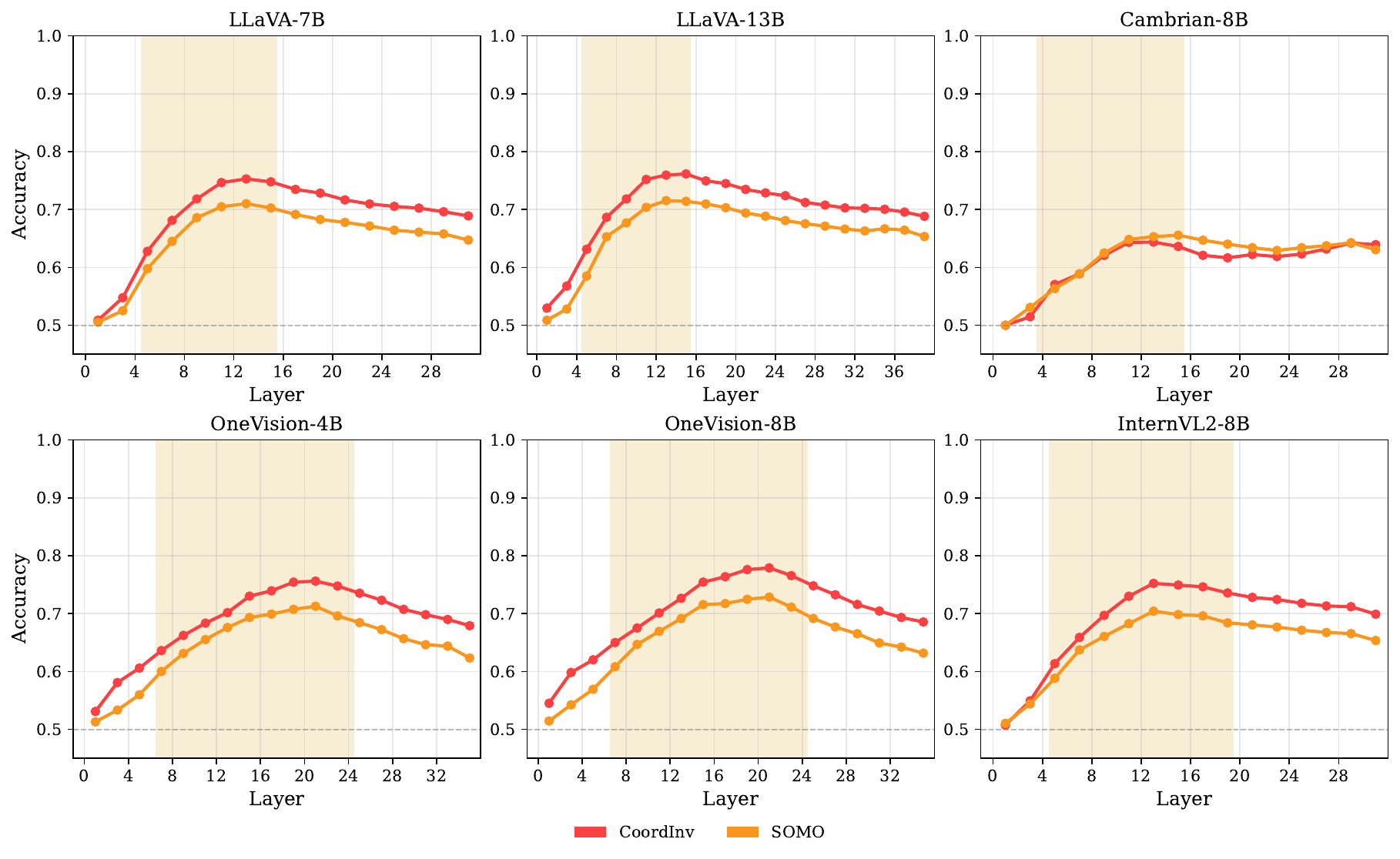}
\end{minipage}
\caption{\textbf{Probing semantic linguistic properties.} Layer-wise accuracy of a logistic-regression probe on the SentEval tasks \cite{conneau-etal-2018-cram} coordination inversion (CoordInv) and semantic odd man out (SOMO), 5,000 sentences each; the dashed gray line marks chance.}

\label{fig:app:linguistic_probing}
\end{figure}

\clearpage 

\section{Information imbalance planes}
\label{app:sec:imbalances}

\begin{figure}[H]
\centering
\begin{minipage}{0.48\textwidth}
\centering
\includegraphics[width=\linewidth]{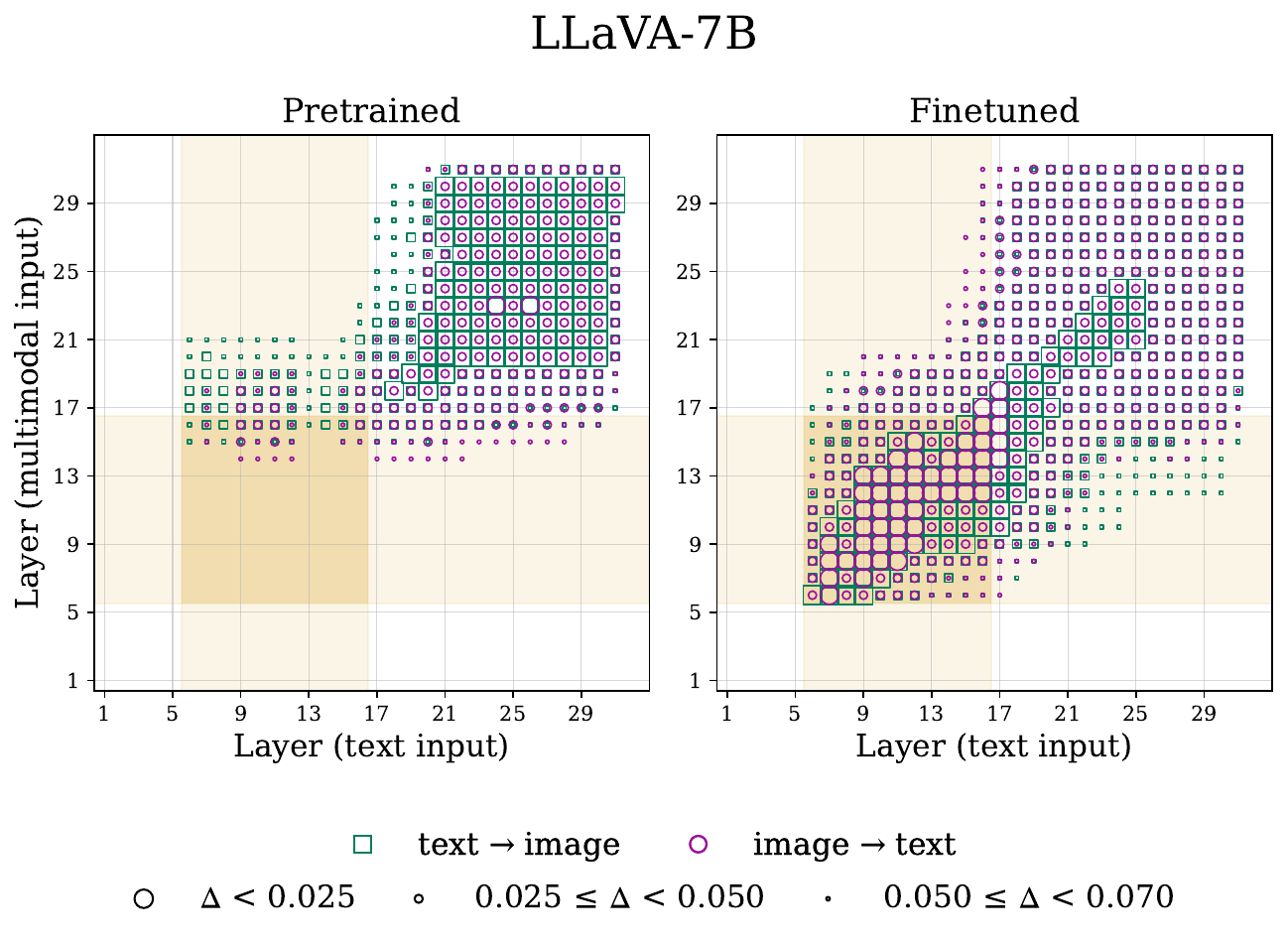}\hfill
\end{minipage}\hfill
\begin{minipage}{0.48\textwidth}
\centering
\includegraphics[width=\linewidth]{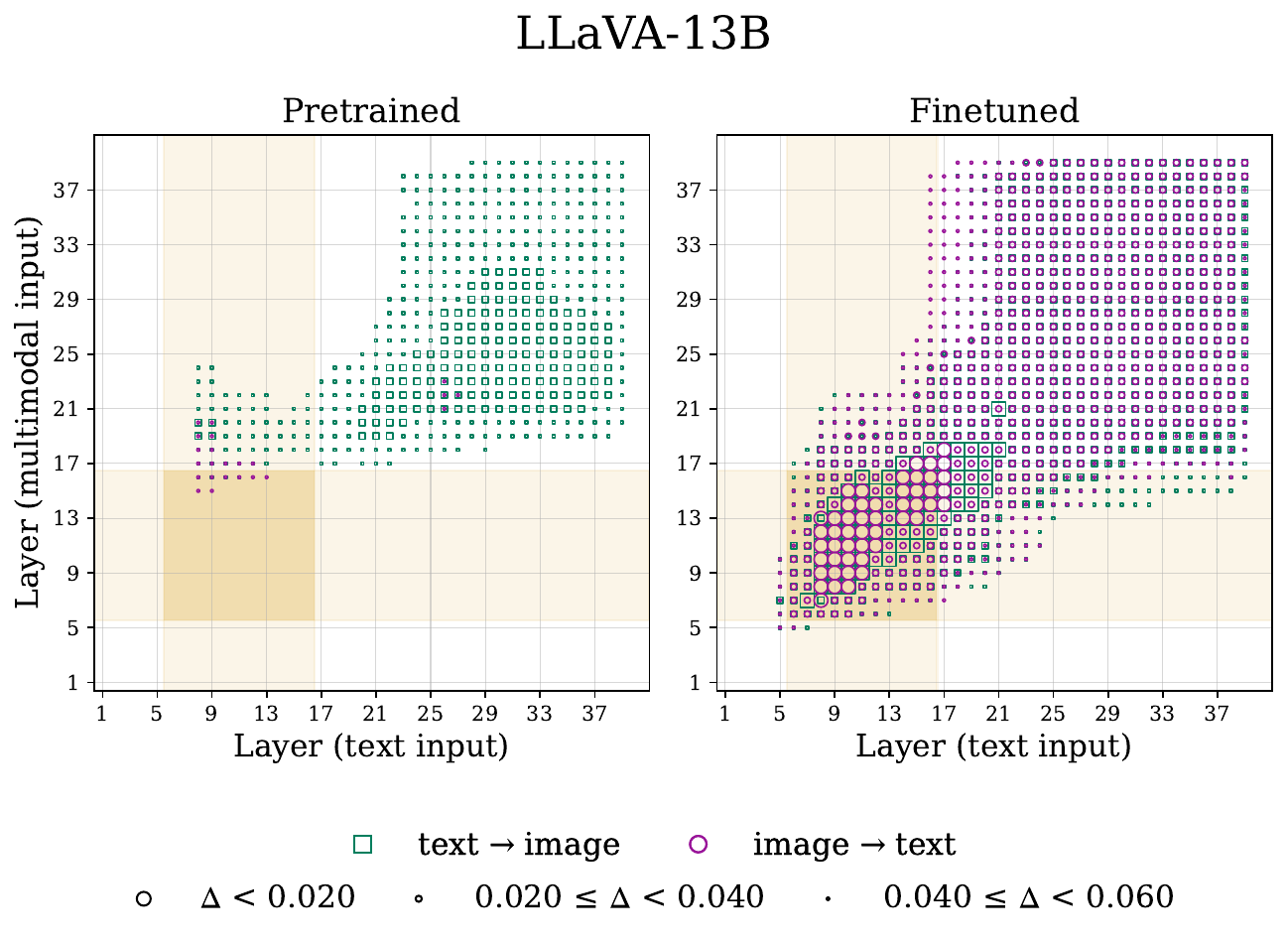}\hfill
\end{minipage}\hfill
\caption{\textbf{Information imbalance planes for LLaVA-1.5 (7B and 13B).} Same setup as \Cref{fig:information_imbalance}, before and after instruction tuning.}
\label{fig:app:information_imbalance_llava}
\end{figure}


\begin{figure}[H]
\centering
\begin{minipage}{0.48\textwidth}
\centering
\includegraphics[width=\linewidth]{final_figures/geometry/information_imbalance/llava_ov_4b.pdf}\hfill
\end{minipage}\hfill
\begin{minipage}{0.48\textwidth}
\centering
\includegraphics[width=\linewidth]{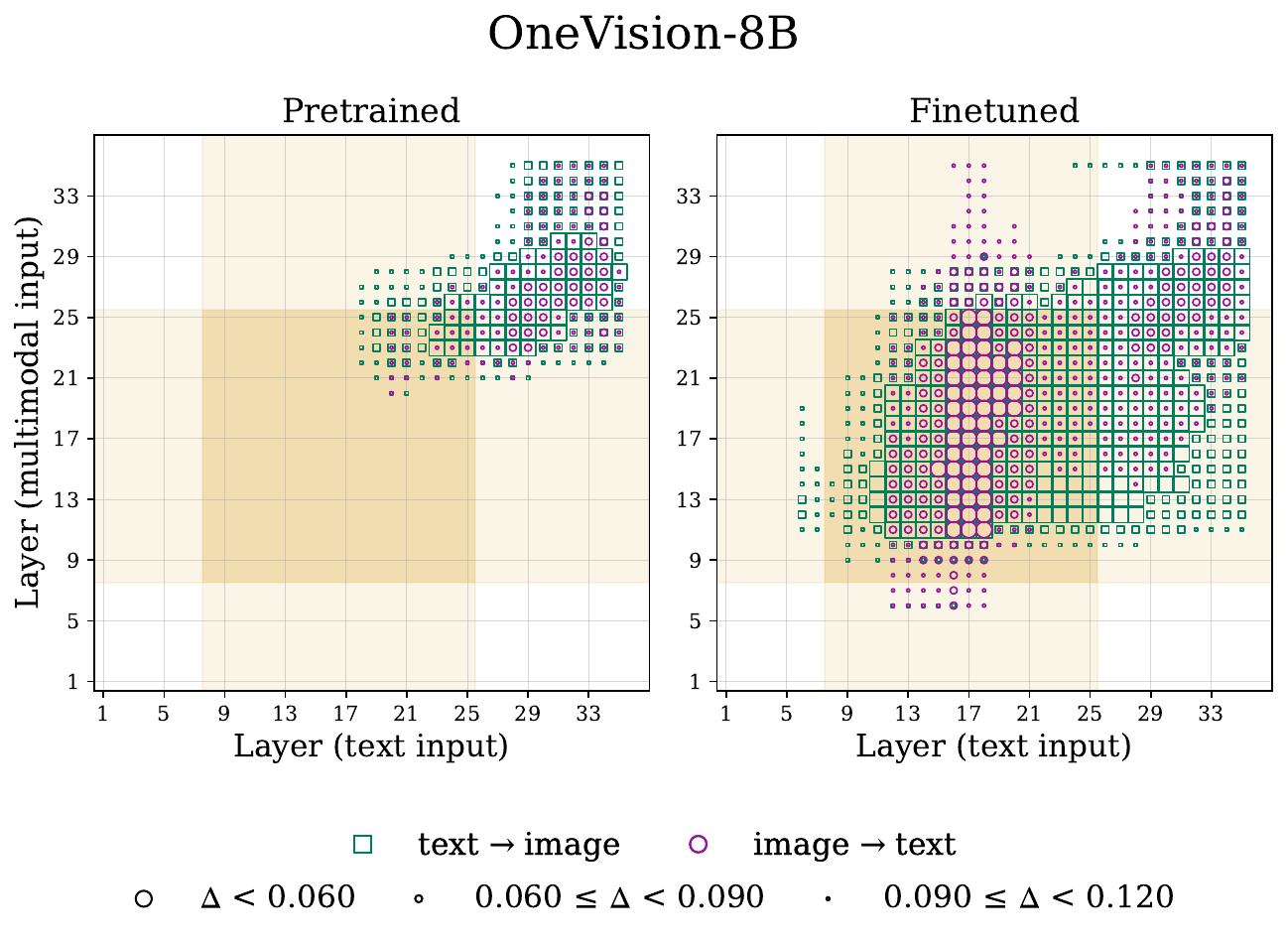}\hfill
\end{minipage}\hfill
\caption{\textbf{Information imbalance planes for LLaVA-OneVision-1.5 (4B and 8B).}}
\label{fig:app:information_imbalance_onevision}
\end{figure}

\begin{figure}[H]
\centering
\begin{minipage}{0.48\textwidth}
\centering
\includegraphics[width=\linewidth]{final_figures/geometry/information_imbalance/internvl2_8b.pdf}\hfill
\end{minipage}\hfill
\begin{minipage}{0.48\textwidth}
\centering
\includegraphics[width=\linewidth]{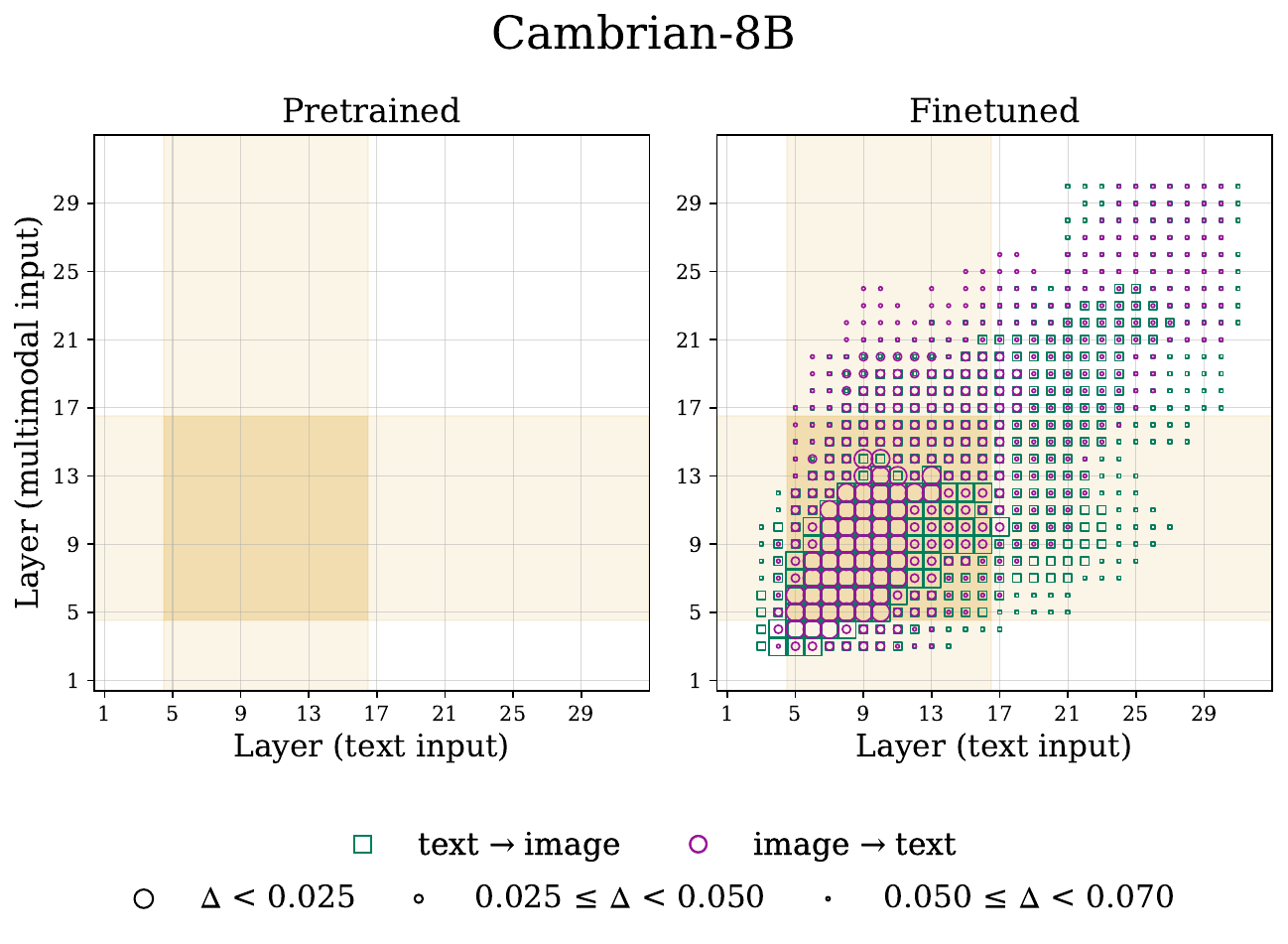}\hfill
\end{minipage}\hfill
\caption{\textbf{Information imbalance planes for InternVL2-8B and Cambrian-8B.} The pre-trained Cambrian-8B has imbalance values above the plotted range, strongly biased towards the textual inputs.}
\label{fig:app:information_imbalance_internvl_cambrian}
\end{figure}

\clearpage 

\section{Other similarity metrics: Linear CKA and Neighborhood overlap.}
\label{app:cka_overlap}

\subsection{Linear CKA}

\begin{figure}[H]
\centering
\begin{minipage}{0.48\textwidth}
\centering
\includegraphics[width=\linewidth]{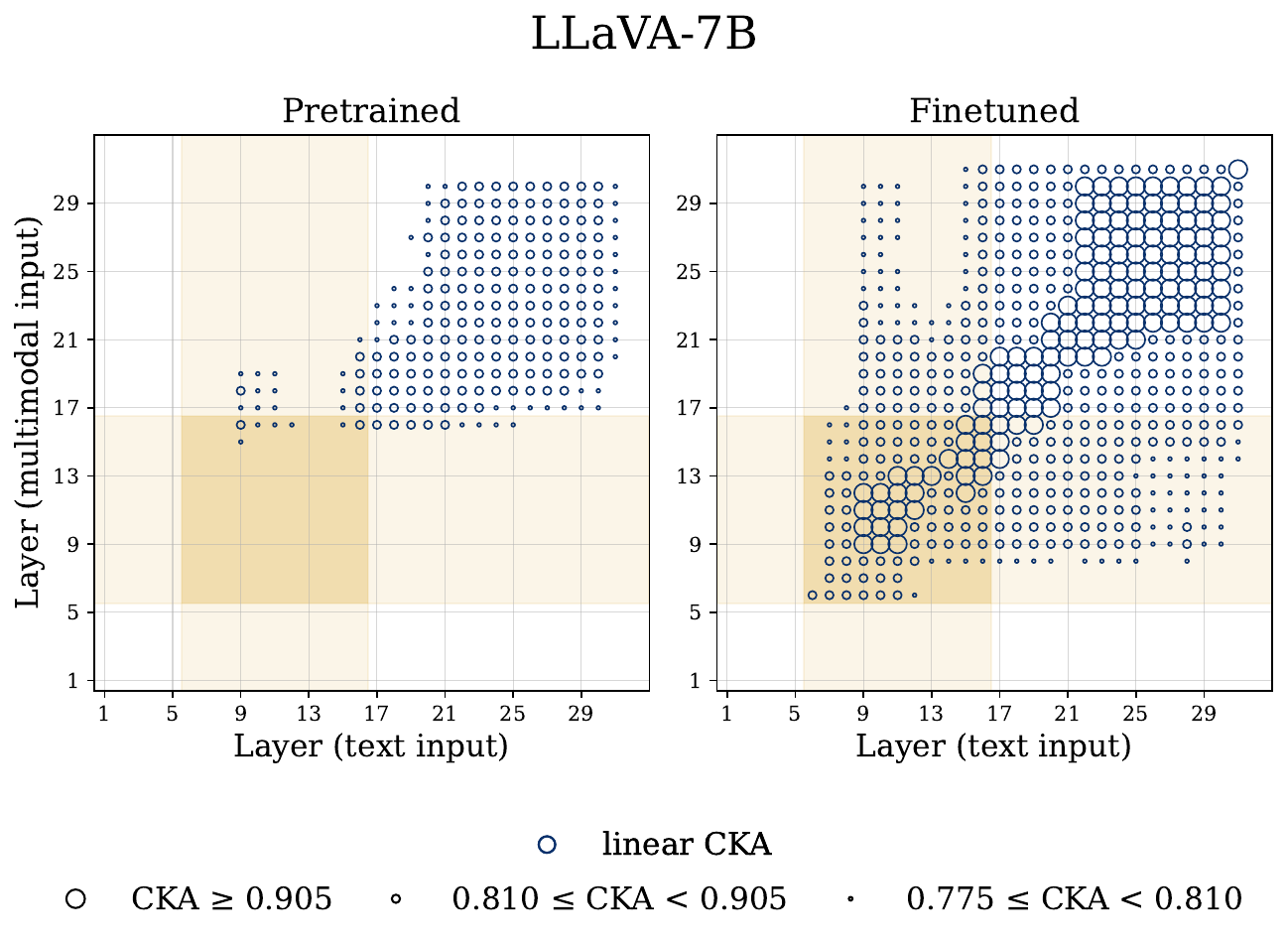}\hfill
\end{minipage}\hfill
\begin{minipage}{0.48\textwidth}
\centering
\includegraphics[width=\linewidth]{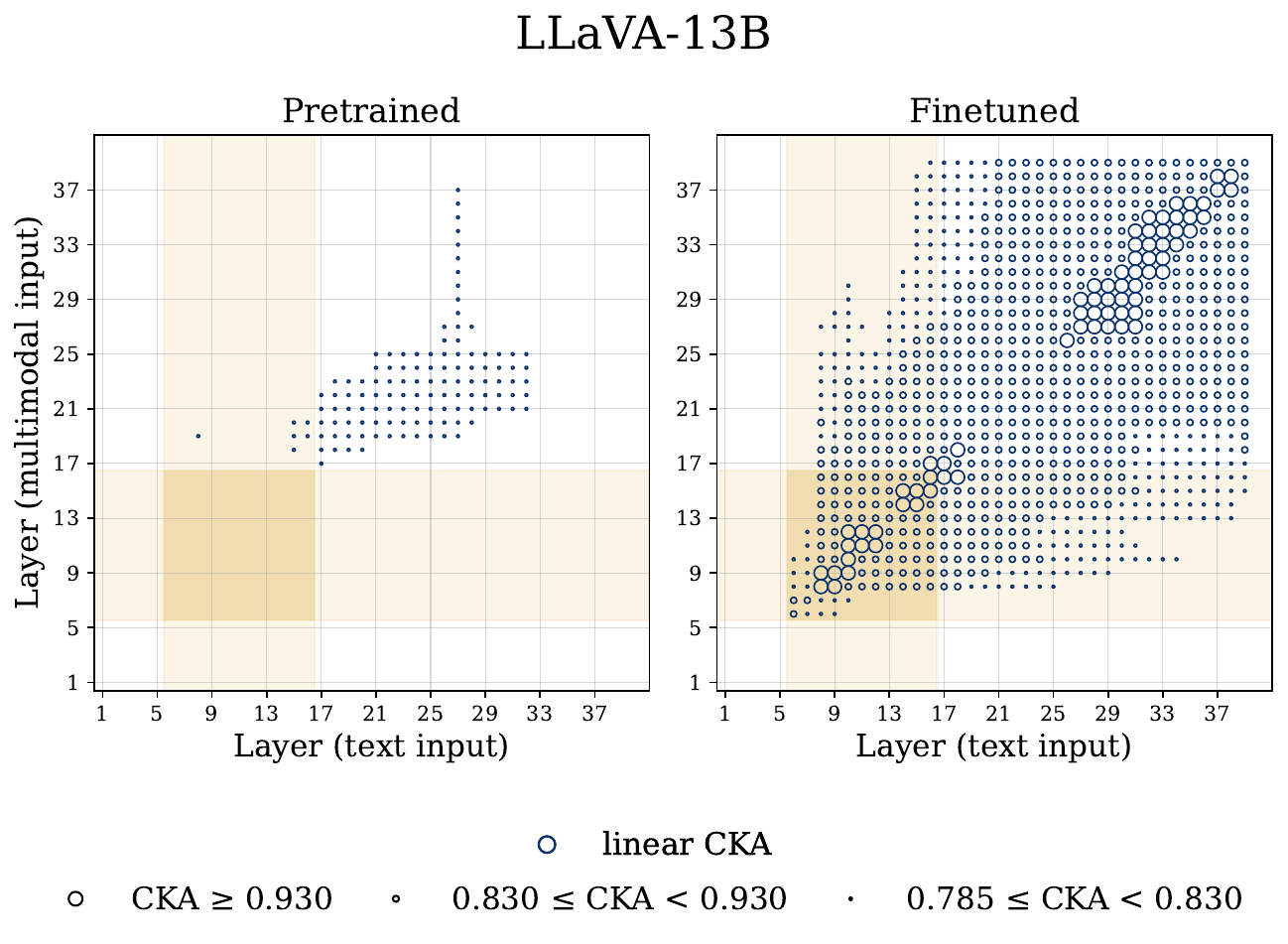}\hfill
\end{minipage}\hfill
\caption{\textbf{Linear CKA between text-only and multimodal representations for LLaVA-1.5 (7B and 13B).} Each panel compares the layers of the text-only prompt (x-axis) with those of the multimodal prompt (y-axis), before and after instruction tuning. Larger markers denote higher CKA, layer pairs below the lowest value in the legend are not drawn, and the yellow band marks the abstraction region. Values are averaged over the same datasets as \Cref{fig:information_imbalance}, using the full datasets.}
\label{fig:app:linear_cka_llava}
\end{figure}


\begin{figure}[H]
\centering
\begin{minipage}{0.48\textwidth}
\centering
\includegraphics[width=\linewidth]{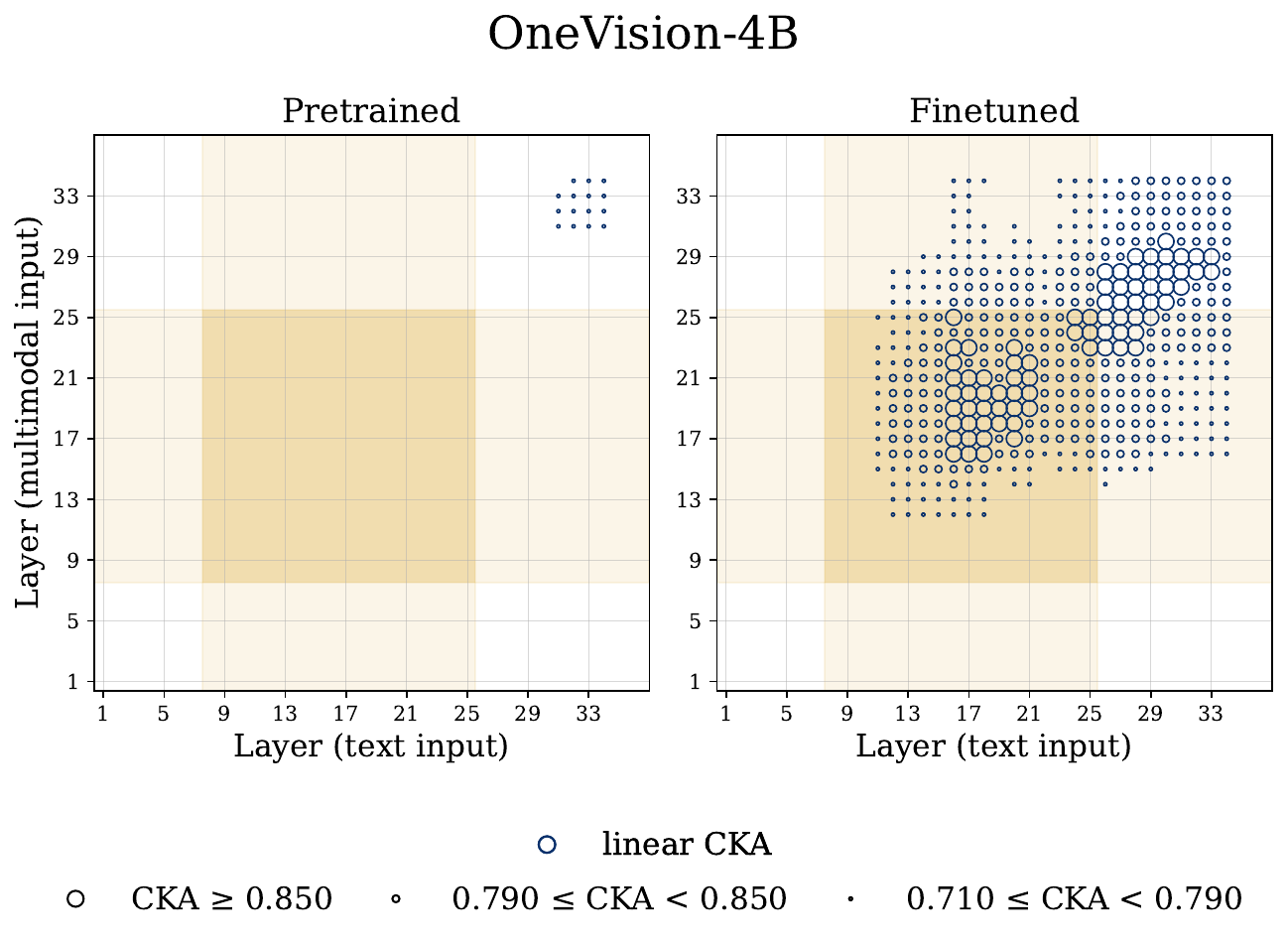}\hfill
\end{minipage}\hfill
\begin{minipage}{0.48\textwidth}
\centering
\includegraphics[width=\linewidth]{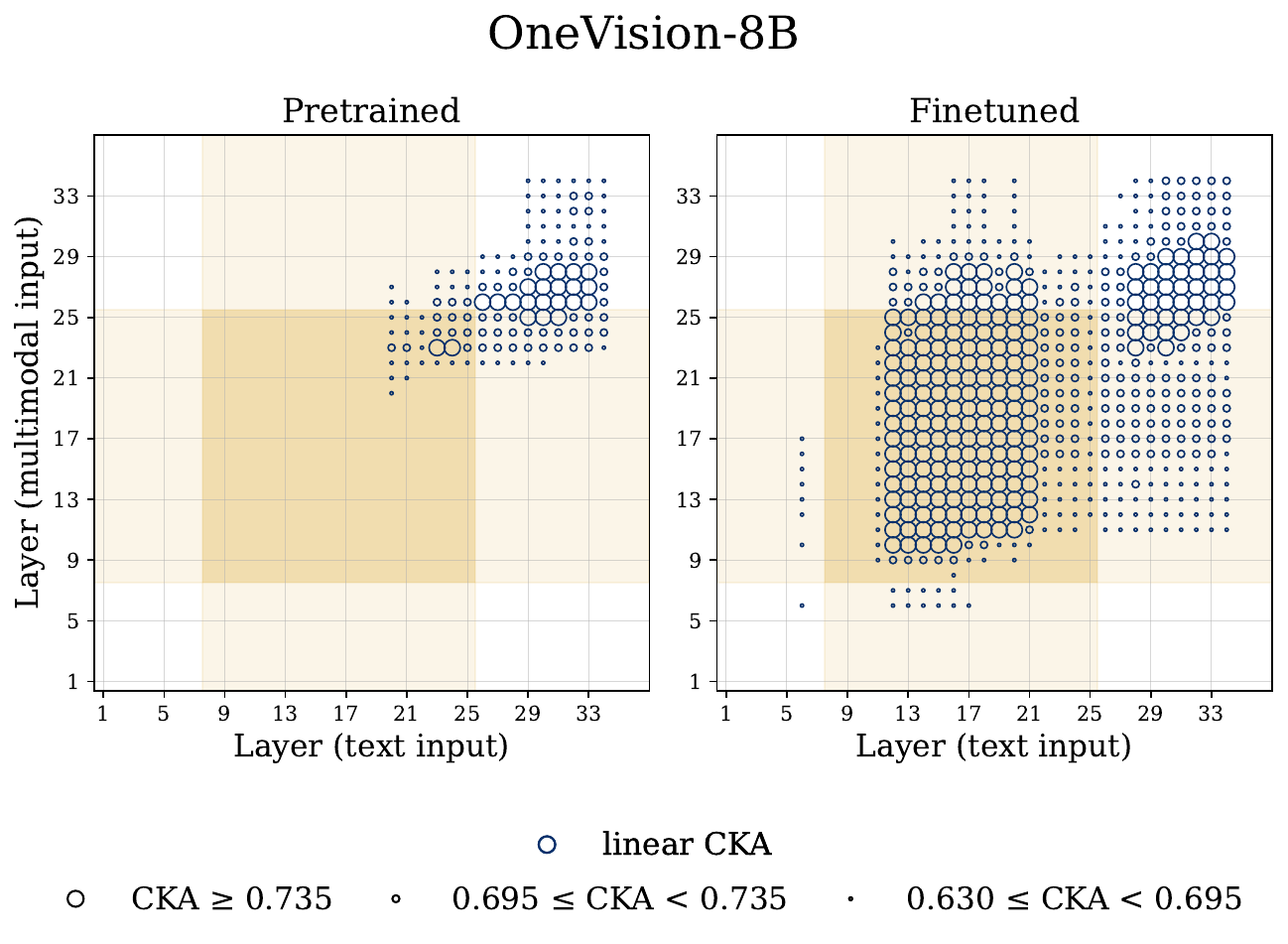}\hfill
\end{minipage}\hfill
\caption{\textbf{Linear CKA for LLaVA-OneVision-1.5 (4B and 8B).}}
\label{fig:app:linear_cka_onevision}
\end{figure}


\begin{figure}[H]
\centering
\begin{minipage}{0.48\textwidth}
\centering
\includegraphics[width=\linewidth]{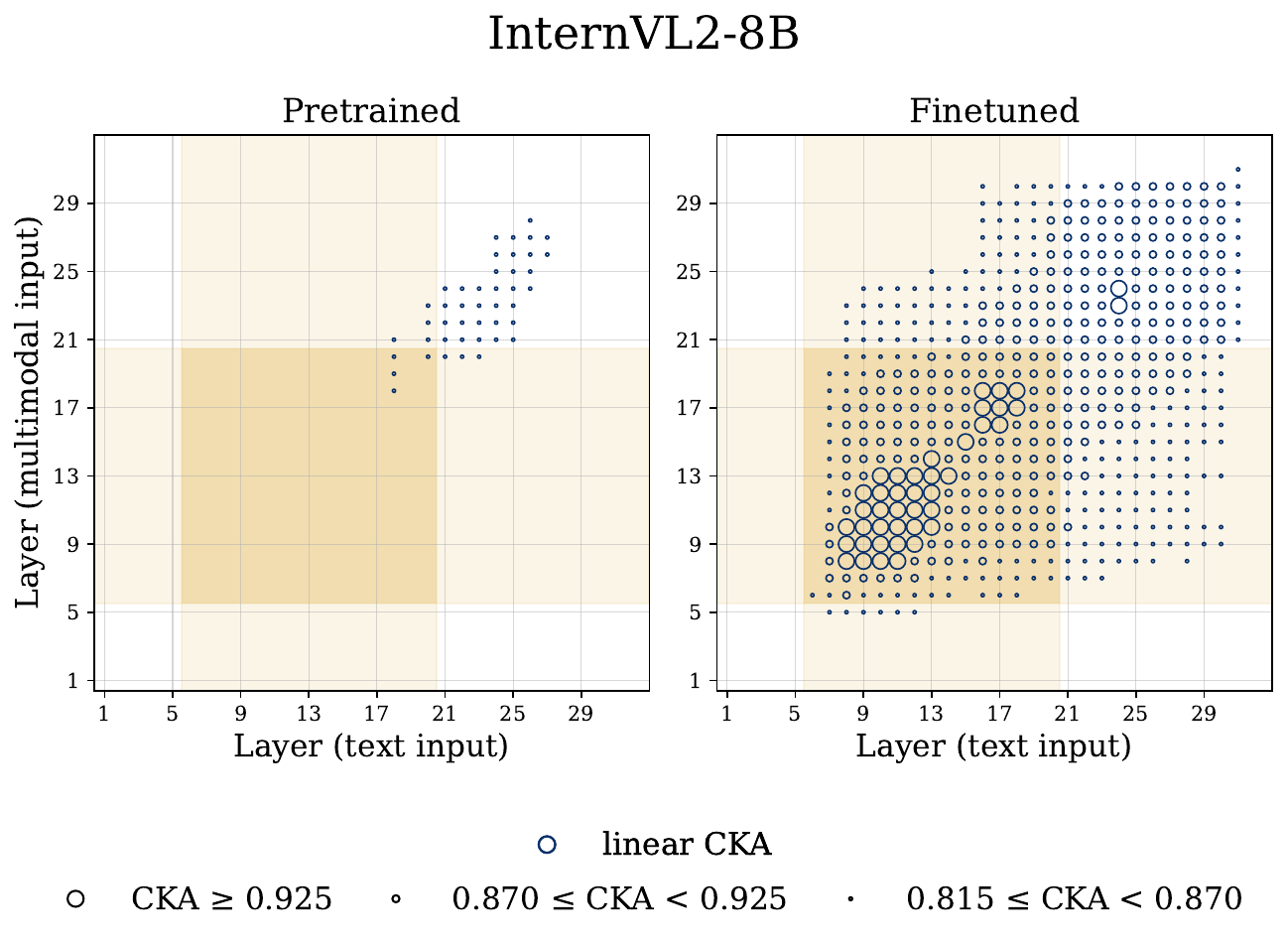}\hfill
\end{minipage}\hfill
\begin{minipage}{0.48\textwidth}
\centering
\includegraphics[width=\linewidth]{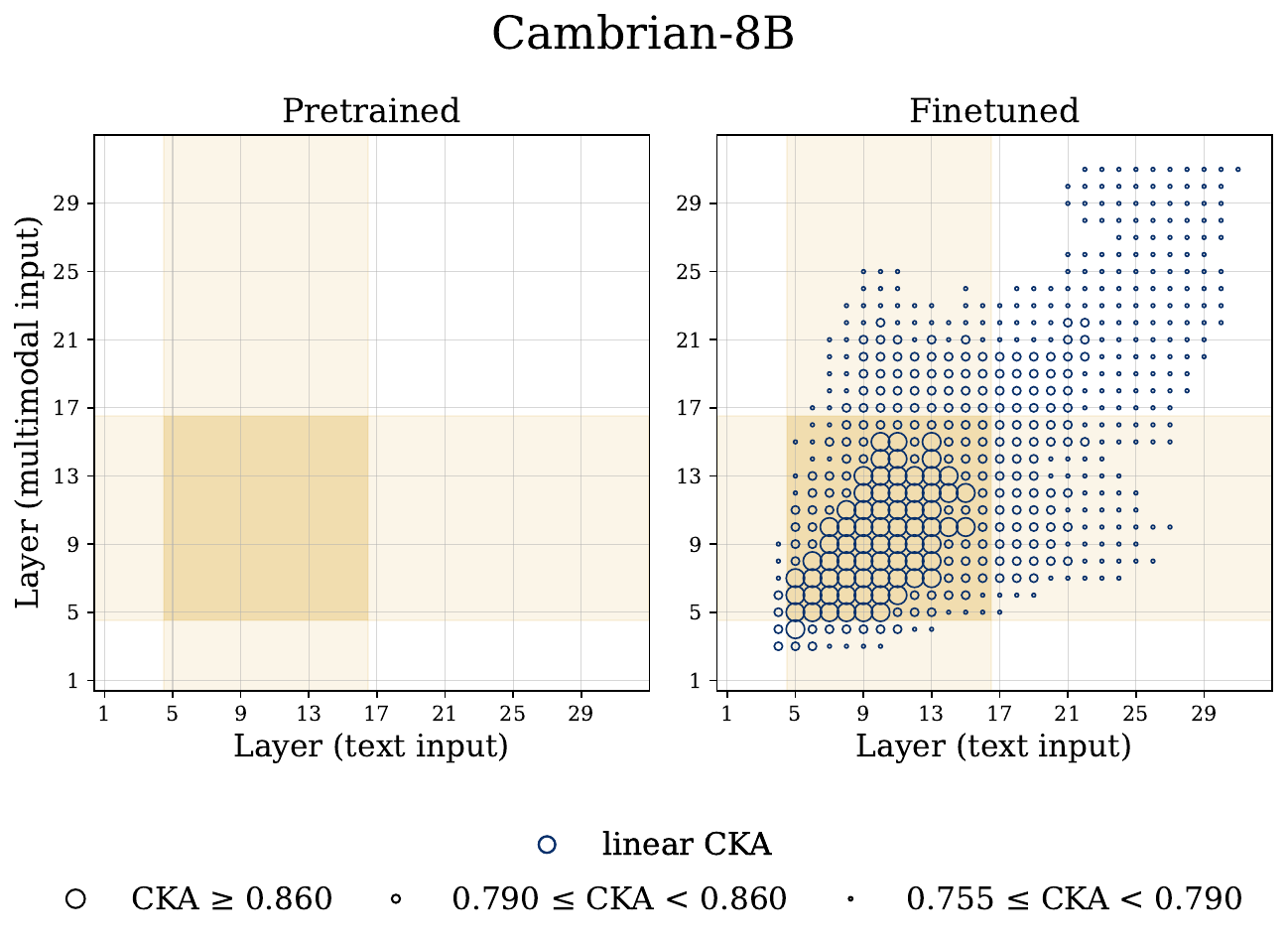}\hfill
\end{minipage}\hfill
\caption{\textbf{Linear CKA for InternVL2-8B and Cambrian-8B.}}
\label{fig:app:linear_cka_internvl_cambrian}
\end{figure}

\subsection{Neighborhood overlap}

\begin{figure}[H]
\centering
\begin{minipage}{0.48\textwidth}
\centering
\includegraphics[width=\linewidth]{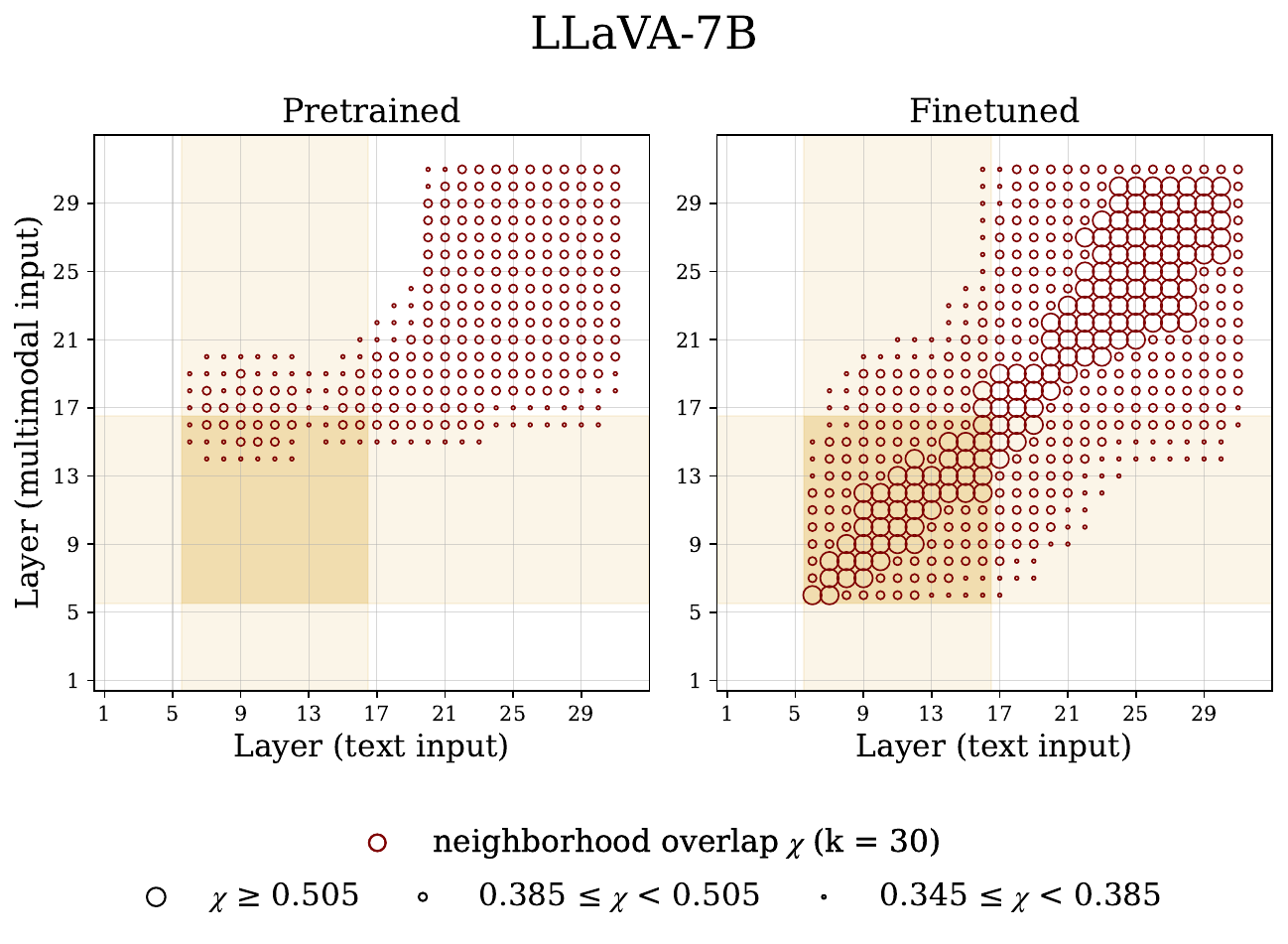}\hfill
\end{minipage}\hfill
\begin{minipage}{0.48\textwidth}
\centering
\includegraphics[width=\linewidth]{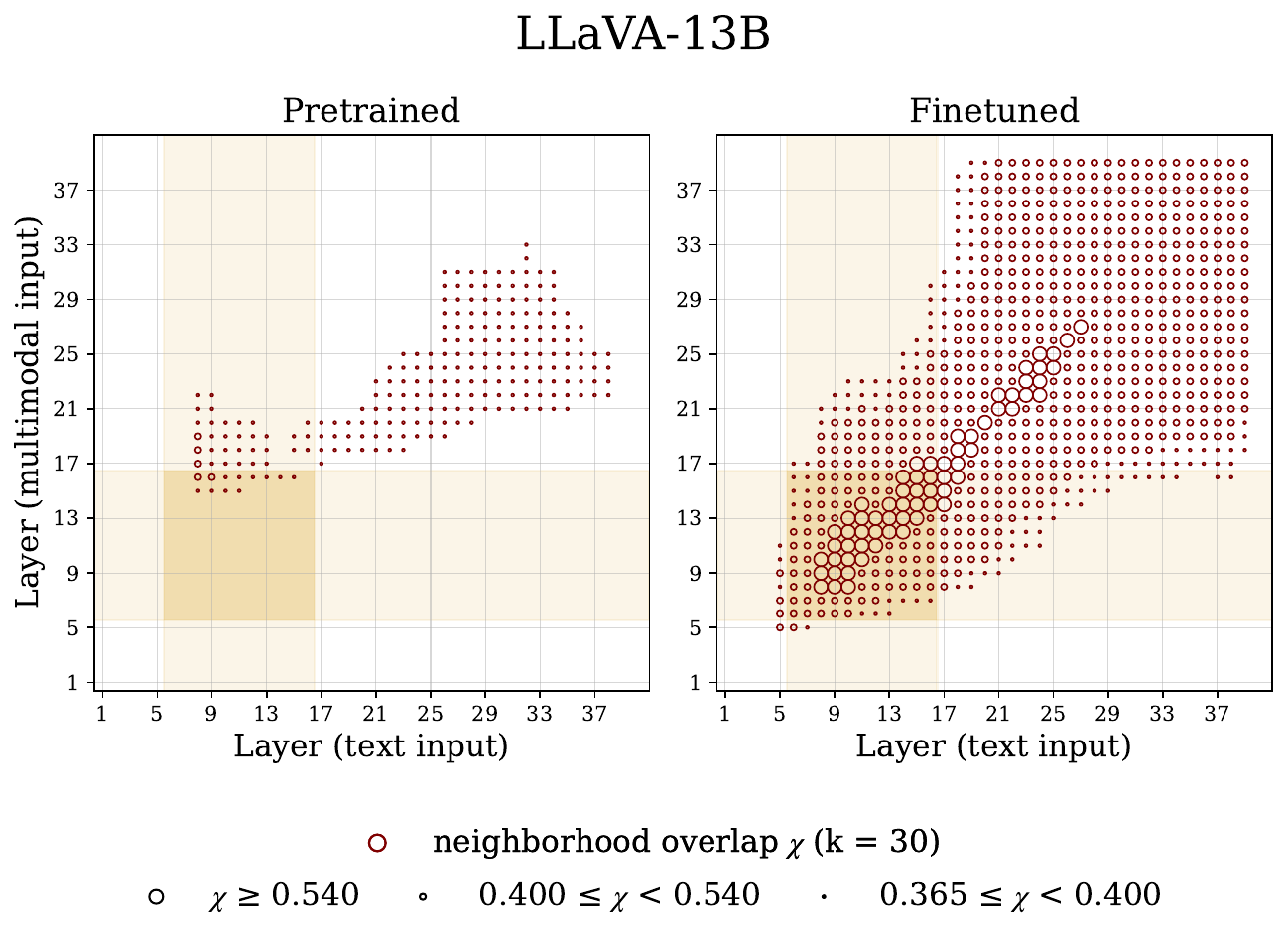}\hfill
\end{minipage}\hfill
\caption{\textbf{Neighborhood overlap between text-only and multimodal representations for LLaVA-1.5 (7B and 13B).} Same layer-pair comparison and marker convention as \Cref{fig:app:linear_cka_llava}, using neighborhood overlap.}
\label{fig:app:neighborhood_overlap_llava}
\end{figure}


\begin{figure}[H]
\centering
\begin{minipage}{0.48\textwidth}
\centering
\includegraphics[width=\linewidth]{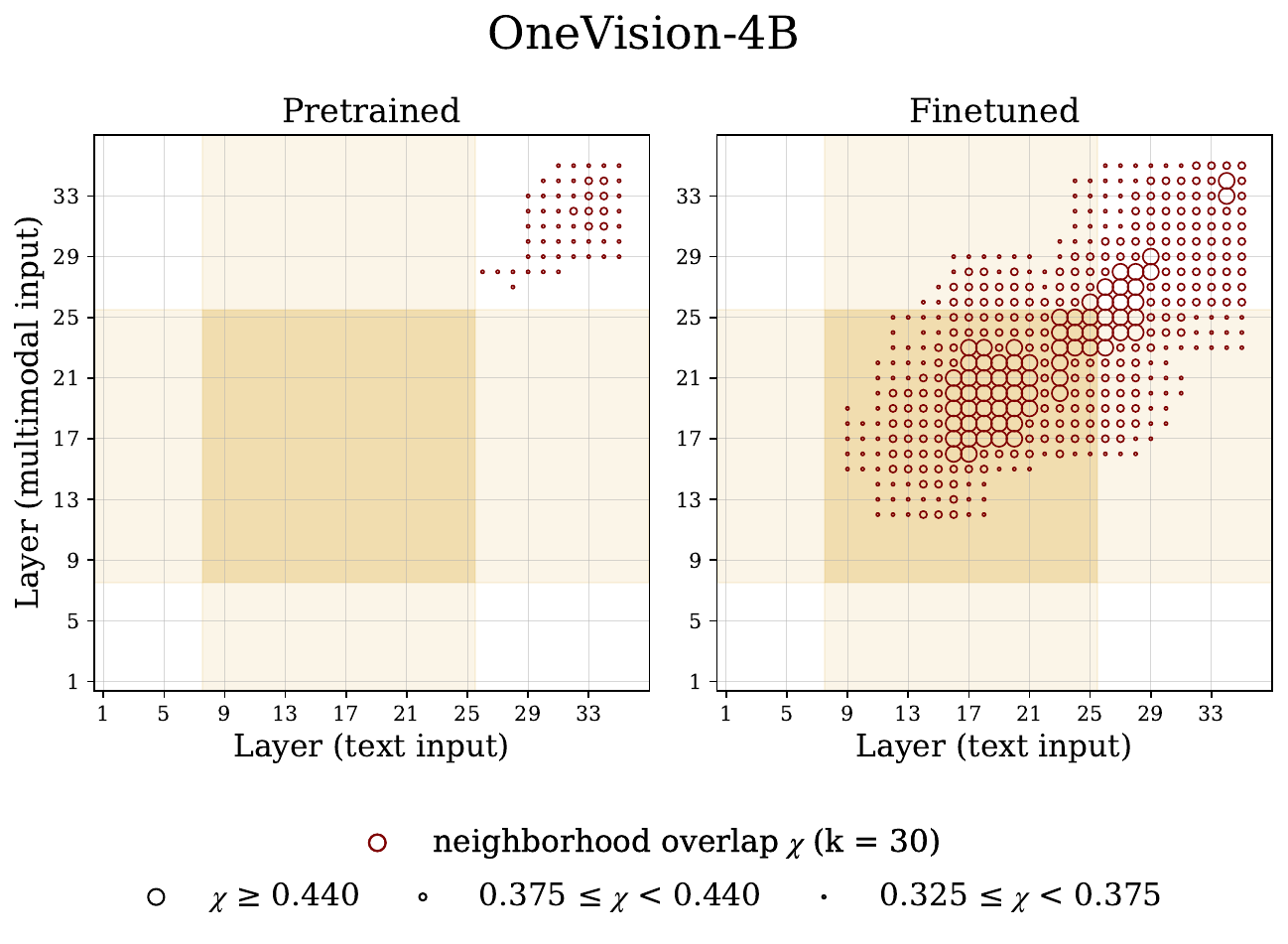}\hfill
\end{minipage}\hfill
\begin{minipage}{0.48\textwidth}
\centering
\includegraphics[width=\linewidth]{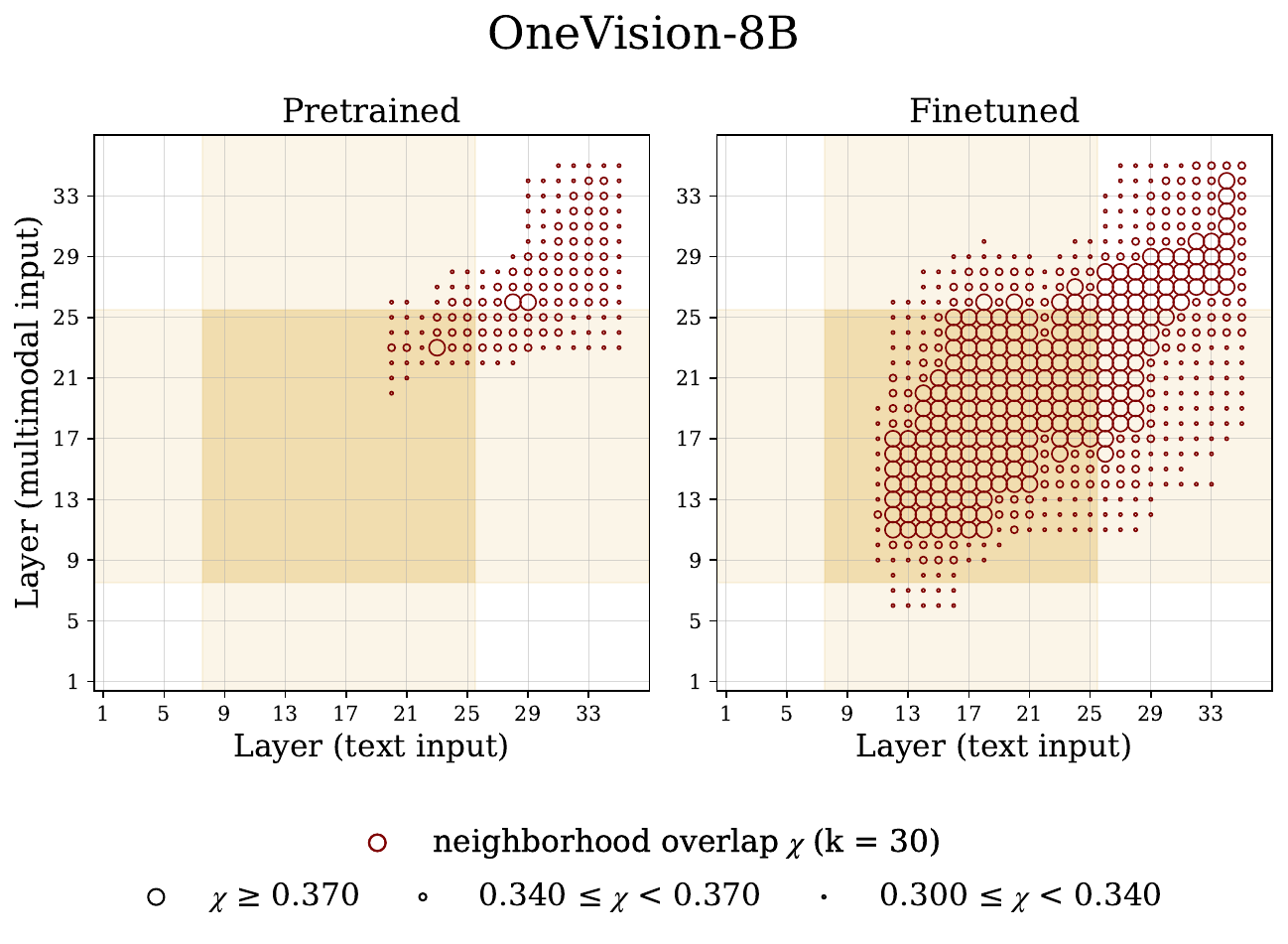}\hfill
\end{minipage}\hfill
\caption{\textbf{Neighborhood overlap for LLaVA-OneVision-1.5 (4B and 8B).}}
\label{fig:app:neighborhood_overlap_onevision}
\end{figure}


\begin{figure}[H]
\centering
\begin{minipage}{0.48\textwidth}
\centering
\includegraphics[width=\linewidth]{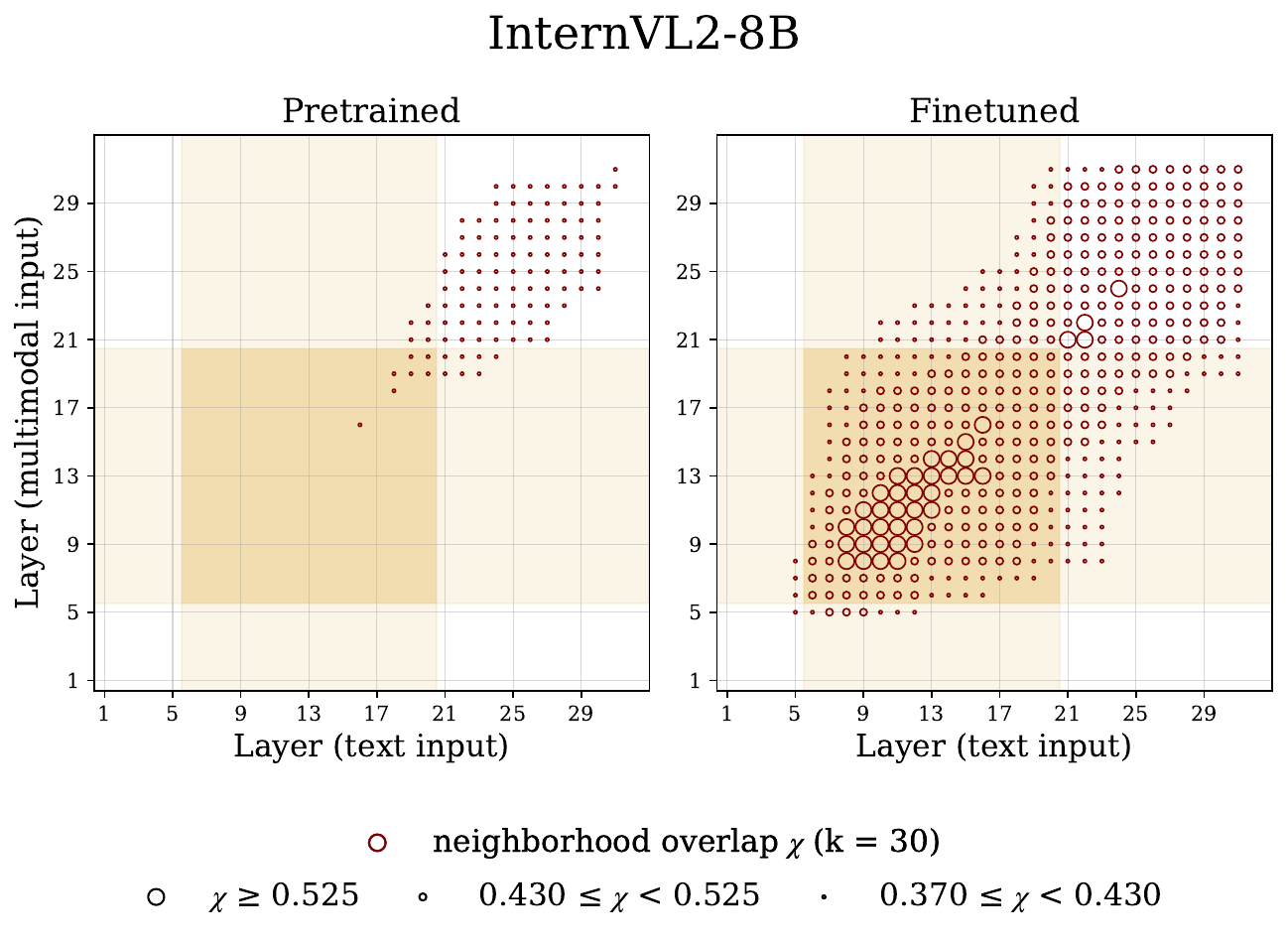}\hfill
\end{minipage}\hfill
\begin{minipage}{0.48\textwidth}
\centering
\includegraphics[width=\linewidth]{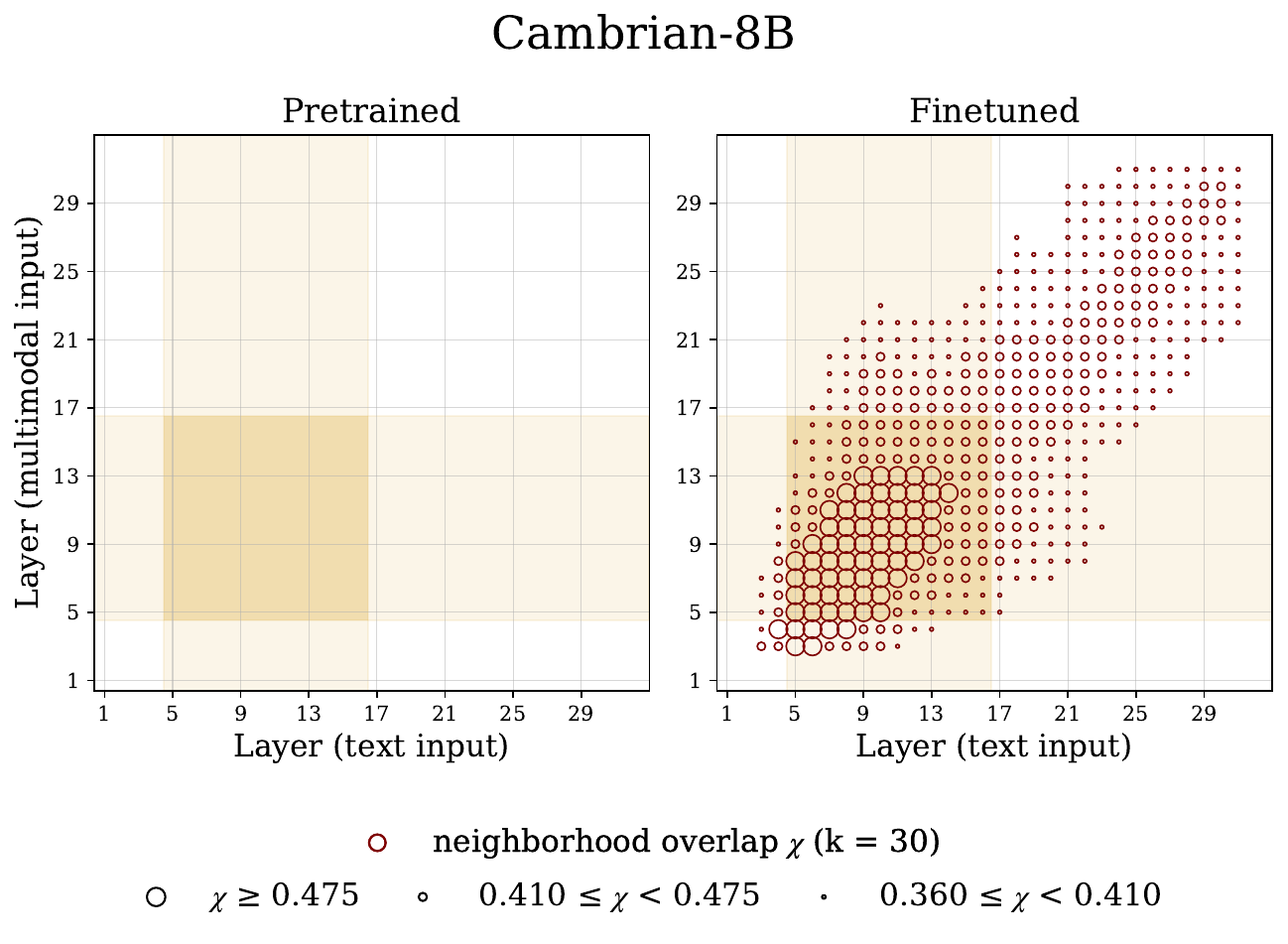}\hfill
\end{minipage}\hfill
\caption{\textbf{Neighborhood overlap for InternVL2-8B and Cambrian-8B.}}
\label{fig:app:neighborhood_overlap_internvl_cambrian}
\end{figure}


\clearpage

\section{Geometric similarity on intermediate sentence tokens.}

\begin{table}[H]
\centering
\small
\resizebox{\columnwidth}{!}{
\begin{tabular}{l|ccc|ccc|ccc}
\toprule
 & \multicolumn{3}{c|}{Pretrained} & \multicolumn{3}{c|}{Fine-tuned} & \multicolumn{3}{c}{Change (\%)} \\
Visual $\backslash$ Text & Early & Middle & Late & Early & Middle & Late & Early & Middle & Late \\
\midrule
Late   & 0.22 & 0.17 & 0.13 & 0.23 & 0.19 & 0.14 & $+6.5$ & $+11.2$ & $+6.7$ \\
Middle & 0.52 & \textbf{0.47} & 0.44 & 0.38 & \textbf{0.33} & 0.29 & $-27.2$ & $\mathbf{-30.3}$ & $-33.3$ \\
Early  & 0.56 & 0.52 & 0.49 & 0.44 & 0.40 & 0.36 & $-21.1$ & $-22.6$ & $-26.1$ \\
\bottomrule
\end{tabular}
}
\caption{\textbf{Cross-modal information imbalance on internal image tokens (LLaVA-1.5-7B).} Mean of the image$\to$text and text$\to$image information imbalance (\Cref{sec:geometry}; lower is better aligned) between image and text representations before and after visual instruction tuning, and the relative change. The image representation is the mean residual stream over the last 30\% of the image tokens (173 of 576) of the multimodal prompt; the text representation is the mean over the caption tokens of the text-only prompt (five human COCO captions for VQAv2 and COCO-QA, one synthetic caption for GQA, MMBench, and SEED-Bench). Values are averaged over these five datasets (3{,}000 samples each; MMBench and SEED-Bench in open-ended format) and over layer blocks: early (1--5), middle (6--17, the abstraction region of \Cref{fig:app:information_imbalance_llava}), and late (18--32).}
\label{tab:imbalance_last_image_tokens}
\end{table}

\clearpage

\section{Extension to multimodal Cross-Attention integration: Llama3.2-90B}
\label{app:llama}
\subsection{Ablation studies: multimodal benchmarks}

\begin{figure}[H]
\centering
\includegraphics[width=\linewidth]{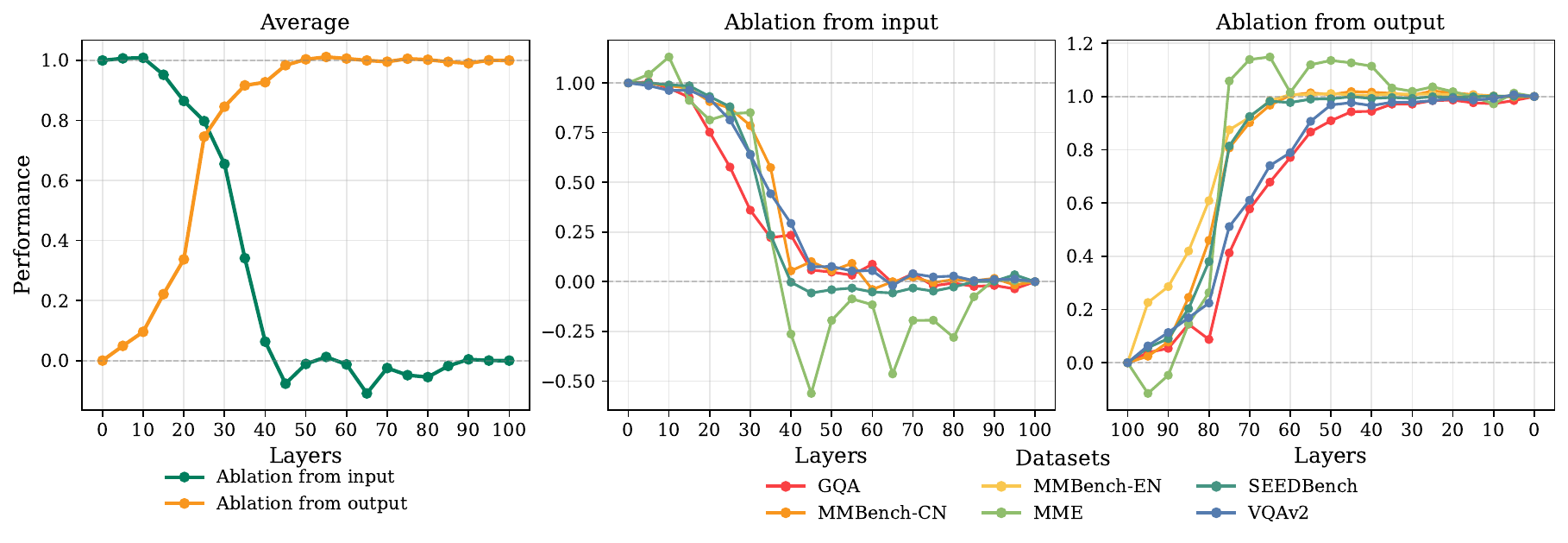}
\caption{\textbf{Cross-attention layer ablations in Llama-3.2-Vision-90B.} Normalized performance when cross-attention layers are ablated progressively from the input or from the output side, averaged over datasets (left) and per dataset (center, right); 5,000 samples per dataset.}
\label{fig:app:llama90b_ablations}
\end{figure}

\subsection{Information imbalance}

\begin{figure}[H]
\centering
\includegraphics[width=0.4\linewidth]{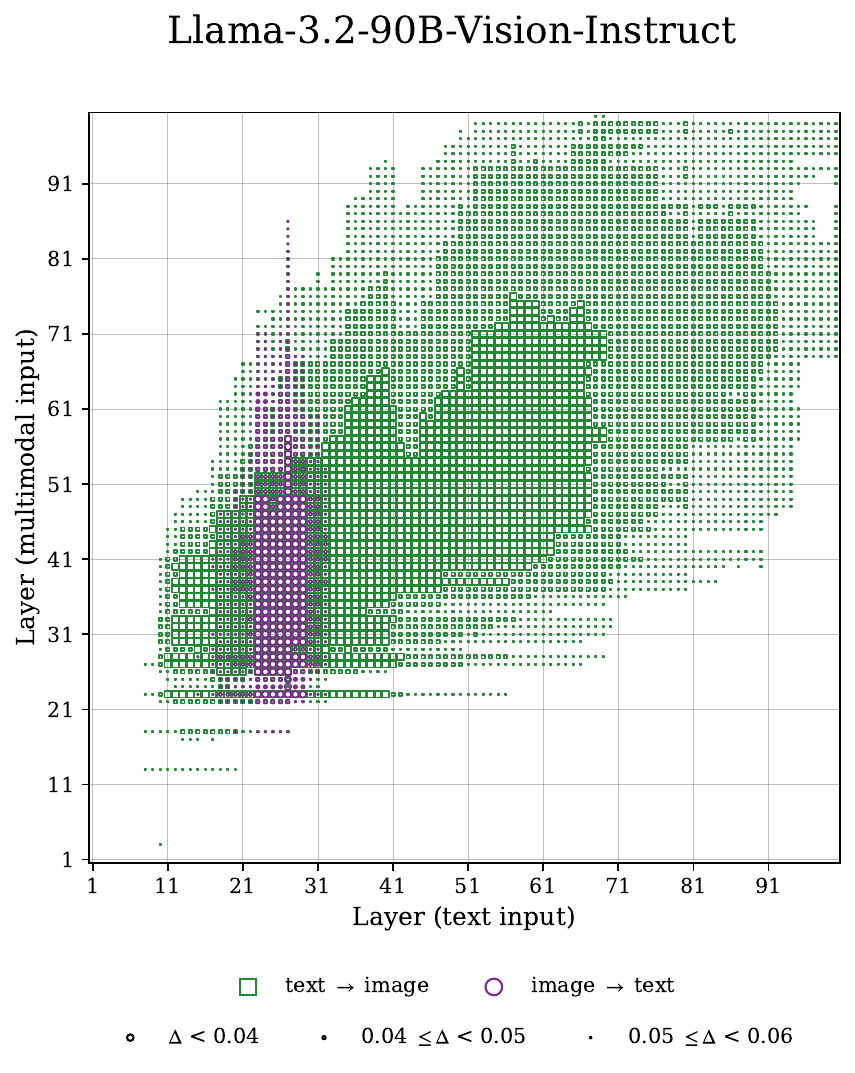}\hfill
\caption{\textbf{Information imbalance plane for Llama-3.2-Vision-90B-Instruct.} Same setup as \Cref{fig:information_imbalance} for the instruction-tuned model, on the full VQAv2, GQA, and MMBench sets.}
\label{fig:app:llama90b_information_imbalance}
\end{figure}

\subsection{Visual reliance correlation}

\begin{table}[H]
    \centering
    \fontsize{8pt}{9pt}\selectfont
\begin{tabular}{llcccc}
\toprule
 Model & \makecell{Layers \\ kept} & \makecell{MC \\ VQA} $\uparrow$ & \makecell{OE \\ VQA}$\uparrow$& Compositional $\uparrow$ & Overall $\uparrow$ \\
\midrule
Llama-3-90B & $[10, 65]$    & 108.5 & 96.8 & 96.7  & 103.5 \\
            & $[10, 65]^C$  &  34.0 & 45.0 & 50.5  &  39.5 \\
            & $[15, 65]$    & 104.1 & 94.3 & 93.6  &  99.8 \\
            & $[15, 65]^C$  &  38.4 & 48.0 & 50.1  &  44.6 \\
\bottomrule
\end{tabular}
\caption{\textbf{Selective activation of cross-attention layers in Llama-3.2-Vision-90B.} Performance when only the cross-attention layers in $[a,b]$, or only those in its complement $[a,b]^C$, are kept active, normalized to the full model (5,000 samples per benchmark). MC VQA averages MMBench-CN, MMBench-EN, MME, and SEED-Bench; OE VQA averages GQA and VQAv2; Compositional reports CV-Bench; Overall averages all available tasks. MMBench results are missing for $[15,65]^C$, so its MC VQA entry averages only MME and SEED-Bench.}
    \label{tab:performance_ft_llama90b}
\end{table}

\begin{figure}[H]
\centering
\includegraphics[width=0.6\linewidth]{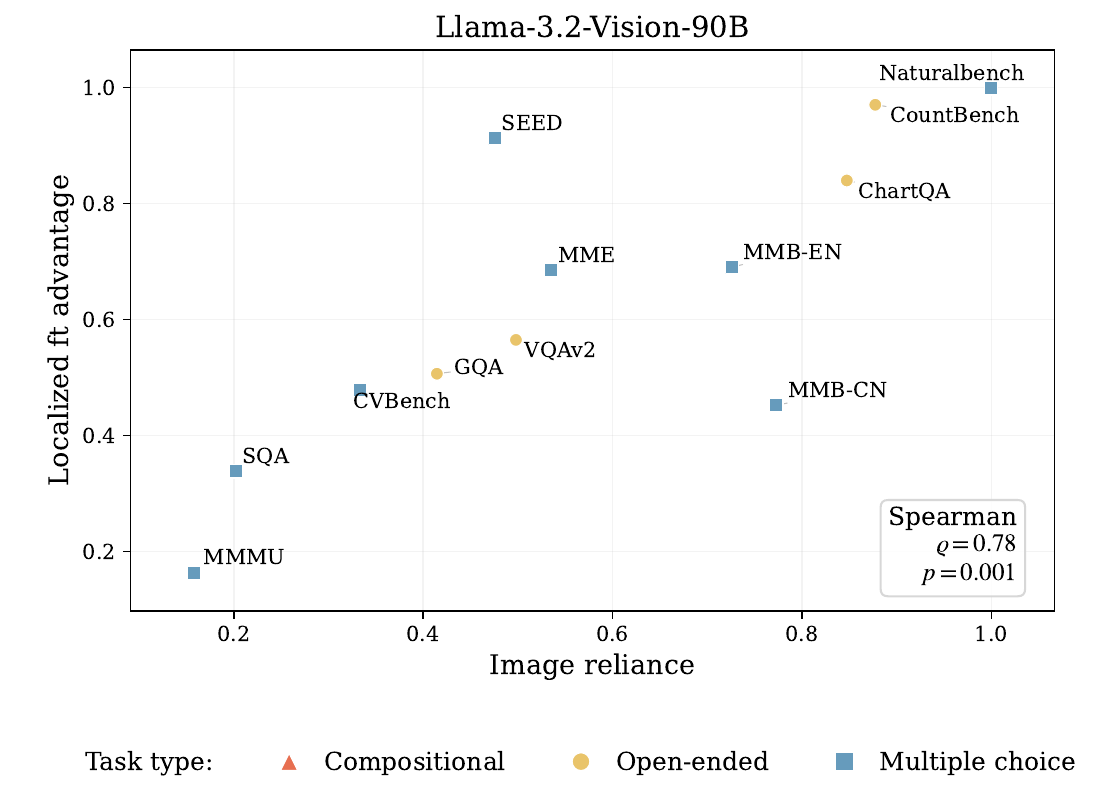}
\caption{\textbf{Image reliance and localized gains in Llama-3.2-Vision-90B.} Same analysis as \Cref{fig:visual_reliance_correlation}, with the y-axis measuring the advantage of keeping active only the cross-attention layers in $[10,65]$ over keeping only their complement (\Cref{tab:performance_ft_llama90b}).}
\label{fig:app:llama90b_visual_reliance}
\end{figure}

\end{document}